\documentclass[11pt,twoside]{article}
\usepackage{epsfig}
\usepackage{graphics}
\usepackage{graphicx}
\usepackage{natbib,amssymb,lineno}

\title{An Image Analogies Approach for Multi-Scale Contour Detection}
\author{Slimane Larabi$^1$, Neil M. Robertson$^2$\\[.5ex]
 $^1$ Computer Science Department, USTHB University, \\ $^2$ Queen's University of Belfast}
\date{}

\chardef\bslchar=`\\ 

\providecommand{\qedsymbol}{\leavevmode
  \hbox to.77778em{%
  \hfil\vrule
  \vbox to.675em{\hrule width.6em\vfil\hrule}%
  \vrule\hfil}}

\catcode`\|=0
\begingroup \catcode`\>=13 
\gdef\?#1>{{\normalfont$\langle$\textit{#1}$\rangle$}}
\gdef\0{\relax}
\endgroup
\def\<#1>{{\normalfont$\langle$\textit{#1}$\rangle$}}

\def\latex/{{\protect\LaTeX}}

\usepackage{url}
\usepackage[breaklinks]{hyperref}

\begin{document}
\maketitle
\markboth{An Image Analogies Approach for Multi-Scale Contour Detection}
{An Image Analogies Approach for Multi-Scale Contour Detection}

\begingroup
\small
\tableofcontents
\endgroup


\newpage 

\section{Introduction}
\label{intro}

Contour detection is an important task in many computer vision applications such as object
recognition, motion, medical image analysis, image enhancement and image compression.


Several authors define contours as the boundaries of objects in an image. This definition
would exclude many situations in which contours do not arise from region boundaries \cite{Papari and Petkov 2011}.

We agree with G. Papari and N. Petkov \cite{Papari and Petkov 2011}
considering that concept of contour is broader than the concept of region boundary and
human judgment is the only possible criterion that can be used in order to say if a given
visual feature is a contour or not. Contours are then defined in a given image as the set
of lines that human observers would concent on to be the contours in that image.

There is wide range of methods in the literature devoted to contour detection
\cite{Ziou and Tabbone 1998}, \cite{Freixenet et al 2002}, \cite{Suri et al 2002},
\cite{He et al 2008} \cite{Papari and Petkov 2011}.
The main problem that has been dealt with in the literature is the modelling of the contour pixel.

The first approaches proposed to contour detection are based on local measurements in image.
Local derivative filters have been proposed by Roberts \cite{Roberts 1965},
Sobel \cite{Duda and Hart 1973}, and Prewitt \cite{Prewitt 1970}.
In the next, Marr and Hildreth \cite{Marr and Hildreth  1980} proposed the use of zero crossings
of the Laplacian of Gaussian operator. The Canny detector \cite{Canny 1986} also models
contours as sharp discontinuities in the brightness channel, adding non-maximum suppression and
hysteresis thresholding steps and becames the most popular differential operator.
Many algorithms have been proposed using others filters \cite{Morrone and Owens 1987},
\cite{Freeman and Adelson 1991}, \cite{Perona and Malik 1990} or for locating contour of texture
\cite{Huang and Tseng 1988}, \cite{Chuang and Sher 1993}, \cite{Ruzon and Tomasi 2001}.

Instead of searching for points where there are sharp changes in intensity, local energy and
phase congruency have been used in many algorithms for contour detection and feature extraction
\cite{Kovesi 1999}, \cite{Perona and Malik 1990}, \cite{Reisfeld 1996}, \cite{Robbins and Owens 1997},
\cite{Venkatesh and Owens 1990}, \cite{Ronse 1993}.
Other important techniques proposed for contour detection concerns active contours methods.
Initially proposed by Kass et al \cite{Kass et al 1988}, this work has been improved tacking
into account topology, distance and gradient vector flow \cite{McInerney and Terzopoulos 2000},
\cite{Cohen and Cohen 1993}, \cite{Xu and Prince 1998}. Other improvement are the Level set
algorithm introduced by Malladi et al \cite{Malladi et al 1995} which doesn't make assumption
about the topology of the objects in the image and the geodesic active contour  based on the
relation between active contours and the computation of geodesics or minimal distance curves
\cite{Caselles et al 1997}.

Contour detection has reached high degree of maturity, taking into account multimodal
contour definition. However, the quality of located contours are still far from what people
can do. This may arise from the missing  of human knowledge (high level of vision) and human
expertise to hand draw contours (low level of vision) in the different proposed approaches.


Indeed, humans can do easily this and results are  known to be very similar from person-to-person.

\begin{figure}[ht!]
\centering
 \includegraphics[width=10cm]{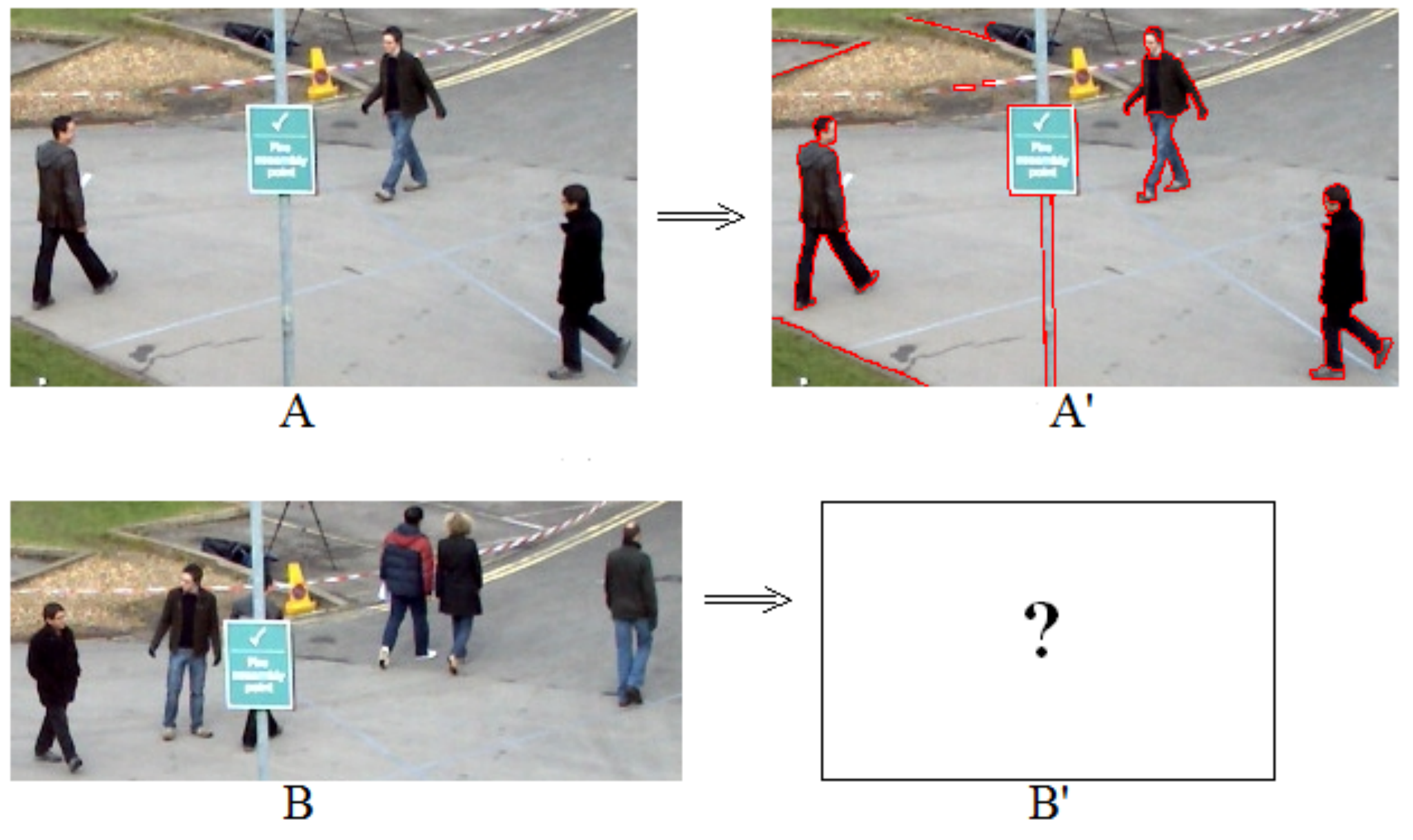}
  \caption{Contour detection by analogy: the basic principle proposed in this work is to use previously
  detected contours to find new ones in similar images: Given the pair of training images $(A, A')$ where
  the contours in $A$ are hand drawn, our aim is to locate the contours of a query image $B$ in the same
  way that it has be done for $A$ }
  \label{fig1n}
\end{figure}

The aim of this work is to introduce image analogies in early stages of computer vision, to model human
expertise and to pass it to the computer for contour detection. Indeed, image analogies constitutes
a natural means of specifying filters and image transformations \cite{Hertzmann et al 2001} and
we can supply an appropriate exemplar and say, in effect, ``Find me pixels which look like this''. The concept
is illustrated in figure \ref{fig1n}.

\subsection{Contributions}

\begin{figure}[ht!]
\centering
\includegraphics[width=4 cm]{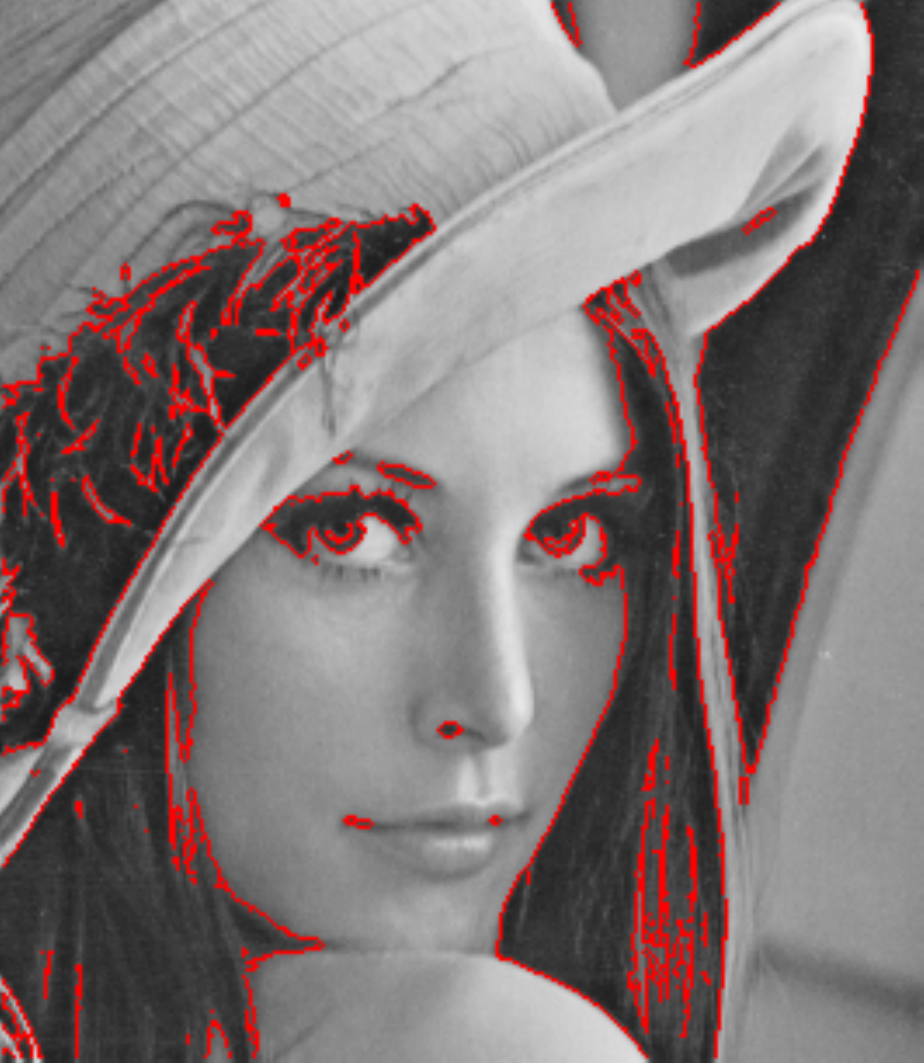}
\includegraphics[width=4 cm]{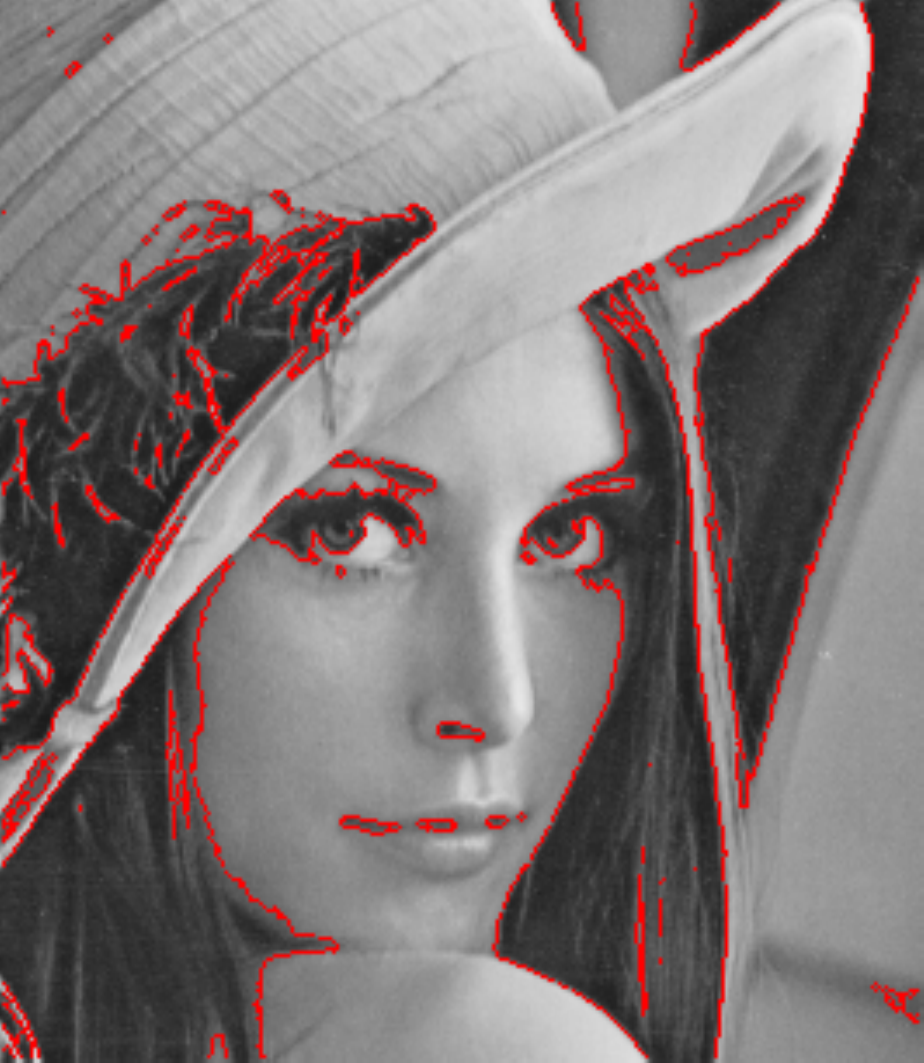}
\includegraphics[width=4 cm]{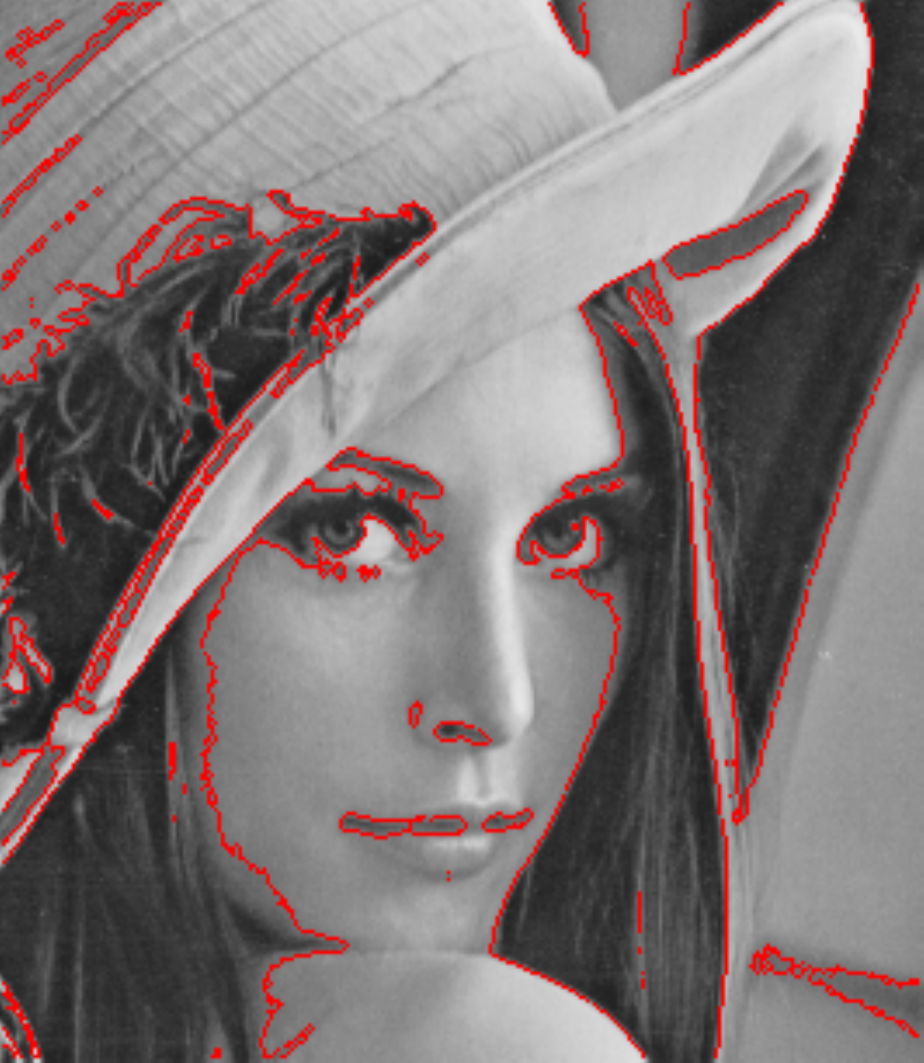}
\includegraphics[width=4 cm]{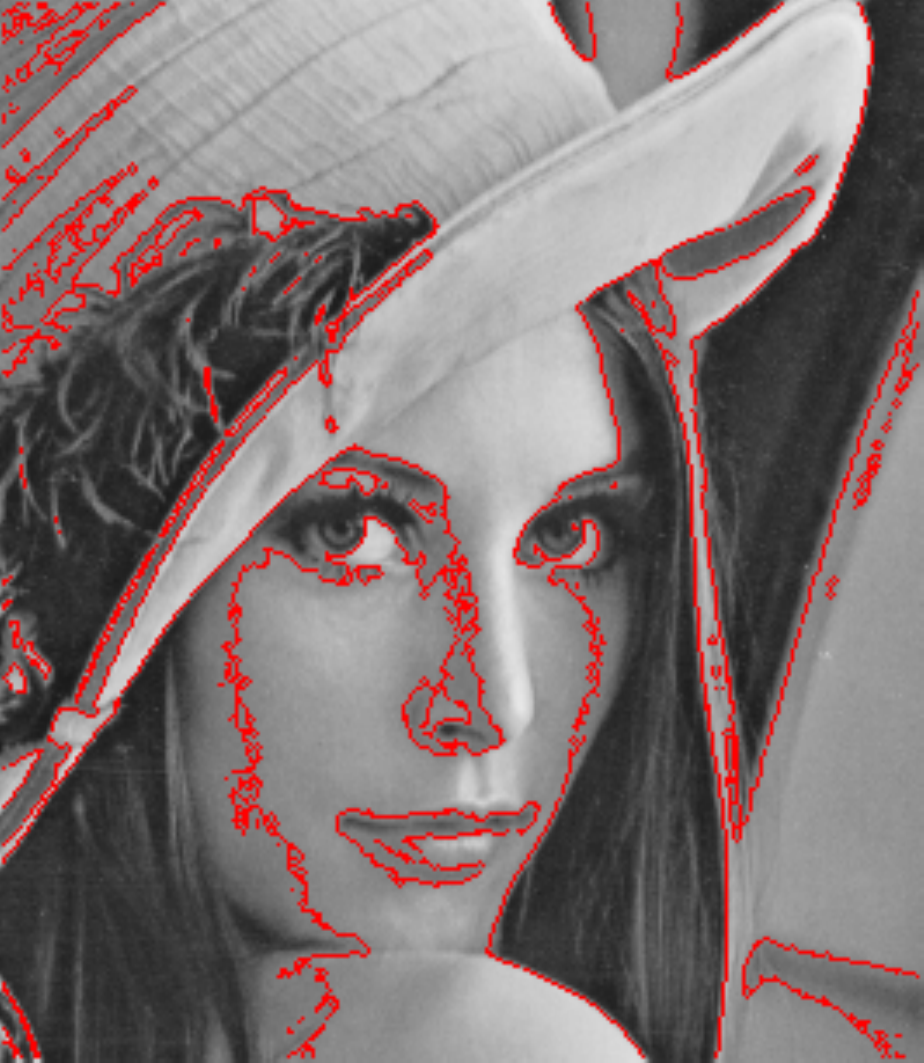}
\caption{Contours located by respectively by the pairs of patterns $(P_5, P'_5)$, $(P_6, P'_6)$, $(P_7, P'_7)$, $(P_8, P'_8)$}\label{fig5n}
\end{figure}

Our contributions are:

- First, Image analogies principle is applied naively to contour detection. Due to the high complexity of this task,
a set of $14$ artificial pairs of patterns $(P_i, P'_i, i=1..14)$ are derived from a
mathematical reasoning so that any contour pixel will be located whatever the lighting conditions in image.
Contours located are related to regions boundaries and are in evolution from darker region to the clearer region
(see figure \ref{fig5n}) where located contours using $4$ pairs of patterns are illustrated).

\begin{figure}[ht!]
\centering
\includegraphics[width=10 cm]{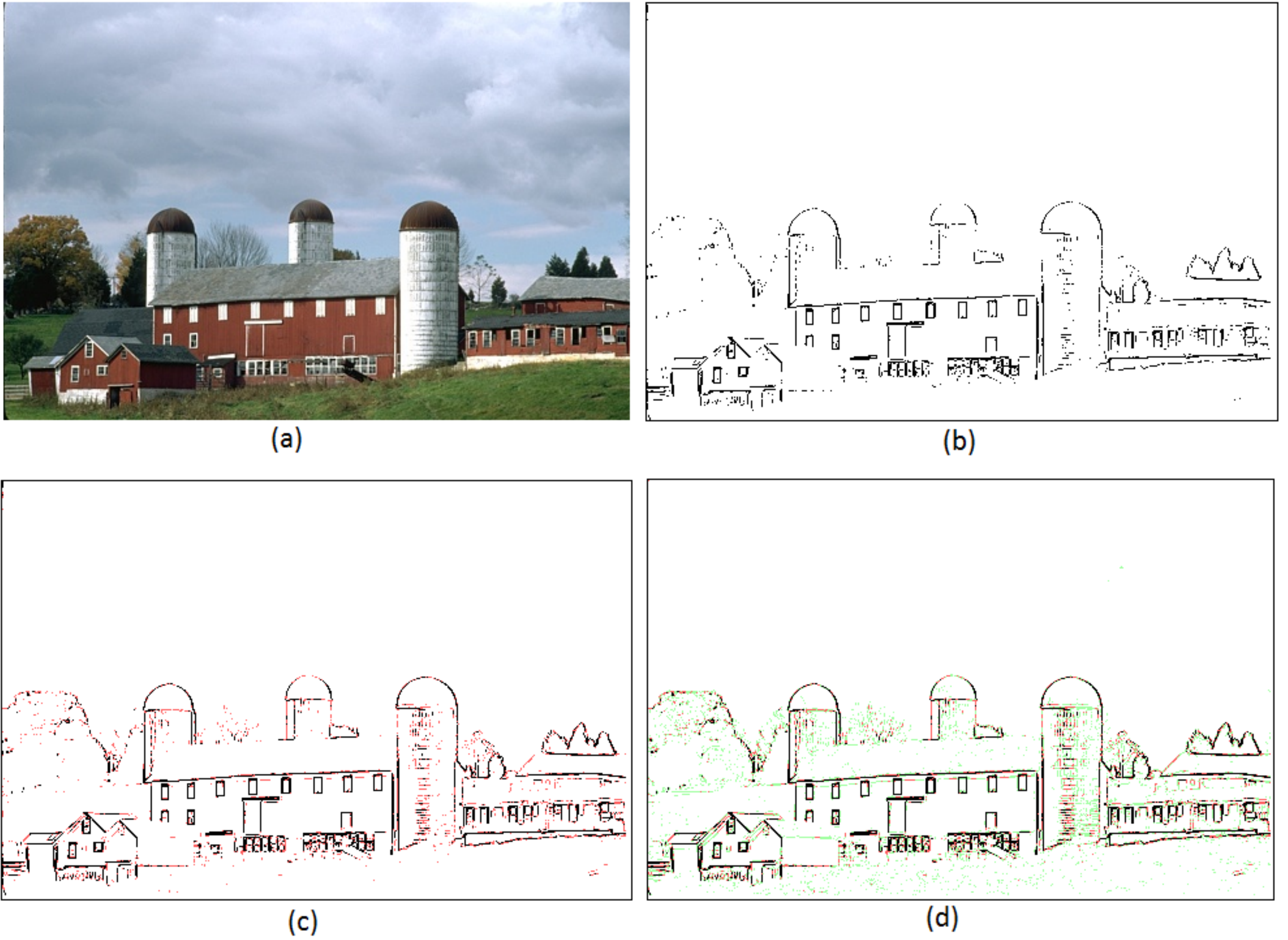}
\caption{(a) A query image, (b) Contours of Low Resolution, (c) Contours of Intermediate Resolution,
new located contours are illustrated with red color, (d) Contours of High Resolution, new located
contours are illustrated with green color}\label{fig10n}
\end{figure}

- Contours are computed for a query image at different scales.
At the low resolution only the low frequency $(LF)$ contours are visible corresponding to large differences
in intensity between regions. The more the resolution increases, the more there are contours
corresponding to intermediate and high frequency $(IF), (HF)$. Figure \ref{fig10n} shows
computed contours by image analogies illustrating this point where the black, red, green are the color
of contour pixels of level $(LF)$, $(IF)$ and $(HF)$.

- Our method requires the fixing of three parameters whose values are known and the result of
contours detection doesn't depend from any other parameter and are unique for all images.

\subsection{Paper roadmap}

In section \ref{A naive}, we present our approach for contour detection using image analogies principle.
Based on stored information in the reference images, contour pixels in a query image are located
applying image analogy technique. A naive application is proposed and the limits of this
approach is explained. Indeed, it is necessary to have many reference images, otherwise some contour
pixels will be not located in the new image since their appearance may not be represented in the training data.
To deal with this constraint, we study  in section \ref{The basic} the required constraints for training images
so that all contour pixels will be located for any query image.
Artificial pairs of patterns are derived from this study and used as reference instead of real images. We explain in section \ref{Making} how these patterns are build.
In section \ref{Algorithm} we study the complexity of the proposed approach and we gives the improvements
made for reducing this complexity. Some details of the implementation and the algorithm are given.

Different data sets including the Berkeley Segmentation Data (BSDS500) \cite{Arbelaez et al 2011},
Weizmann Horses \cite{Borenstein and Ullman 2002} are used to validate this approach. Obtained results their evaluation are presented in Section \ref{Results}.
We conclude this paper with propositions for integrating image analogies at other stages of image analysis.




\section{Image Analogies for Contour Detection: A Naive Application}\label{A naive}

A human is able to detect and draw natural image contours. Applying image analogies principle,
our aim is to locate contours as accurately as a human does it, including within images of low
resolution where objects have small sizes.

\begin{figure}[ht!]
\centering
  \includegraphics[width=12 cm]{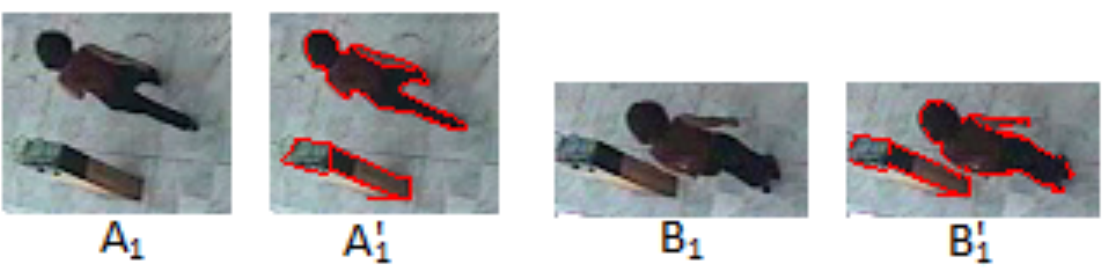}
  \caption{(Left) First training pair $(A_1, A'_1)$ where drawn pixel contours
  are highlighted with red color in $A'_1$, (Right) Query image $B_1$ and the computed image $B'_1$}
  \label{fig13n}
\end{figure}

Let $A$ be the initial image. We assume that contour pixels are manually located on $A$
and marked giving a synthesized image $A'$ (see figure \ref{fig13n} where pixels contours are
highlighted with red colour).
Given a query image $B$, the problem is then how to compute the synthesized image $B'$
that contains contours located and highlighted in the same way as those located in $A'$.
The key idea is to classify each pixel $q$ of $B$ using the
knowledge that may be inferred from $(A, A')$: for each pixel $q$, the synthesized pixel $q'$
will be the same pixel as $q$, in addition it will be marked contour pixel if the pixel $p'^*$ in $A'$
associated to $p^*$ in $A$ is marked so as $p^*$ is the best match of $q$.
The similarity measure considered as Euclidean distance is computed taking into account the neighbours of
$q$ and $p$ and concerns only the brightness of pixels.

In the case where the query image has the same background \textbf{like} the training pair of images $(A, A')$, the algorithm find
almost all contour pixels because the best match $p^*$ of a query contour pixel $q$ will be found that it
coincides with a marked pixel (see figure \ref{fig13n}).

\begin{figure}[ht!]
\centering
  \includegraphics[width=10cm]{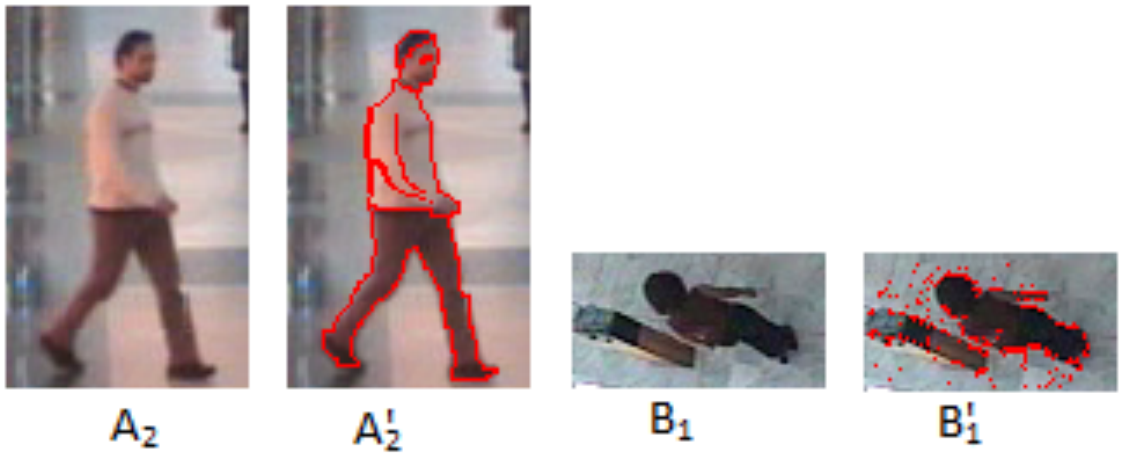}
  \caption{(Left) Second training pair $A_2, A'_2$ with different brightness
  than the first one, (Right) For the same query image $B_1$, the computed $B'_1$ where many contour
  pixels are not located} \label{fig14n}
\end{figure}

\begin{figure}[ht!]
\centering
  \includegraphics[width=3cm]{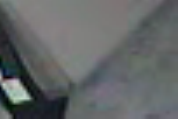}
  \includegraphics[width=3cm]{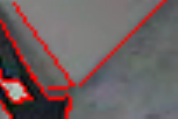}
  \includegraphics[width=3cm]{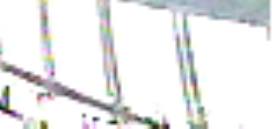}
  \includegraphics[width=3cm]{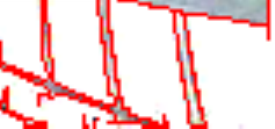}
 \caption{A third and fourth training images pair $(A_3, A'_3)$, $(A_4, A'_4)$ }
  \label{fig15n}
\end{figure}

\begin{figure}[ht!]
  \includegraphics[width=4cm]{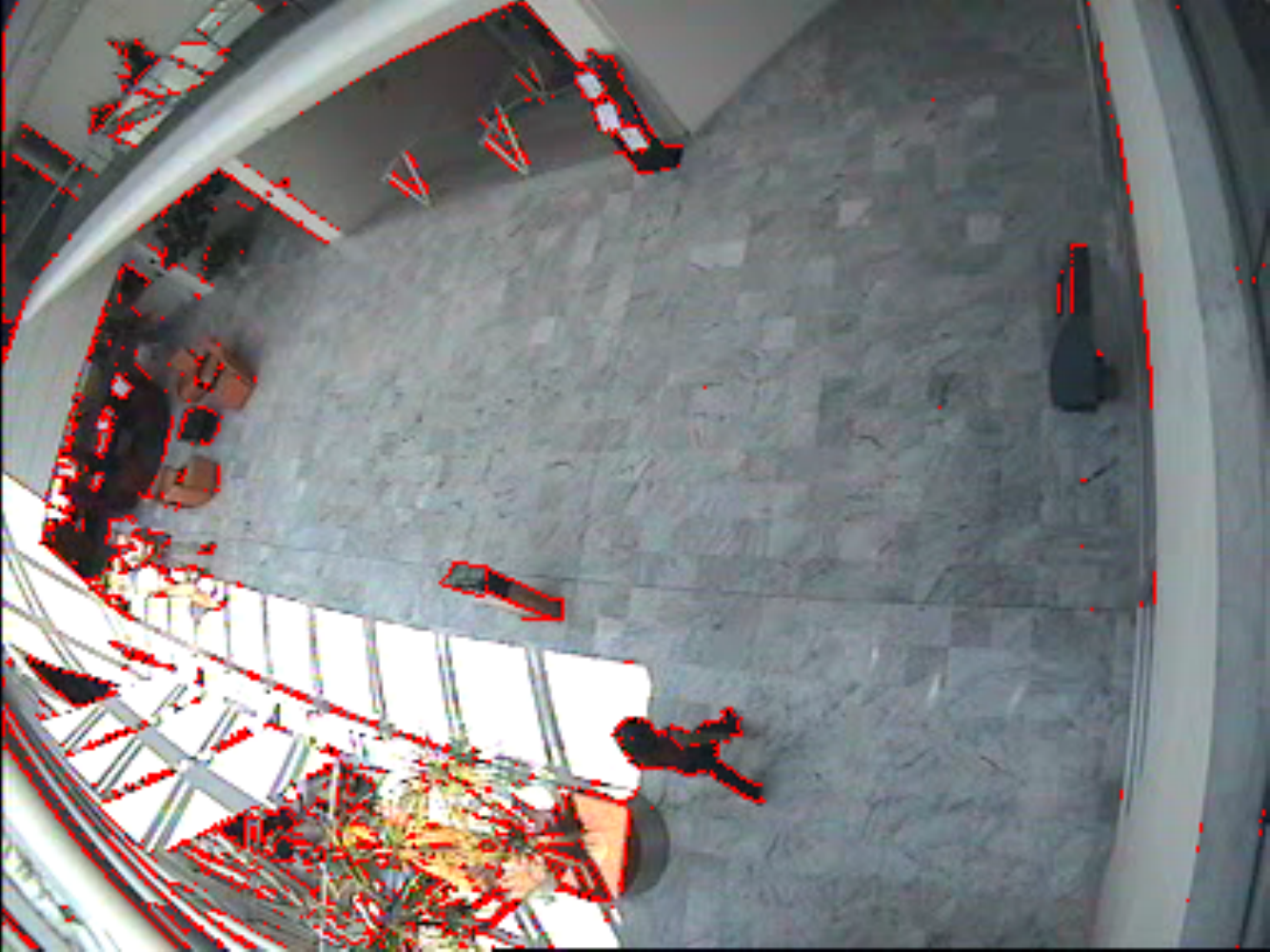}
  \includegraphics[width=4cm]{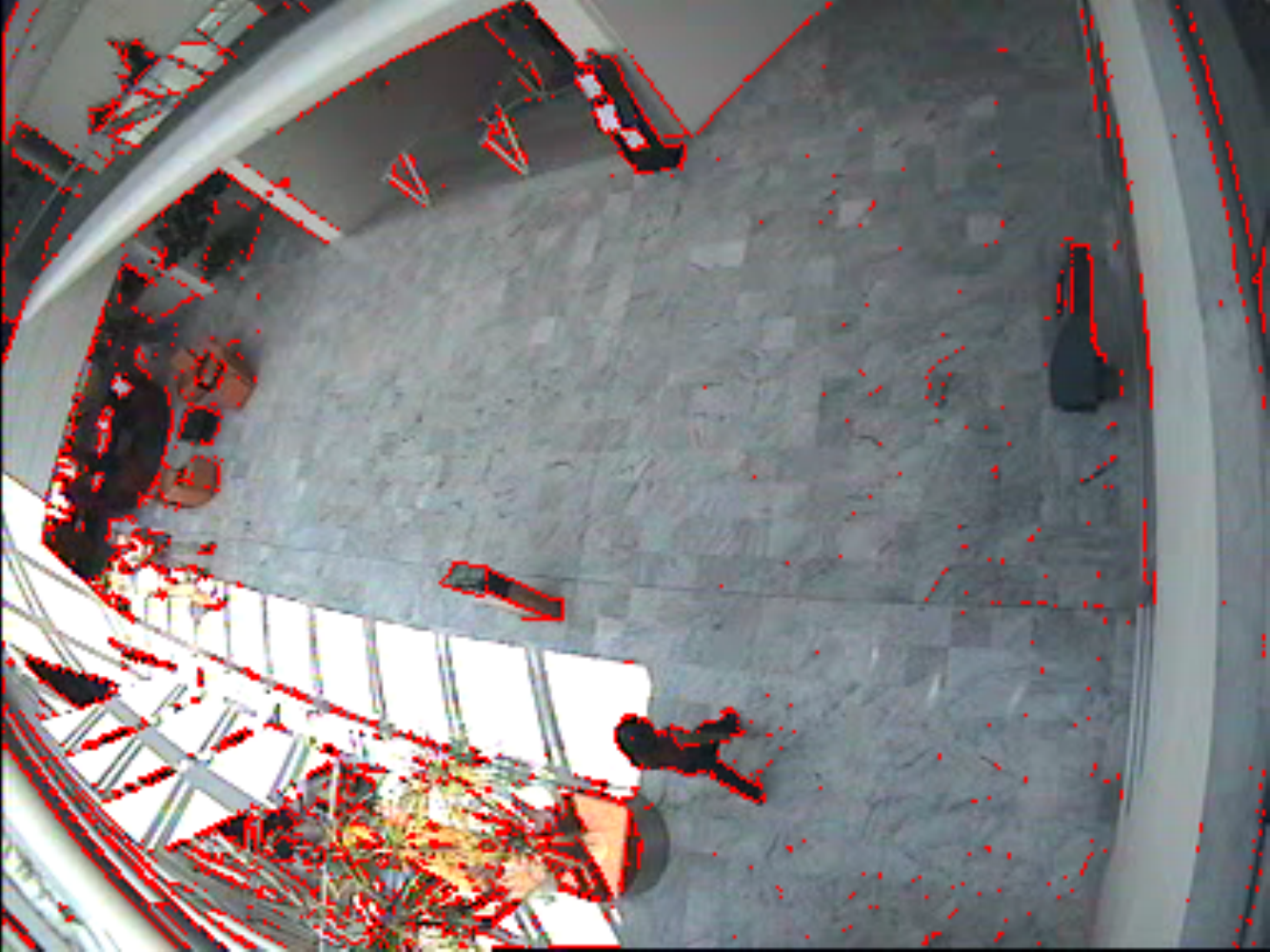}
  \includegraphics[width=4cm]{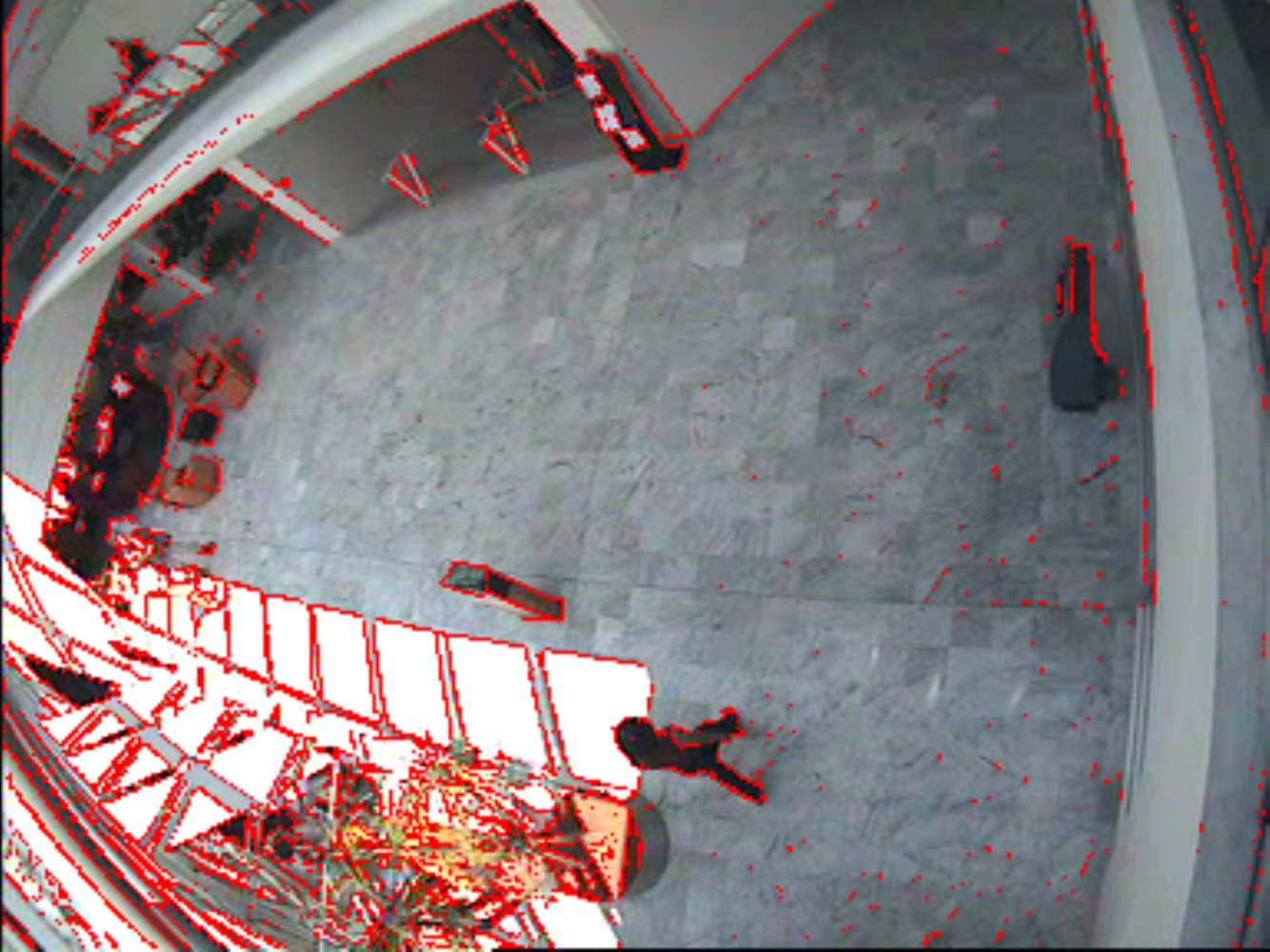}
 \caption{From the left to right: the computed $B'_2$ for a second query image $B_2$ respectively using (first),
 (first and third),(first, third and fourth) training images pairs. We note the increasing of located contour
 pixels when more training images related to the query image are used}
  \label{fig16n}
\end{figure}

The naive application of image analogies for contour detection has been our first task: we
found that this method does not work. Figure \ref{fig14n} shows an example where the training
images pair is not sufficient to locate all contour pixels in the query images because its brightness is
completely different from that of the images training.
However, if we increase the number of learning images, this will increase the probability to match correctly
query pixels, and then the result of image analogies may be better if there are similarity between query
image and some reference images (see figures \ref{fig15n}, \ref{fig16n}).
This way to improve the quality of contour detection implies a considerable increasing
of time processing.

When the query images are of the same nature as the training images, the results may be good.
This is the case in \cite{Lackey and Colagrosso 2004}, image segmentation is done in this work applying
directly image analogies technique in particular case of set of consecutive Visible Human slices:
${m_0, m_1, . . . , m_n}$.
Given a human segmented image $s_0$, Image Analogies is applied firstly using the same segmented
$s_0$ for all $m_i$ and progressively for $s_i$, the computed segmented image $s_{i-1}$ is used.
The performance is improved significantly:
If other query image $m_i$ (different from the slices) is used, the result of image analogies will depend on
the texture, color of the regions contained in image $m_0$. If $m_i$ and $m_0$ are different, poor
segmentation will be obtained. This is due to the step of pixel matching between $m_i$ and $m_0$. If we
cannot found for a query pixel $q$ in $m_i$ a good match $p$ in $m_0$, then the pixel $q$ in $s_i$ will
be misclassified.

We investigated in this work, how can we avoid this constraint (more learning images are required) in order
to guarantee that all contour pixels will be located for any query image.
This limitation may be avoided using many pairs of training images in different conditions of
illumination. This will increase the processing time and also it is a hard task to acquire sufficient
manually-labelled data. In the next section, we propose a new way to deal with this limitation which
allows independence from the task of obtaining manually-located outlines.


\section{The basic principle of our method}\label{The basic}

\begin{figure}[ht!]
\centering
  \includegraphics[width=12 cm]{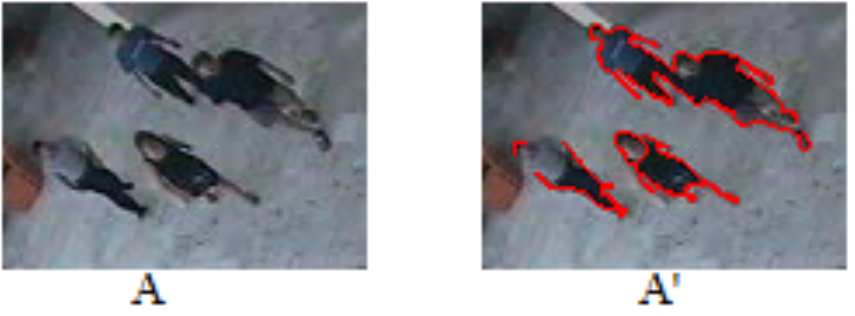}
  \caption{Initial image $A$ and the synthesized image $A'$ obtained as $A$ in addition contour
  pixels are marked (illustrated with red color). Note here that the difficulty to locate contours in
  such images of low-resolution is due to the similarity of the background and some parts of objects}
  \label{fig17n}
\end{figure}

Before the describing of the proposed method, we discuss the scope of human expertise for
contour detection. Given an image $A$, we believe that the human takes into account
two criteria for locating contours. The first one is the neatness of the difference of gray level
intensity (or colour) between two neighboring sets of pixels. The second one is the knowledge of
outline shape geometry inferred from context or some features such as outlines of dominant parts
\cite{DeWinter and Wagemans 2006}.
Indeed, during the process of outline drawing, a human cannot localize some parts of the outline due
to the high similarity between pixels of background and object part but can avoid this difficulty
using the prior knowledge (see figure \ref{fig17n}).

In this paper we deal only with the first criterion and we present our approach in order to allow 
computer locating contour pixels in similar way that human do. We model hand drawing contours, using only the brightness feature.

Let $(A, A')$ be pair of training images such that $A'$ be the synthesized image identical to $A$,
in addition contour pixels are marked.
Let $B$ be a query image. Applying image analogies principle means that for each pixel $q$ of $B$, its best
match $p^*$ is searched in $A$ using:\\
- The  brightness similarity between the neighbors of $p$ and $q$ (\textbf{best approximate match}).\\
- The selected $p^*$ must verify the \textbf{best coherence match}, which means if the neighbors
of  $q$ are pixels of contours, it will be also the case for the neighbors of the selected $p^*$.

\subsection{The best approximate match}

\begin{figure}[ht]
\centering
  \includegraphics[width=12 cm]{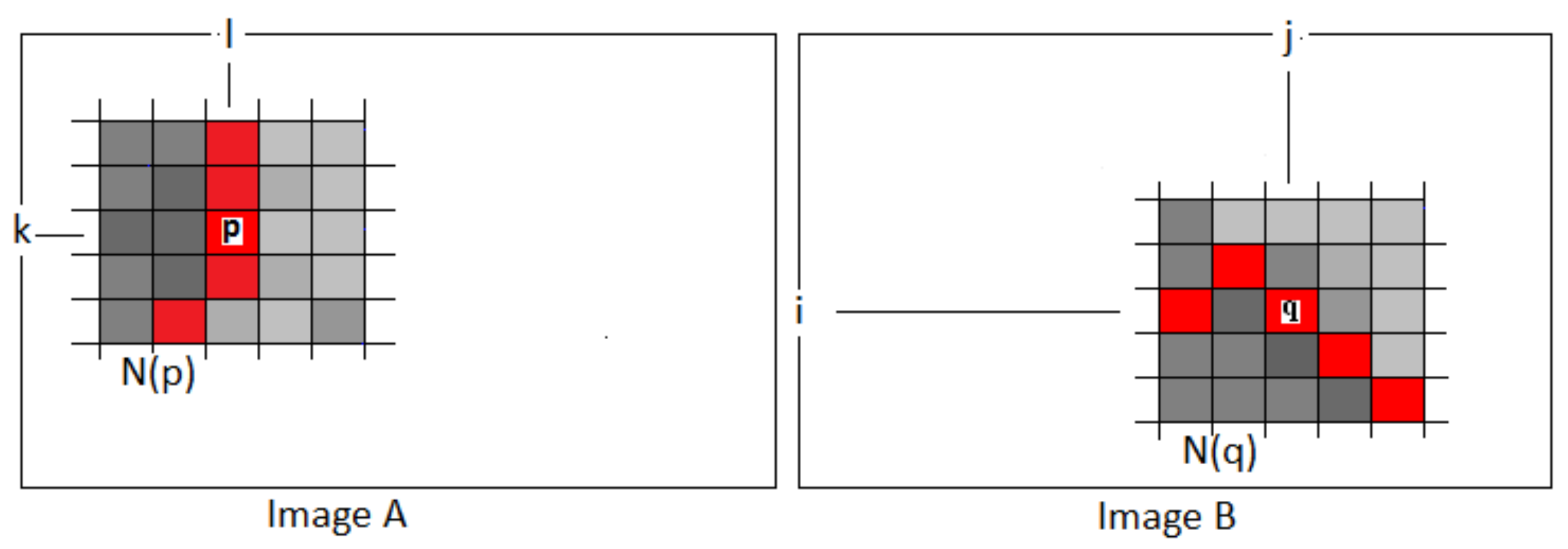}
  \caption{Example of two neighbors $N(p), N(q)$, in red color boundary pixels}
  \label{fig18n}
\end{figure}

Let $N(p)$, $N(q)$ be the $(m \times m)$ neighborhood of $p, q$ in images $A,B$.
Our aim is to search in $A$ the best match $N(p^*)$ of $N(q)$.
The similarity measure $S(q,p)$ between $N(q)$ and $N(p)$, given by equation \ref{eq1}, is computed as the
Euclidian distance between the intensities of corresponding pixels in $N(q)$ and $N(p)$ (see figure \ref{fig18n}).
{\small
\begin{equation}\label{eq1}
    S(q,p)= \sum_{u=-w}^{u=+w} \sum_{v=-w}^{v=+w} (N(q)(i+u, j+v)- N(p)(k+u,l+v))^2
\end{equation}
}
Where:

\begin{enumerate}
\item $(i,j), (k,l)$ are the coordinates of the pixels $q$, $p$ in images $B$, $A$
\item $N(q)(i+u, j+v)$, $N(p)(k+u,l+v)$ are the intensities of pixels $(i+u, j+v)$ and $(k+u,l+v)$
in images $B$, $A$.
\item $m \times m$ is the size of $N(p)$ and $N(q)$ and $w=(m-1)/2$
\end{enumerate}

The proposed similarity measure must guarantee that any pixel $q$ cannot be
misclassified if the knowledge inferred from $(A, A')$ is sufficient.

Based on the proposed similarity measure, we study in annex \ref{annex} the necessary constraints that must
be verified in the training images $(A, A')$ in order to guarantee that all contour pixels and only
contour pixels will be selected.

\subsection{The best coherence match}\label{coherence}

\begin{figure}[ht]
\centering
  \includegraphics[width=9cm]{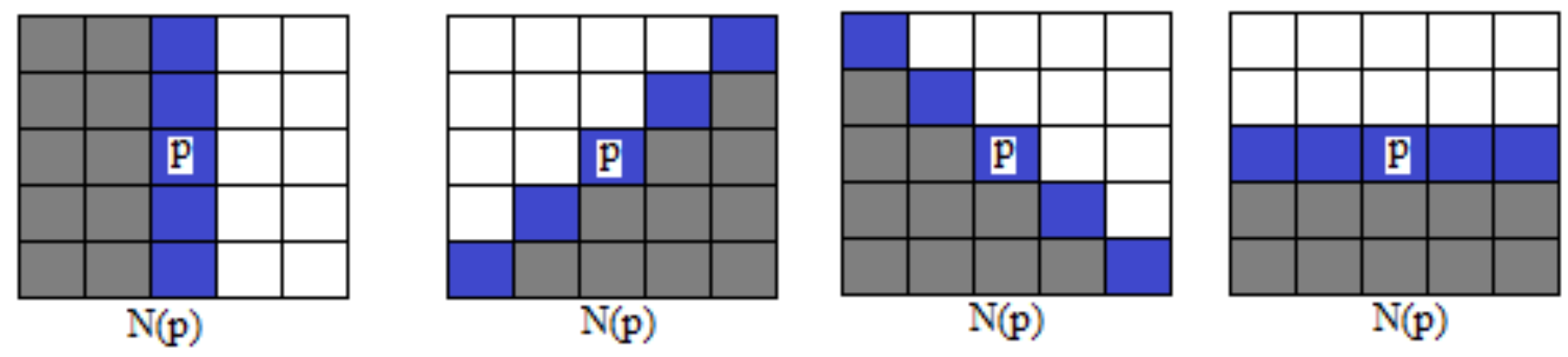}
  \caption{Case of four directions of the boundary in $N(p)$, in blue color are represented contour pixels}
  \label{fig35n}
\end{figure}

This criterion is considered in order to favour the matching of aligned pixel contours in both
neighboring $N(q), N(p)$.  Then, in addition to the lighting conditions, it is necessary to have all directions of
contours pixels in the training images. We consider then the presence of $n$ directions of contours in $A, A'$.
For example if $n=4$ (see figure \ref{fig35n}), the directions are horizontal, vertical and the two diagonals
directions are considered.

Considering these occurrences in the training images, any query $N(q)$ will be matched with $N(q)$ having
the same direction of the boundary and nearest intensities corresponding to the minimal value of the similarity measure.

\section{Making the Training Images}
\label{Making}


From the conducted study presented in annex \ref{annex} devoted to the constraints required for the pairs of  training images, there are some constraints that must be verified in the image $A$ in order
to ensure that all  $q$ outline pixels will be correctly classified, otherwise, the contour pixel $q$
will be misclassified. 

Two constraints on the value of $d_A$ must be verified (see annex \ref{annex}):\\

{ \small
if $d_A>0$:
\begin{equation}\label{eq1ref}
  2d_b+2\bigtriangleup I^b_A<d_A<2d_B+2d_b-2\bigtriangleup I^b_A
\end{equation}
}
if $d_A<0$:
{ \small
\begin{equation}\label{eq3}
  2d_B+2d_b+2\bigtriangleup I^b_A<d_A<2d_b
\end{equation}
}

\textbf{Case 1}

Our goal is to have in $A$ the located contour pixels which verify for each $N(p)$:
{ \small
\begin{equation}\label{eq4}
  d_A \in ]2d_b+2\bigtriangleup I^b_A, 2d_G+2d_B-2\bigtriangleup I^b_A[
\end{equation}
}
This interval depends on the image $B$, particularly on the values $I^b_B, I^f_B$ of each $N(q)$.

It is unrealistic to have all possible pairs of images $(A, A')$ whose located contour pixels verify
the constraint given by the equation \ref{eq1ref} for all pixels $q$ of any $B$ image.\\
What we propose here is to use artificial images $(A,A')$ which allow to have for any $N(q)$ the
constraint satisfied.

\begin{figure}[ht]
\centering
  \includegraphics[width=8 cm]{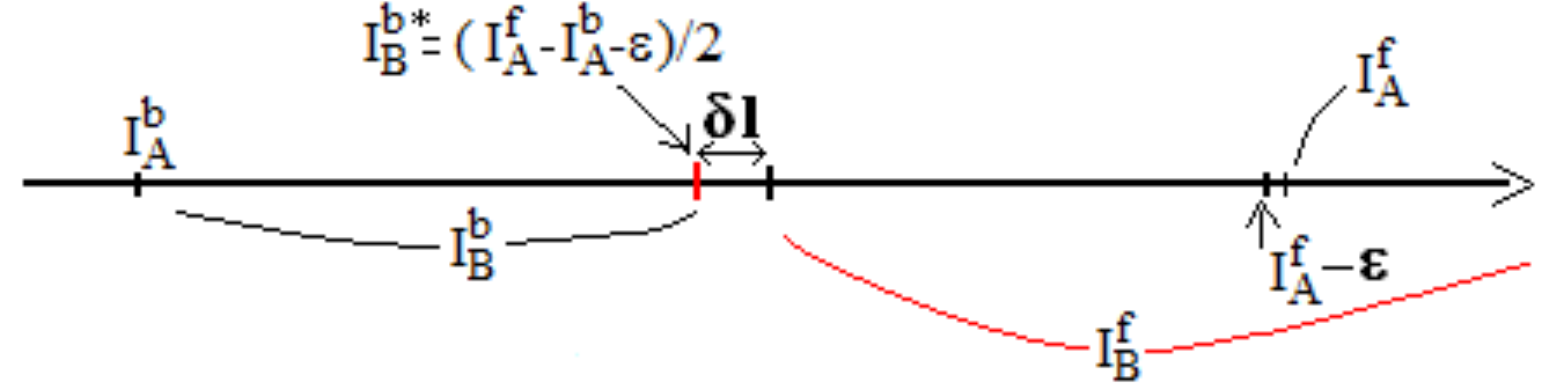}
  \caption{For given values $(I^b_A, I^f_A)$ of $N(p)$, all $N(q)$ such that $I^b_B<I^{b*}_B$ and
  $I^f_B \geq I^{b*}_B+ \delta l$ satisfy the equation \ref{eq1ref}}
  \label{fig24n}
\end{figure}

Consequently, we assume that $I^b_A, I^f_A$ are fixed values in $N(p)$, where $p$ is a contour pixel.
Then, the values of $I^b_B, I^f_B$ satisfying the equation \ref{eq1ref} correspond respectively to
$I^{b*}_B=(I^f_A-\varepsilon-I^b_A)/2$, and any value of $I^f_B$ greater than $I^b_B+\delta l$, where $\varepsilon$
is a smallest intensity ($\varepsilon=1$) and $\delta l$ is the smallest difference of intensity
between two regions (see figure \ref{fig24n}).
Then any $N(q)$ with these values of $I^b_B, I^f_B$ verifies the equation \ref{eq1ref} and the pixel
$q$ will be classified correctly using $N(p)$.

\begin{figure}[ht]
\centering
  \includegraphics[width=8 cm]{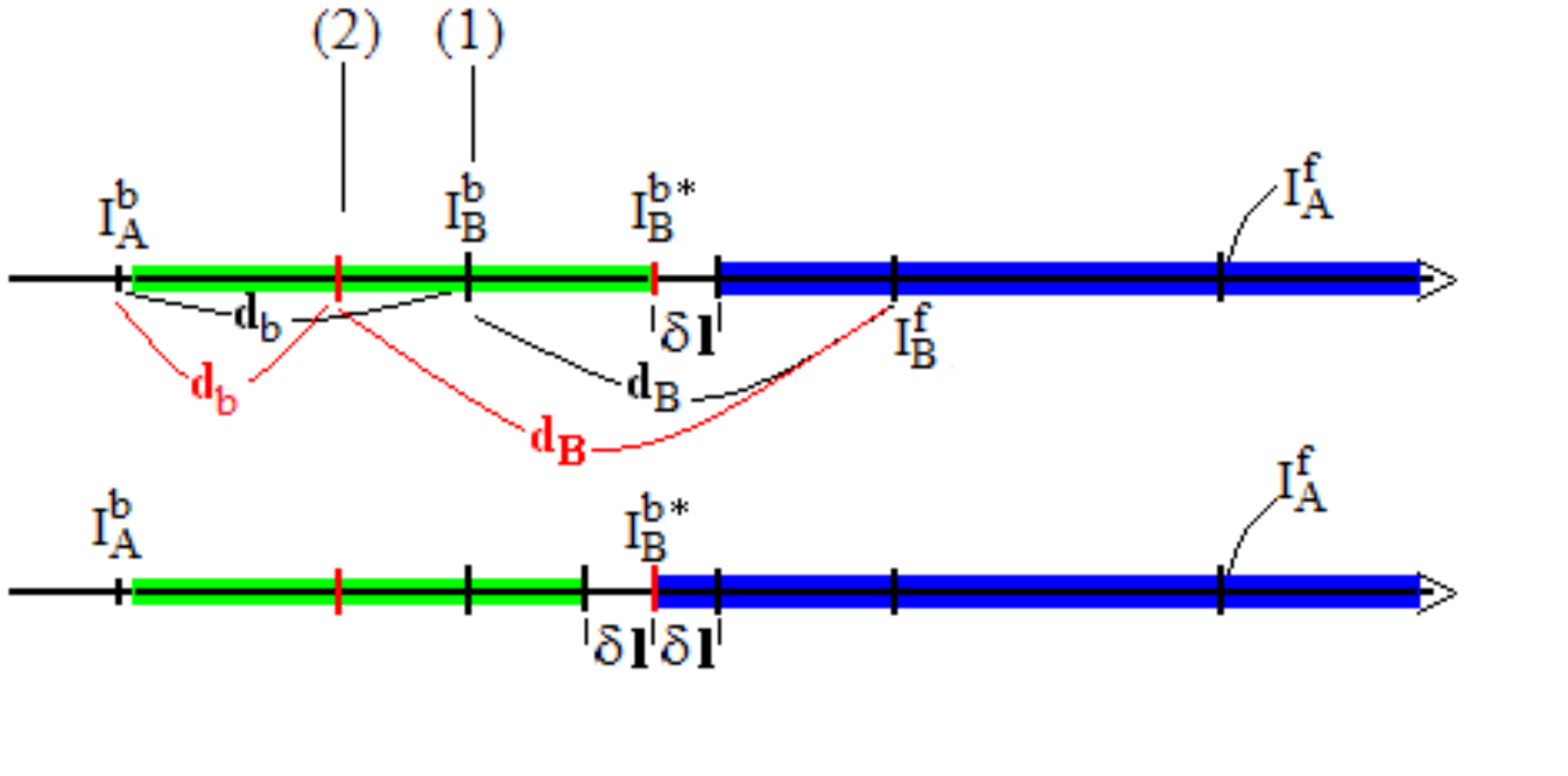}
  \caption{For a given $I^f_A$, possible values of $I^b_B, I^f_B$ illustrated respectively in green and
  blue colors. If the value of $I^b_B$ moves from the position $(1)$ to the position $(2)$, the value of
  $d_b+d_B$ is constant for a fixed value of $I^f_B$. If $I^b_B$ is distant from $I^{b*}_B$ by $\delta l$,
  the values of $I^f_B$ may begin from $I^{b*}_B$.}
  \label{fig25n}
\end{figure}

In addition, all $(I^b_B, I^f_B)$ of $N(q)$ so as $I^b_A < I^b_B \leq I^{b*}_B$ and $I^f_B \geq I^{b*}_B+ \delta l$
verify the equation \ref{eq1ref}. Indeed, if $I^b_B$ decreases towards $I^b_A$, the value of $d_b$ decreases
but $d_B$ increases with the same amount and then $2d_b+2d_B$ becomes constant. In the other hand, if
$I^b_B=I^{b*}_B-\delta l$, all $I^f_B \geq I^{b*}_B$ are possible because $I^f_A$ will belong always to the
interval given in the equation \ref{eq1ref} (see figure \ref{fig25n}).\\

It is necessary to have all combinations $(I^b_B, I^f_B)$ in order to classify correctly all $q$ outline
pixels. To do this, we will use the following results:\\

\begin{figure}[ht]
\centering
  \includegraphics[width=8 cm]{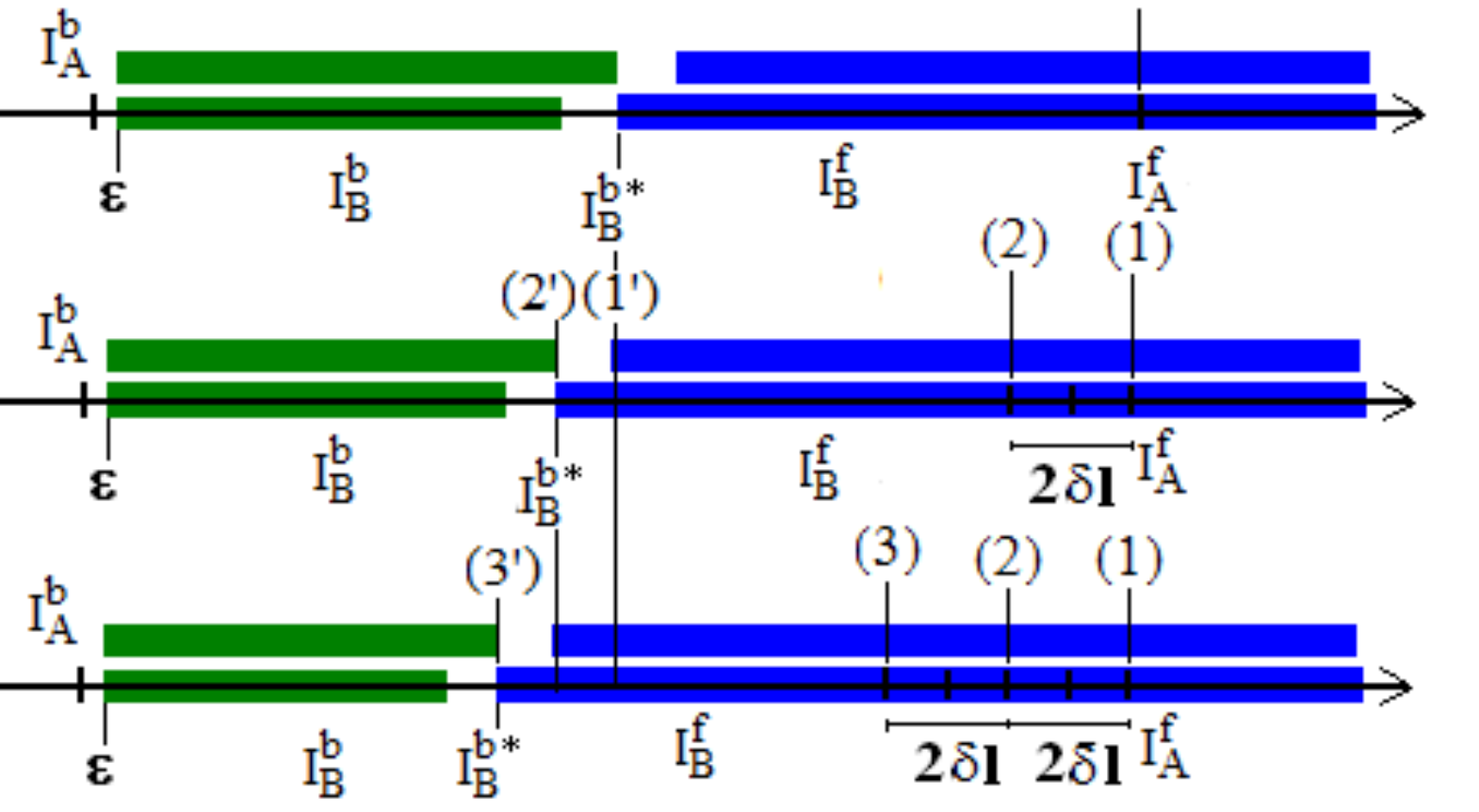}
  \caption{When the value of $I^f_A$ is decreased by $2\delta l$, and translated from the position (1)
  to (2), the value of $I^{b*}_B$ is decreased by $\delta l$ and translated from (1') to (2'). Possible
  values of $(I^b_B, I^f_B)$ are illustrated respectively by the green and blue colors.The same remark
  is valid when $I^f_A$ is decreased again by $2\delta l$ from (2) to (3).}
  \label{fig26n}
\end{figure}

- if we decrease $I^f_A$ by $2\delta l$, the value of $I^{b*}_B$ is decreased by $\delta l$ (see figure
\ref{fig26n}) and thus we get new possible values of $(I^b_B, I^f_B)$.

\begin{figure}[ht]
\centering
  \includegraphics[width=8 cm]{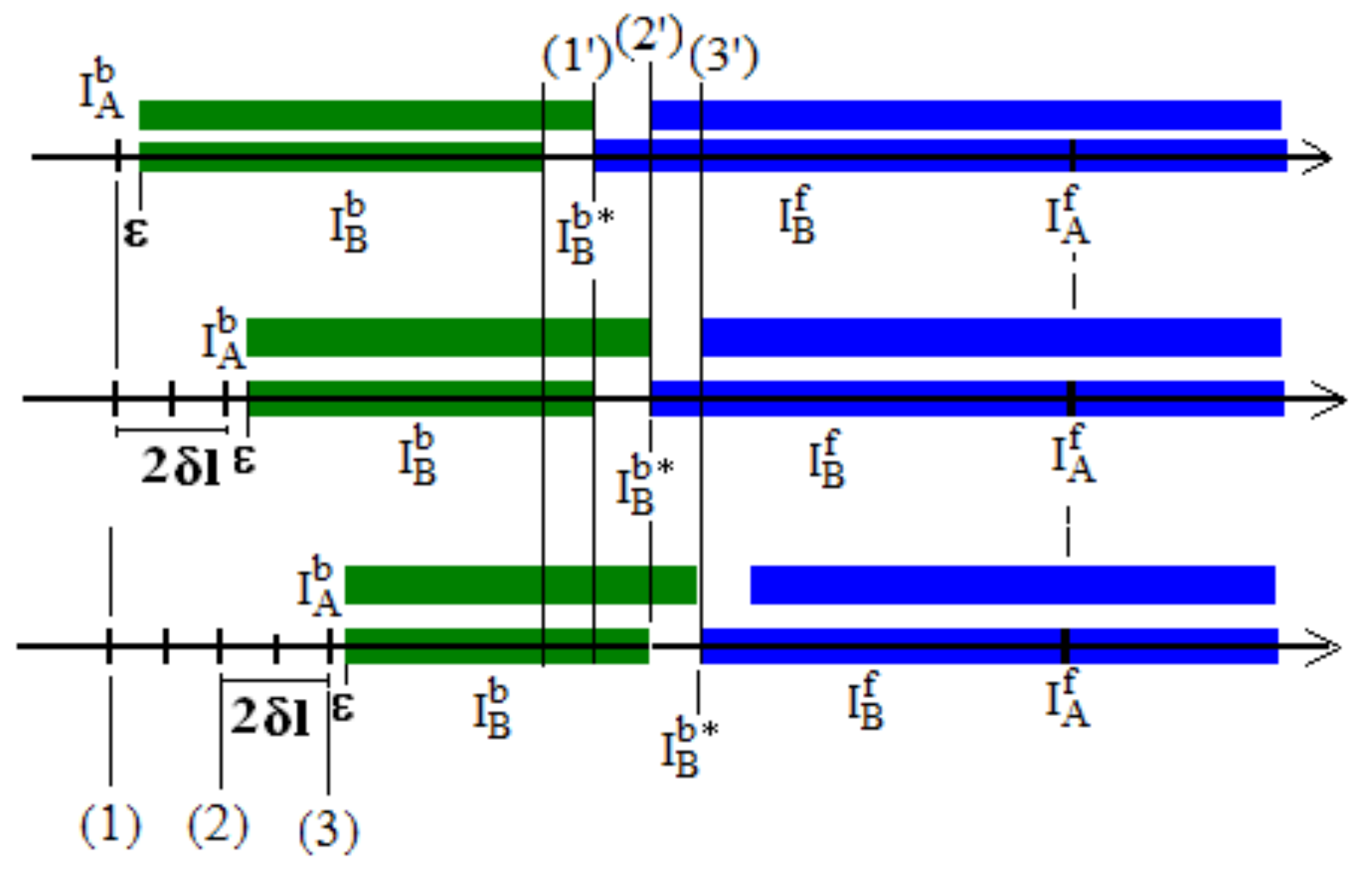}
  \caption{When the value of $I^b_A$ is increased by $2\delta l$, and translated from the position (1) to (2),
  the value of $I^{b*}_B$ is increased by $\delta l$ and translated from (1') to (2'). Possible values of
  $(I^b_B, I^f_B)$ are illustrated respectively by the green and blue colors.The same remark is valid when
  $I^b_A$ is increased again by $2\delta l$ from (2) to (3).}
  \label{fig27n}
\end{figure}

- if we increase $I^b_A$ by $2\delta l$, the value of $I^{b*}_B$ is increased by $\delta l$ (see figure
\ref{fig27n}) and thus we get new possible values of $(I^b_B, I^f_B)$.

Using these three results, we can get all combinations of $(I^b_B, I^f_B)$ performing the following steps:

(1)- Set $(I^b_A=0)$ and $(I^f_A=2\delta l)$, this implies that $I^{b*}_B=\delta l$. \\
(2)- We increase the value of $I^f_A$ with a step of $2\delta l$, this implies that $I^{b*}_B$
increases with $\delta l$. \\
(3)- We repeat the step (2) until that $I^f_A$ reaches the high value of intensity $224$, then $I^{b*}_B$
reaches the value $112$. $I^f_A$ can't reach the value $255$ because we can't have in this case $I^f_B>I^f_A$.\\
(4)- The rest of $I^b_B$ values are obtained by moving $I^b_A$ and $I^f_A$.\\

\begin{figure}[ht]
\centering
  \includegraphics[width=8 cm]{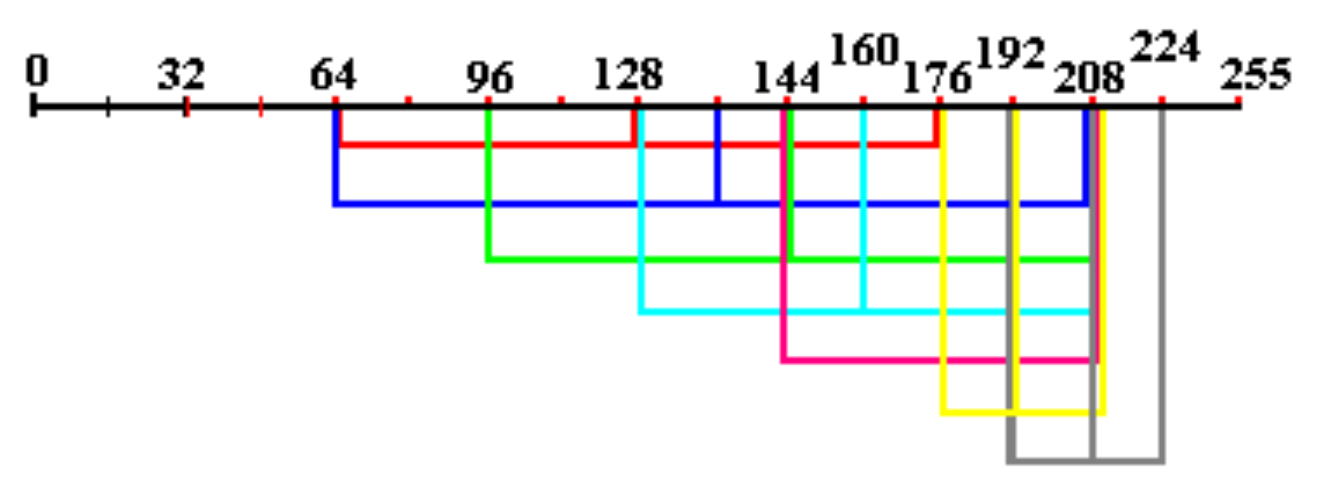}
  \caption{Used values of $I^b_A, I^f_A$ for obtaining all $I^b_B,I^f_B$ values }
  \label{fig28n}
\end{figure}

If we take $\delta l=16$, the set of $(I^b_A, I^f_A)are:\\
(0,32), (0,64), (0,96), (0,128), (0,160), (0,192), (0,224),\\
(64, 192), (64, 224), (96, 224), (128, 224), (160, 224)$,\\
$(192, 224), (208, 240)$ which correspond to the following values of $I^b_B$:\\
$16, 32, 48, 64, 80, 96, 112, 128, 144, 160, 176, 192, 208, 224$ (see figure \ref{fig28n}).

\begin{figure}[ht]
\centering
  \includegraphics[width=8cm]{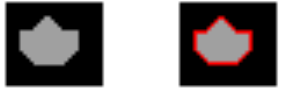}
  \caption{Sample of used pair of patterns $(A, A')$ where four directions are considered }
  \label{fig29n}
\end{figure}

What we propose in this paper is the use of artificial patterns $P_{1,i}$ instead of real images. The
key idea is to generate the image $A$ so as the background is set to zero $(I^b_A)$ and the foreground
is a shape having intensity $I^f_A$ and representing the four main directions of the contours $(d_i, i=0,3)$.
This allows any $q$ of $N(q)$ so as $I^f_B>I^b_B=(I^f_A-I^b_A)/2$  to be classified correctly
(see figure \ref{fig29n}). The pattern $I^f_A$, identical to $A$ but in addition contour pixels are marked.

We can now give the following result based on the previous reasoning:

\textbf{Let $(A, A')$ be an artificial pattern having $I^b_A$ and $I^f_A$ as intensities of the
background and foreground. Let $N(q)$ be the neighborhood of the query pixel $q$ so as $I^b_B$,
$I^f_B$ intensities of the two regions of $N(q)$ so as $(I^f_B>I^b_B)$ and $d_b=(I^b_B-I^b_A)>0$. The pixel
$q$ will be classified correctly if $2d_b+2d_B>I^f_A>2d_b$.
}

\begin{figure}[ht]
\centering
  \includegraphics[width=12 cm]{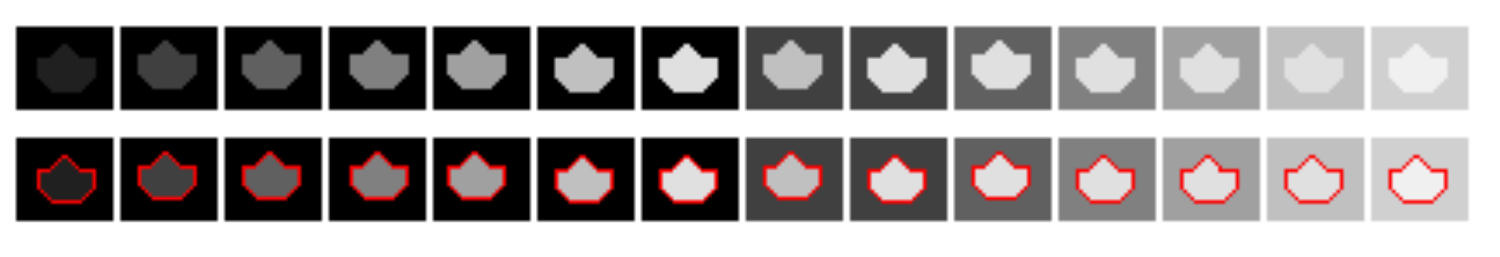}
  \caption{The set of patterns $P_{1,i}$ in case of $d_b>0$}
  \label{fig30n}
\end{figure}

The set of generated pairs of patterns $P_{1,i}$ as illustrated by figure \ref{fig30n}.

\textbf{Case 2}

Our goal is to have the located outline pixels verifying:$d_b<0$ and $d_A \in [2d_b+2d_B, 2d_b]$.

Following the same reasoning as case $1$, we get the following result:

\textbf{Let $(A, A')$ be an artificial pattern having $I^b_A, I^f_A$ as intensities of the background
and foreground.
Let $N(q)$ be the neighborhood of the query pixel $q$ so as $I^b_B, I^f_B$ intensities of the two
regions of  $N(q)$ verifying $(I^f_B<I^b_B)$ and $d_b<0$. The pixel $q$ will be classified correctly
if $2d_b+2d_B<I^f_A<2d_b$.}

If we take $\delta l=16$, the set values of $(I^b_A,I^f_A)$ are:\\
 $(255,224), (255,192), (255,160), (255,128), (255,96), \\
(255,64), (255,32)$ giving the values of $I^b_B$ equal to $240, 224, 208, 192, 176, 160, 144$.\\
The value $0$ for $I^f_A$ is excluded because this implies that $I^b_B=128$ and then we can't have $I^f_B<I^f_A$.

To obtain other combinations, it is sufficient to take the following values for $(I^b_A, I^f_A)$:\\

$(224, 32)$, $(192, 32)$, $(160, 32)$, $(128, 32)$, $(96, 32)$, $(64, 32)$, $(48, 16)$ which
correspond to the following values of $I^b_B$: $128, 112, 96, 80, 64, 48, 32$.

\begin{figure}[ht]
\centering
  \includegraphics[width=10 cm]{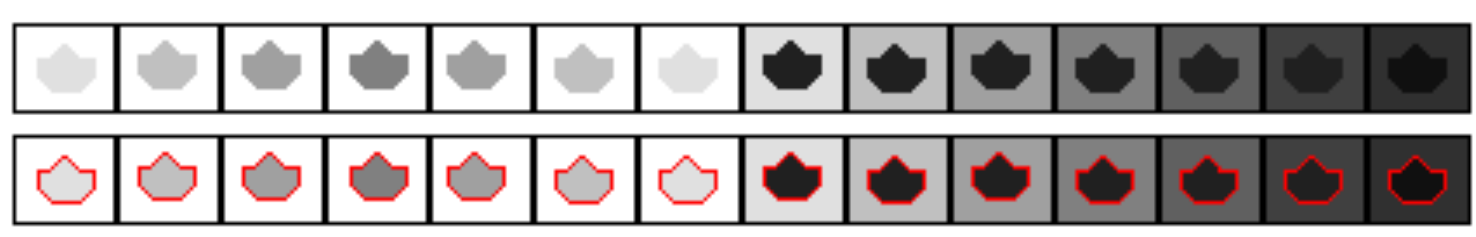}
  \caption{The set of patterns $P_{2,i}$ in case of $d_b<0$}
  \label{fig31n}
\end{figure}

The set of pairs of patterns $P_{2,i}$ as illustrated by figure \ref{fig31n}.

\subsection{Duality of the two sets of artificial patterns}

The two sets of artificial patterns proposed for contour detection ($P_{1,i}$ for the case $d_b>0$ and
$P_{2,i}$ for the case $d_b<0$) perform the same task, the unique difference is the position of the
computed contour which is outside of the shape for the first one and inside the shape for second one.

This duality is due to the fact that each case may be considered as the other case interchanging the
role of the considered regions: the darkest region is considered as background and the clearest one
as shape for the set of patterns and vice versa.

\begin{figure}[ht]
\centering
  \includegraphics[width=7 cm]{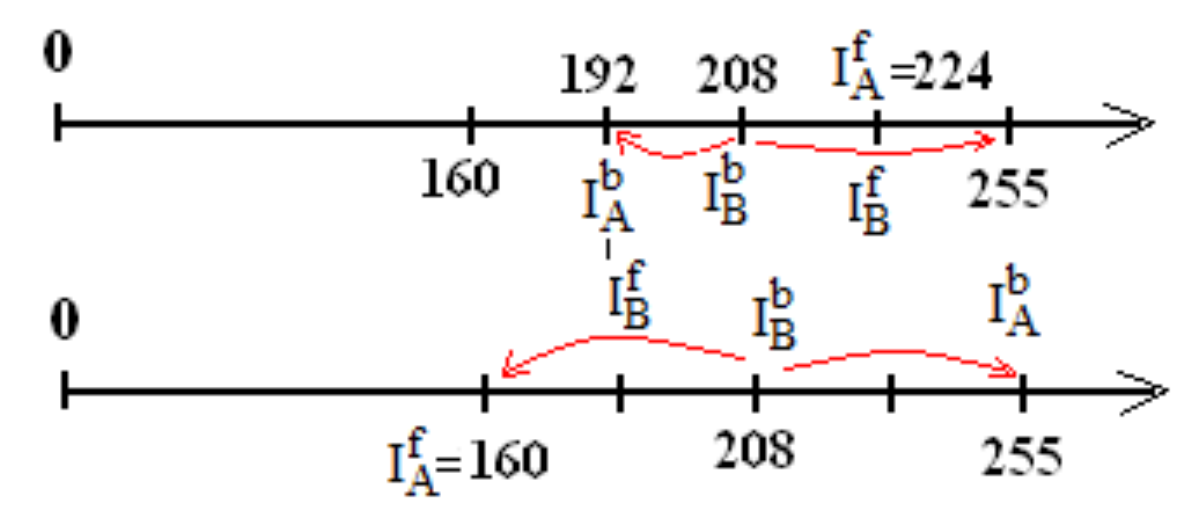}
  \caption{Example of the duality between patterns $P_{1,i}$ and $P_{2,j}$}
  \label{fig32n}
\end{figure}

Figure \ref{fig32n} illustrates this duality for example between $P_{1,13}$ and $P_{2,3}$. For the pattern
$P_{1,13}$, the possible values of $I^f_B$ are from $208$ to $255$ and for $I^b_B$ are from $208$ to $192$.
Concerning the pattern $P_{2,3}$, the possible values of $I^f_B$ are from $208$ to $255$ and for $I^b_B$ are from
$208$ to $160$. The second is then equivalent to the first one if we interchange between $I^b_B$ and $I^f_B$.

This relation is verified for the following pairs of patterns:\\
$(P_{1,1}, P_{2,1})$, $(P_{1,2}, P_{2,14})$, $(P_{1,3}, P_{2,13})$,\\
 $(P_{1,4}, P_{2,12})$, $(P_{1,5}, P_{2,11})$, $(P_{1,6}, P_{2,10})$,\\
 $(P_{1,7}, P_{2,9})$, $(P_{1,8}, P_{2,8})$, $(P_{1,9}, P_{2,7})$, \\
 $(P_{1,10}, P_{2,6})$, $(P_{1,11}, P_{2,5})$, $(P_{1,12}, P_{2,4})$,\\
  $(P_{1,13}, P_{2,3})$, $(P_{1,14}, P_{2,2})$.\\

\subsection{Obtaining of most significant outline}\label{LevelsOfContours}

\begin{figure}[ht]
\centering
  \includegraphics[width=8 cm]{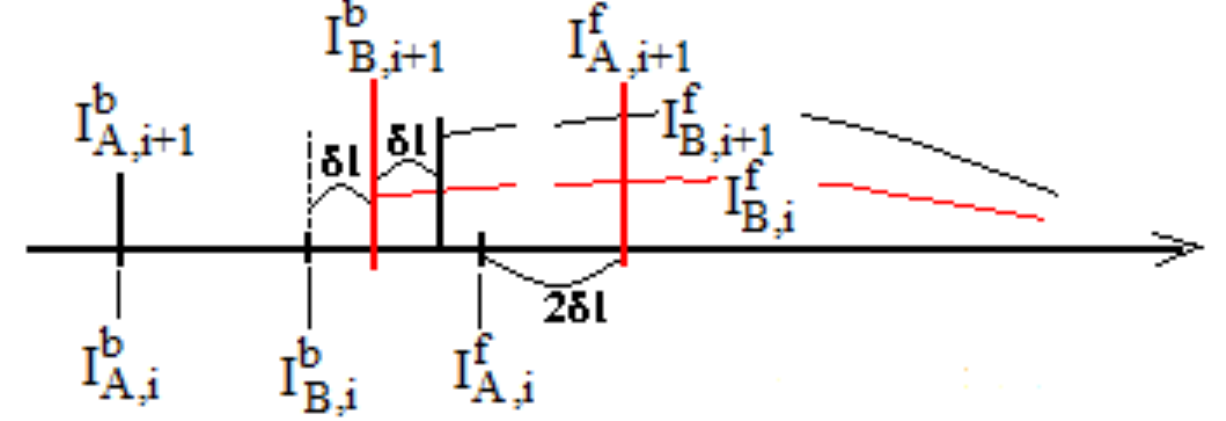}
  \caption{Evolution of the contour using successive patterns $P_{1,i}$ and $P_{1,i+1}$, here we took
  $I^b_{A,i}=I^b_{A,i+1}$}
  \label{fig33n}
\end{figure}

When we visualize the located contour using two successive patterns $P_{1,i}$ and $P_{1,i+1}$ we
find that the outline is either growing or shrinking. This motion of the outline is explained by the
illustration in Figure \ref{fig33n}. If we assume that outline are located for the pattern $P_{1,i}$ with
$(I^b_{A,i}, I^f_{A,i})$, then all contour pixels which verify $I^f_B>I^b_{B,i}$ will be located where $I^b_{B,i}=(I^b_{A,i}+I^f_{A,i})/2$.

\begin{figure}[ht]
\centering
  \includegraphics[width=6 cm]{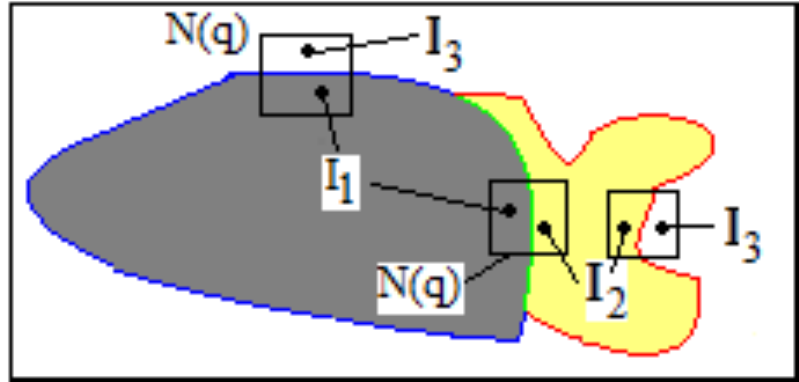}
  \caption{Example of the evolution of the contour: First the use of the pattern $P_{1,i}$ allows locating the outline in blue and green color because the intensities $I_1,I_2$ of $N(q)$ with (grey, White) colors and $I_1,I_3$ of $N(q)$ with (grey, yellow) colors verify the equation \ref{eq1ref} where the $N(p)$ is taken from the defined $P_{1,i}$. The outline in red color is located only by the next pattern $P_{1,i+1}$ because the intensities $I_2,I_3$ of $N(q)$ with (yellow, white) color verify the equation \ref{eq1ref} where the $N(p)$ is taken from the defined $P_{1,i+1}$.}
  \label{fig34n}
\end{figure}

If the pattern $P_{1,i+1}$ is used, then $(I^b_{A,i+1}, I^f_{A,i+1})$ replaces $(I^b_{A,i}, I^f_{A,i})$
and $I^b_{B,i+1}$ instead of $I^b_{B,i}$ (see figures \ref{fig33n}, \ref{fig34n}).

For the second pattern $P_{1,i+1}$, all pixels verifying $I^f_B>I^b_{B,i+1}$ $(I^b_{B,i} < I^b_{B,i+1})$
will be located. Consequently, the pixels for which the value of $I^f_B$ is between $I^b_{B,i+1}$ and $I^b_{B,i}$
will not appear in the new located outlines. These pixels correspond to low variation of intensity and are
considered as high frequency information. Figure \ref{fig34n} illustrates an example where the green and
blue colored contour (rightmost region in the segmentation) are located using $P_{1,i}$ pattern. However,
the green outline does not appear when the pattern $P_{1,i+1}$ is used. Consequently, only the red and
blue contour will be located.

The result of application of all patterns to the query image produces a set of outlines that are moving
showing the propagation of high frequency outlines. We will associate a level of outlines the number of
times where it appear applying a successive patterns $P_{1,i}$. For example, if we apply two successive
patterns $P^1_i$ and $P^1_{i+1}$, outlines that appears only with $P_{1,i}$ are of level $1$. Those which
appear with both $P_{1,i}$ and $P_{1,i+1}$ are of level $2$ and so on.

In order to decrease the interval of recovering $[I^b_{B,i}, I^b_{B,i+1}]$, we will use $\delta l=8$
instead of $\delta l=16$. This allows to obtain more levels of contours. In this case, the number of
patterns $P_{1,i}$ is $28$ instead of $14$ and the values of $I^b_A, I^f_A$ become:

$(0,16)$, $(0,32)$, $(0,48)$, $(0,64)$, $(0,80)$, $(0,96)$, $(0,112)$,\\
$(0,128)$, $(0,144)$, $(0,160)$, $(0,176)$, $(0,192)$, $(0,208)$, $(0,224)$,\\
$(0,240)$, $(64,192)$, $(64,208)$, $(64,224)$, $(64,240)$, $(96,224)$, $(96,240)$,\\
$(128,224)$, $(128,240)$, $(160,224)$, $(160,240)$, $(192,224)$,\\
$(192,240)$, $(208,240)$.

The corresponding values of $I^b_B$ are therefore:\\
$8, 16, 24, 32, 40, 48, 56, 64, 72, 80, 88, 96, 104, 112, 120, 128,\\
136, 144, 152, 160, 168, 176, 184, 192, 200, 208, 216, 224$.


\section{Algorithm, Complexity and Details of the Implementation}\label{Algorithm}

The proposed method uses the $14$ artificial  artificial patterns represented by pairs of images illustrated
by figure \ref{fig30n}.
For a query image $B$ of size $(N \times M)$, and for a size of the neighbors $N(q)$ equals to $3$, each $N(q)$
associated to a pixel $q$, similarity measures $S(q,p)$ is computed for all $N(p)$ of the image $A$.

Let $(N' \times M')$ be the size of the image $A$. As $S(q,p)$ is computed $N'M'$ times for each $q$,
the number of times of the computation of $S(q,p)$ for all pixels $q$ is equal to $(N \times M \times N' \times M')$.
The number of operations required for the computation of the similarity measure is equal to: $(m \times m)$ substractions,
$(m \times m)$ multiplications, and $(m \times m)-1$ additions, assuming that $m \times m$ is the size of $N(q)$
(see equation \ref{eq1}).

Then for each pattern, to compute the pixels contours we need to perform:
$(N \times M \times N' \times M') (3m \times m)$ arithmetic operations.

\begin{figure}[ht]
\centering
  \includegraphics[width=6 cm]{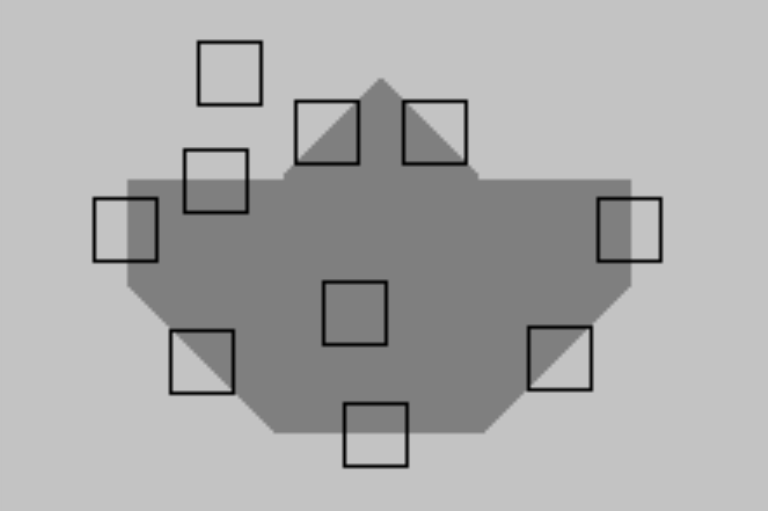}
  \caption{The considered $N(p)$ in the training image $A$}
  \label{fig88n}
\end{figure}

To reduce this complexity, we improved the computation of the similarity measure which instead to concern
all the image $A$, we used only some neighbors $N(p)$ representing all the information of the image.
Indeed, there is many time computation without any profit for the computation of $S(q,p)$ such as
when $N(p)$ concern only the background or the foreground of $A$ (see figure \ref{fig88n}).
To avoid this loss of time, we considered only all possibilities of $N(p)$ appertaining to the border
of the shape inside $A$ and to  the background or the foreground.

\begin{figure}[ht]
\centering
  \includegraphics[width=6 cm]{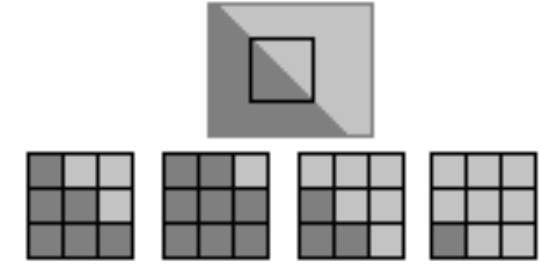}
  \caption{Example of considered $N(p)$ and the different configurations taken into account for such neighbor}
  \label{fig89n}
\end{figure}

There are 8 considered Neighbors $N(p)$ considered as illustrated by figure \ref{fig88n}. For each one
containing a border, four configurations are considered giving all possibilities of the border in $N(p)$
as illustrated by figure \ref{fig89n}.

With this new improvement, the number of used neighbors $N(p)$ in $A$ is then equal to $24+2$
(2 for the foreground and background).

The number of arithmetic operations using the $14$ patterns is then equal to:
$(14 \times N \times M \times 26\times (3m \times m))$ giving a complexity of $O(n^2)$.

If we consider for example, one image of BSD500 dataset having $(481 \times 321)$ rows and columns,
the number of operations is equal to:
$(14 \times 481 \times 321 \times 26 \times 27)=1517453028$ operations.

Most microprocessors today can do $4$ FLOPs per clock cycle. Therefore, a single-core $2.5$ GHz
processor has a theoretical performance of $10$ billion FLOPS = $10$ GFLOPS, this gives as computation
time equal to $0,151 sec$.

\begin{tabbing}
\hspace{1cm}\=\hspace{1cm}\=\kill
\textbf{The Algorithm}\\
\textbf{Begin}\\
-$(A,A')$ is the pair of artificial training pattern\\
-$B$ is the query image\\
-$B'$ is the computed image, identical to $B$, in addition contour pixels will be marked\\
-The $26$ reference $N(p)$ in $A$ are used instead of all $N(p)$ of $A$\\

\textbf{For} each $q$ from $B$\\
\textbf{Do} \textbf{For} each reference $N(p)$ of $A$\\
    \>  \textbf{Do}  Compute the similarity measure $S(q,p)$\\
    \> \textbf{EndFor}\\
    \> Select $p^*$ so as $S(q,p^*)$ is minimal\\
    \>         \textbf{If}($p^*$ in $A'$ is a contour pixel)\\
     \>        \textbf{Then} $q$ is set in $B'$ as contour pixel\\
     \>         \textbf{Else} $q$ is set in $B'$ as non contour pixel\\
\textbf{EndFor}\\
\textbf{End.}
\end{tabbing}

\section{Results} \label{Results}

The first part of this section is devoted to the parameter setting and to results obtained using
images of hand-drawn contours as training images. In particular we discuss the limitations
of this approach.
We show how the detected contour varies through the image depending on the specific
variation of luminosity chosen, resulting in a contour "level". The problem of the choice
of the suitable patterns in order to find the expected contour has been studied in
\ref{LevelsOfContours} and the result of computation of contours of different levels
is presented.
We applied our method to the problem of detecting people in video sequences
(a focus of CAVIAR data set). We show that the detection is easily achieved
due to the quality of contours acquired by our method.

We present in the next, a study related to the invariance of located contours to scale change and rotation.
Finally, we present a qualitative and quantitative evaluation of our method on different data sets
of real images: Berkeley Segmentation
Data Set $(BSDS500)$ \cite{Arbelaez et al 2011}, Weizmann Horses \cite{Borenstein and Ullman 2002}.
The BSD 500 consists of 500 natural images, with hand drawn contours by five different subjects. Weizmann
Horses data set consists of 328 images of horses manually segmented. Only the outlines of horses are drawn.
The obtained results are compared with the state of the art methods.
Finally, we studied the invariance of the proposed method to scale change and rotation.

\subsection{New benchmarks for BSD 500 dataset}

The definition of contour such is given by is:

Since human judgment is the only possible criterion that can be
used in order to say if a given visual feature is a contour or not, we
operationally define contours in a given image as the set of lines that
human observers would concent on to be the contours in that image (one
could give a similar operational definition of other concepts used in
the image processing and visual pattern recognition literature, such as
face). On the light of this, research in contour detection aims at
understanding and modeling mathematically the features which
people (consciously or consciously) use to recognize such line sets
(such as contrast, good continuation, and closure).

\subsection{Parameter settings}

There are three parameters whose values are justified: the size of neighborhoods $N(p)$ and $N(q)$,
the number of directions of contours in training images (artificial patterns) and the value of $\delta l$.

- Size of neighborhoods $N(p)$ and $N(q)$: In \ref{coherence}, we have seen that
the optimal match $p$ of a query pixel $q$ will be obtained so as the directions of contour pixels
in both $N(p)$ and $N(q)$ are the same.

\begin{figure}[ht!]
\centering
\includegraphics[width=6 cm]{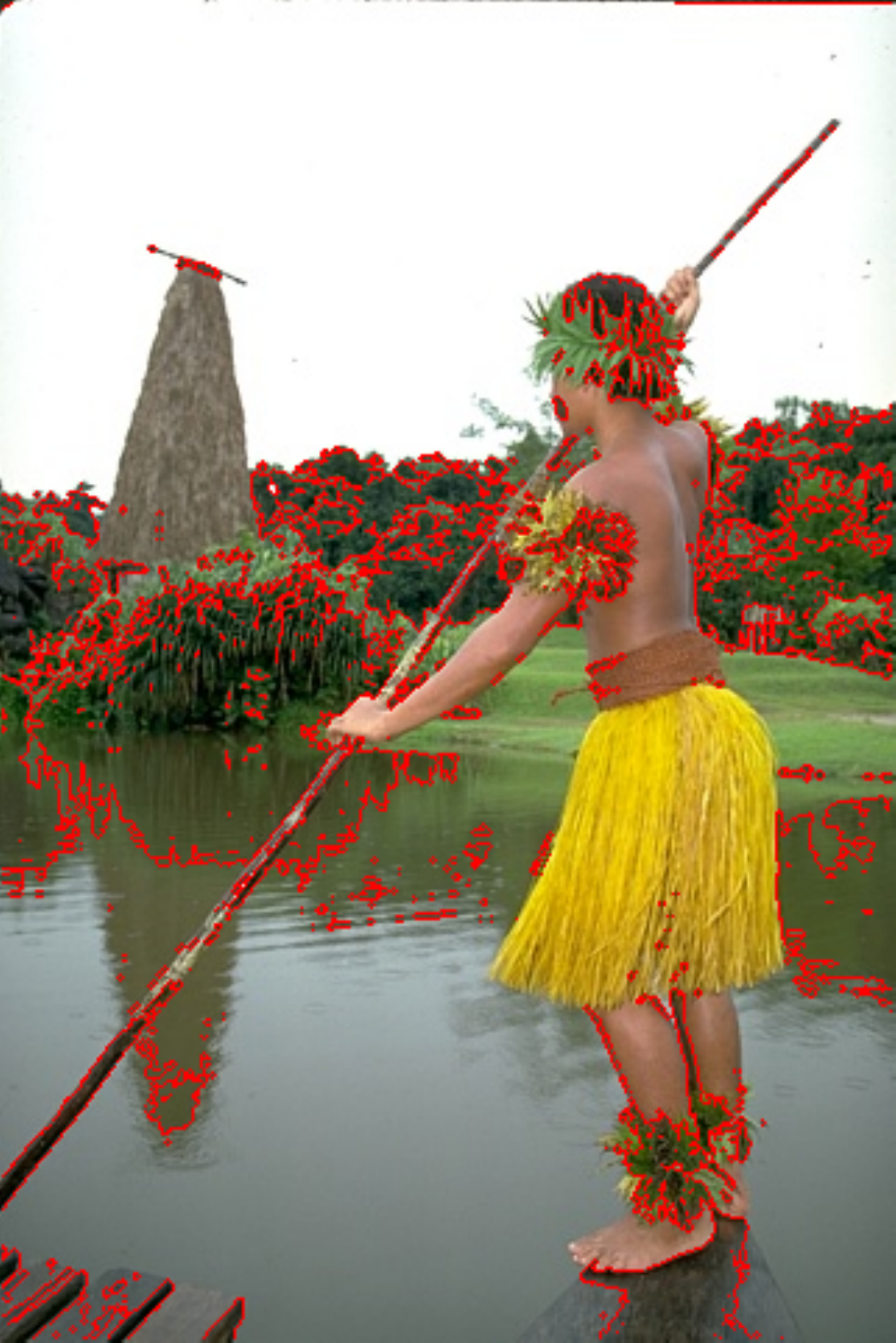}
\caption{Located contours using the pattern $P_5$ for sizes of neighborhood $N(p)$
  equal to $3$}\label{fig36n}
 \end{figure}

\begin{figure}[ht!]
\centering
\includegraphics[width=6 cm]{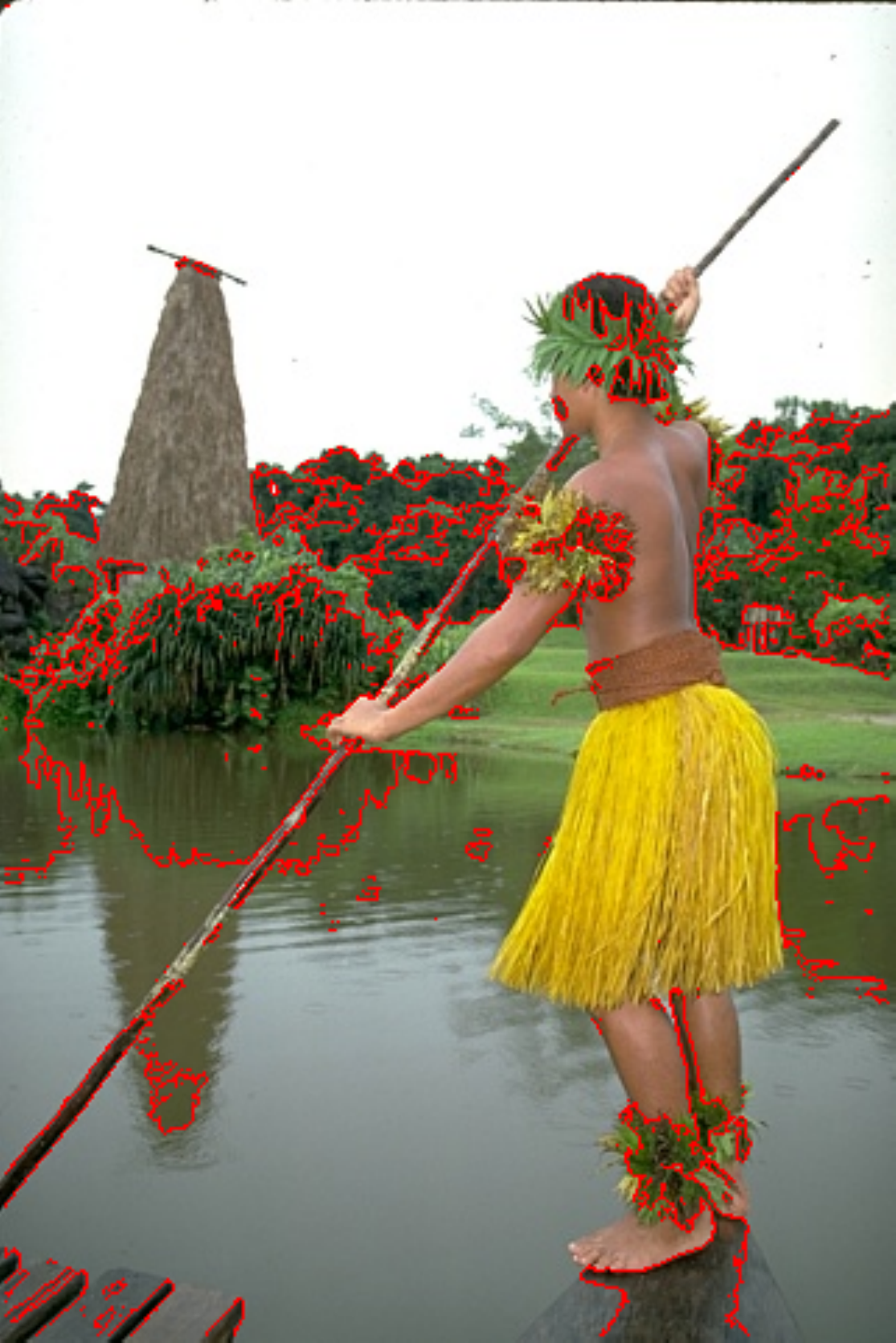}
\caption{Located contours using the pattern $P_5$ for sizes of neighborhood $N(p)$
  equal to $5$}\label{fig37n}
\end{figure}

\begin{figure}[ht!]
\centering
\includegraphics[width=6 cm]{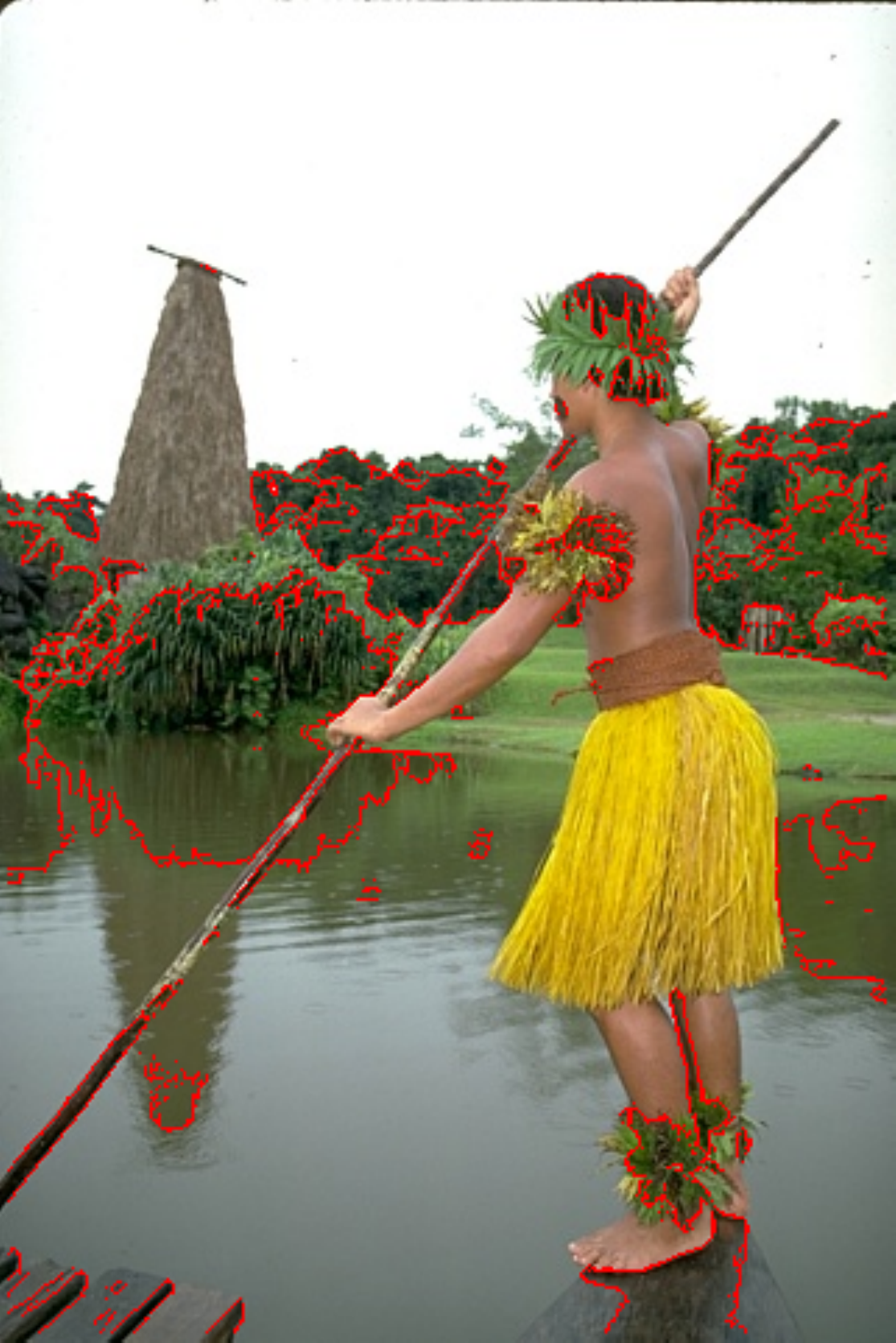}
\caption{Located contours using the pattern $P_5$ for sizes of neighborhood $N(p)$
  equal to $7$}\label{fig38n}
\end{figure}

When this  size of $N(p)$ and $N(q)$ is equal to $3$, more contours pixels will
be located because the three aligned contour pixels will more numerous than with size more greater
($5$ or $7$). Figures \ref{fig36n}, \ref{fig37n}, \ref{fig38n} illustrate the located contours using the
pattern $P_5$ for the value of size is equal to $3, 5$ and $7$. We can see that with $size=3$, contour
pixels encompass with more accuracy the regions and are more numerous than with other sizes. When the
size is equal to $5$ or $7$, in the query image, there are less possibilities to have aligned contour
pixels having the same direction such as $N(p)$. Indeed, despite the quality of the contours are the same, some
contours pixels will be not located in addition of the time of computation that will increase considerably.

\begin{figure}[ht!]
\centering
\includegraphics[width=6 cm]{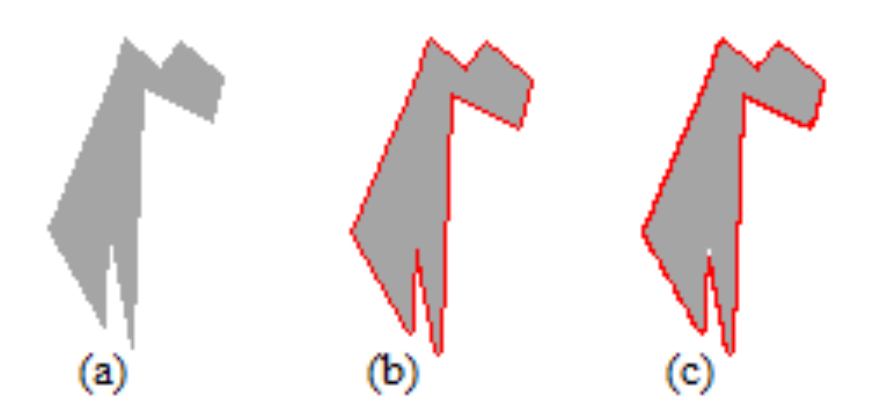}
\caption{(a) Initial image, (b) Located contours using the size of the
neighborhood $N(q)$ equals to $3$, (c) Located contours using the size of the
neighborhood $N(q)$ equals to $5$}\label{fig39n}
\end{figure}

- Number of directions: We considered in the different illustrative figures of the paper
four main directions. With size of the neighborhood $N(q)$ equal to $3$,
all directions are represented with the four considered directions and then it is sufficient to
locate any contour orientation. Figure \ref{fig39n} illustrates a
sample of shape whose boundaries have other directions than the four considered, but all contour
pixels are located using size of the neighborhood $N(q)$ equals to $3$ which looks with uniform
connexity ( a $8$-connected); while with a neighborhood size equals to 5 the contour are with two
different connexities (pixels are $4$ and $8$-connected).

- Value of $\delta l$: The set of artificial patterns are based on the value of $\delta l$: the $14$
patterns will be used if this value of $\delta l$ is equal to $16$ and $28$ patterns will be used if
value is equal to $8$ .

The values of these parameters are known and the computed contours does not depend from any other
parameter and are unique for each image.

\subsection{Using hand-drawn contours as training images}

To detect the outline of a query image, we require a pair of training images
$(A, A')$ of a scene where $A'$ is identical to $A$, in addition it contains hand-drawn
contours.

\begin{figure}[ht!]
\centering
\includegraphics[width=4 cm]{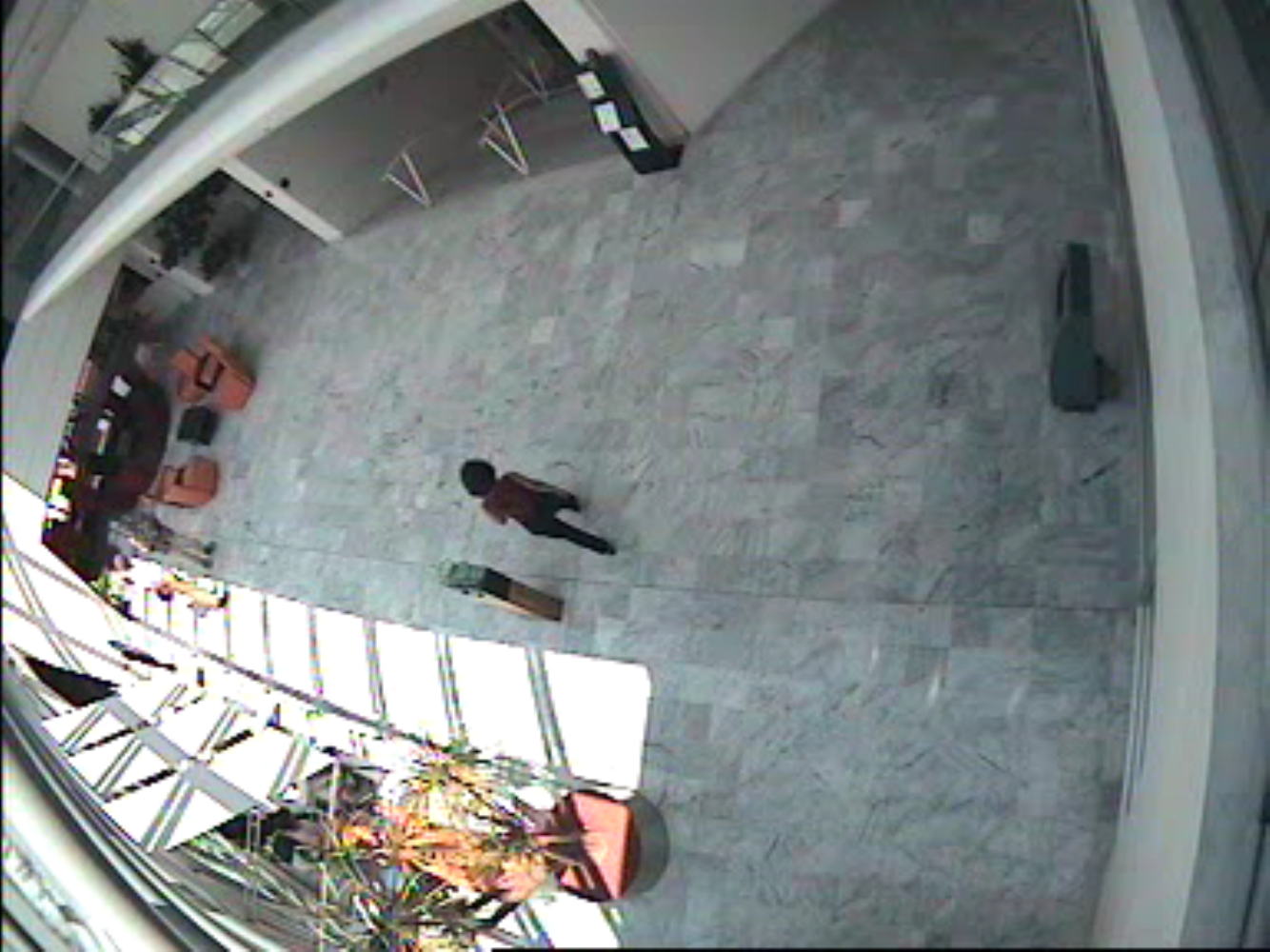}
\includegraphics[width=4 cm]{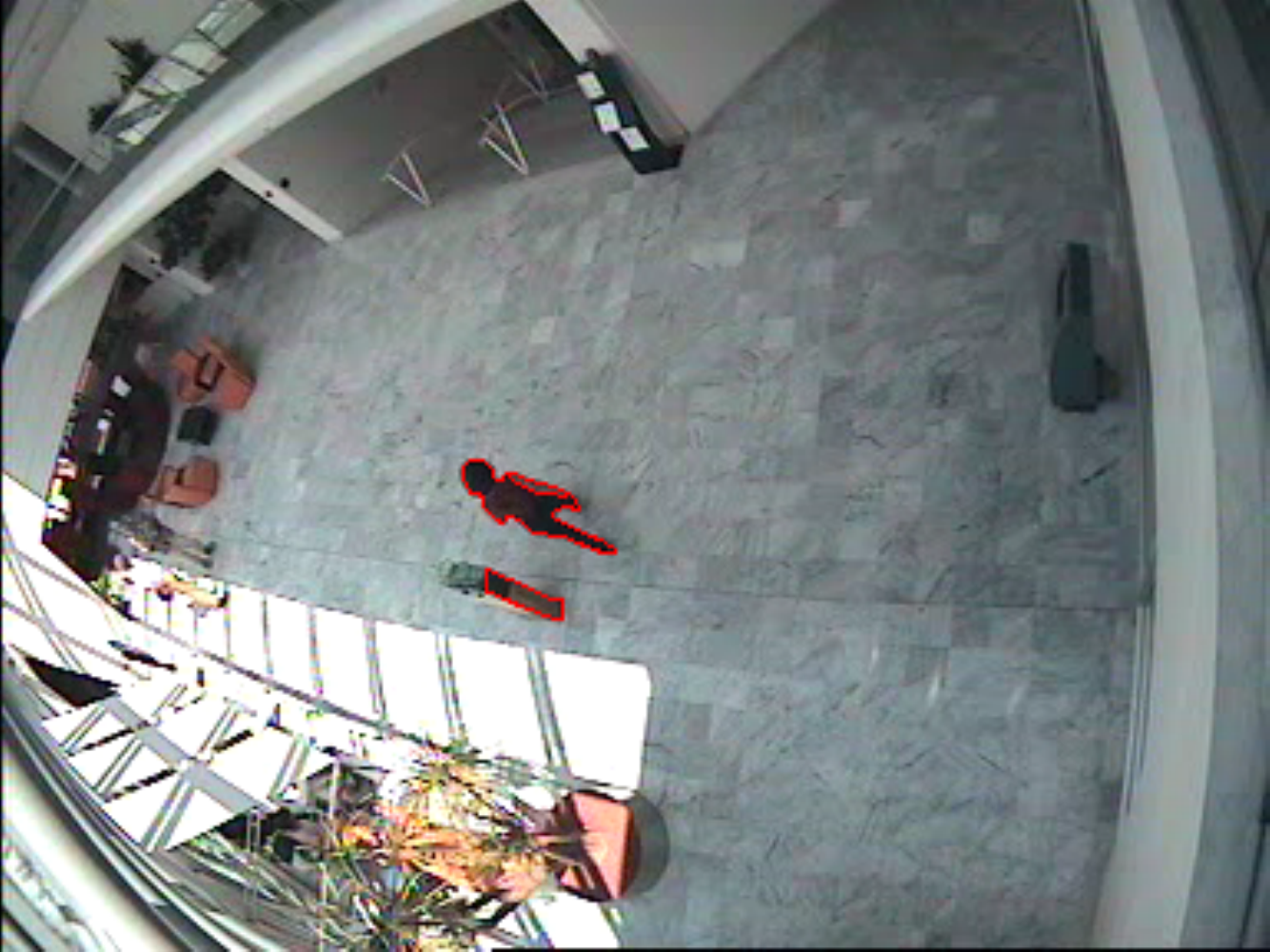}
\includegraphics[width=4 cm]{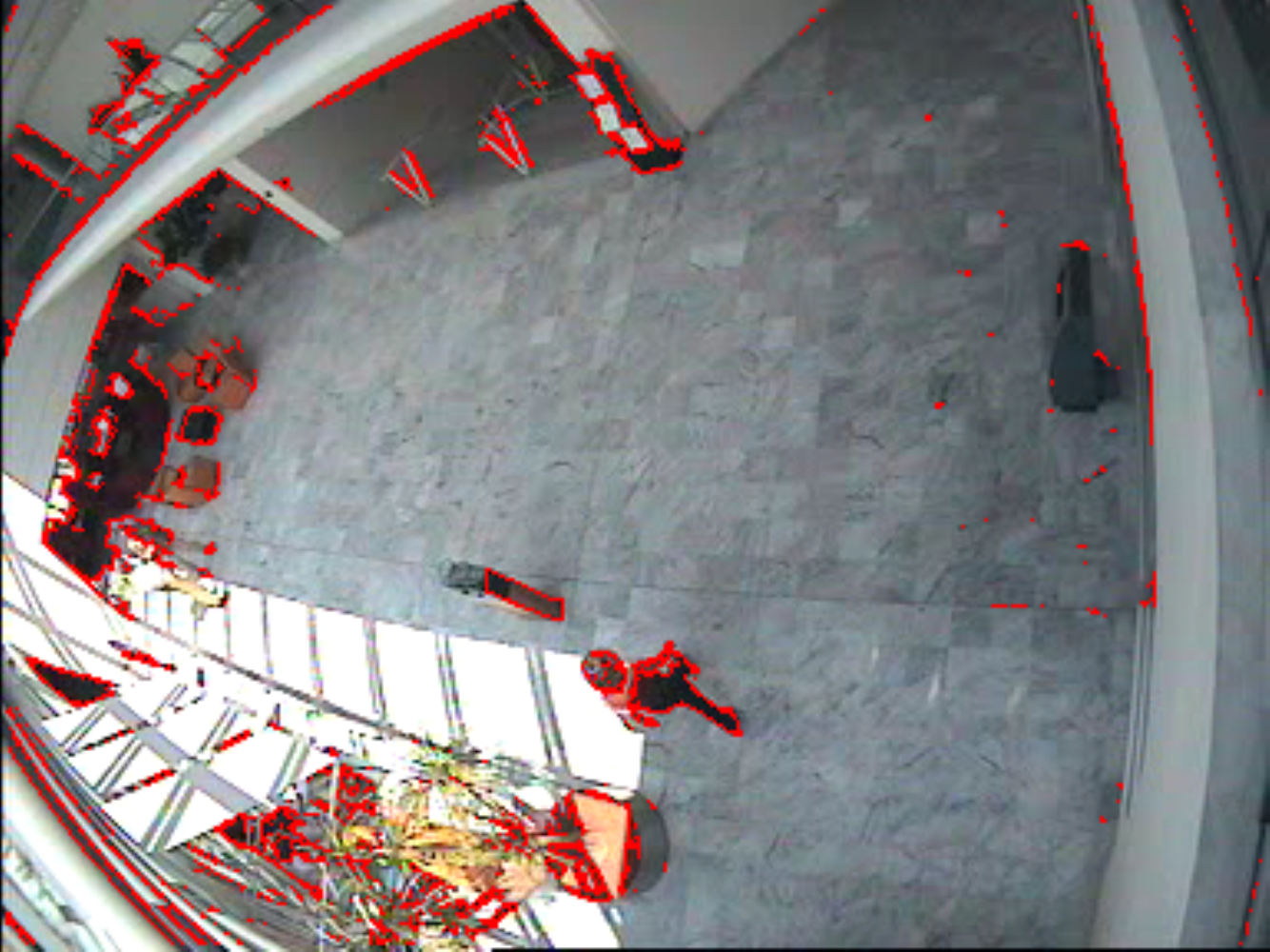}
\includegraphics[width=4 cm]{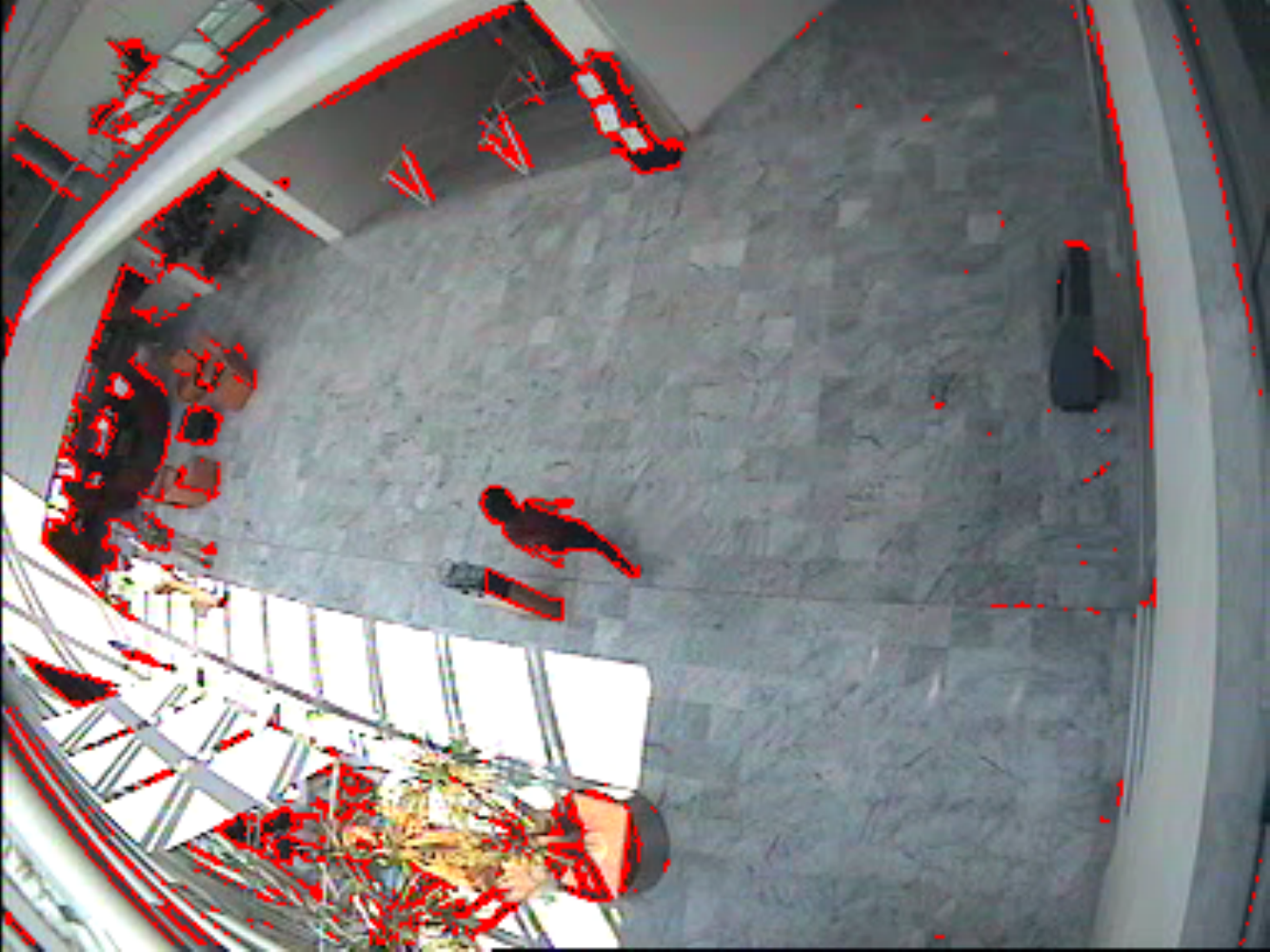}
\caption{From top to bottom, left to right: Initial image $A$, The synthesized image $A'$,
Located contours for a new frame $B$, Located contours for another frame $B$}\label{fig43n}
\end{figure}

Figure \ref{fig43n} illustrates images $A, A'$ from the CAVIAR data set (where
hand-drawn outline shapes are highlighted with red color). The automatically located contours
for some images of the same video are illustrated by the same figure \ref{fig43n}.
It can be seen that in the query images some outlines are located but many others aren't located.
This is because the neighboring to considered pixels do not verify the required constraints (see annex \ref{annex}).

\begin{figure}[ht!]
\centering
\includegraphics[width=4 cm]{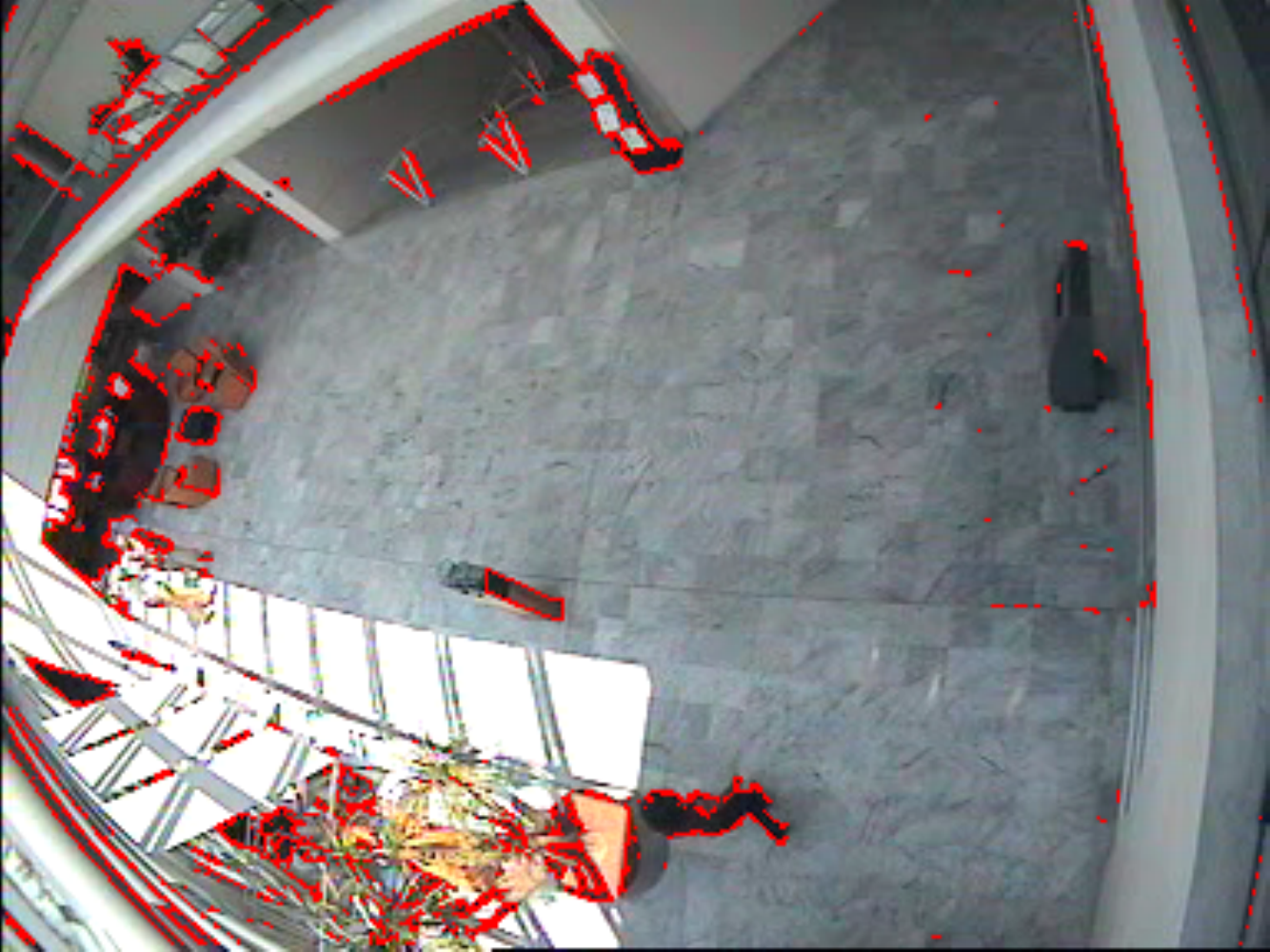}
\includegraphics[width=4 cm]{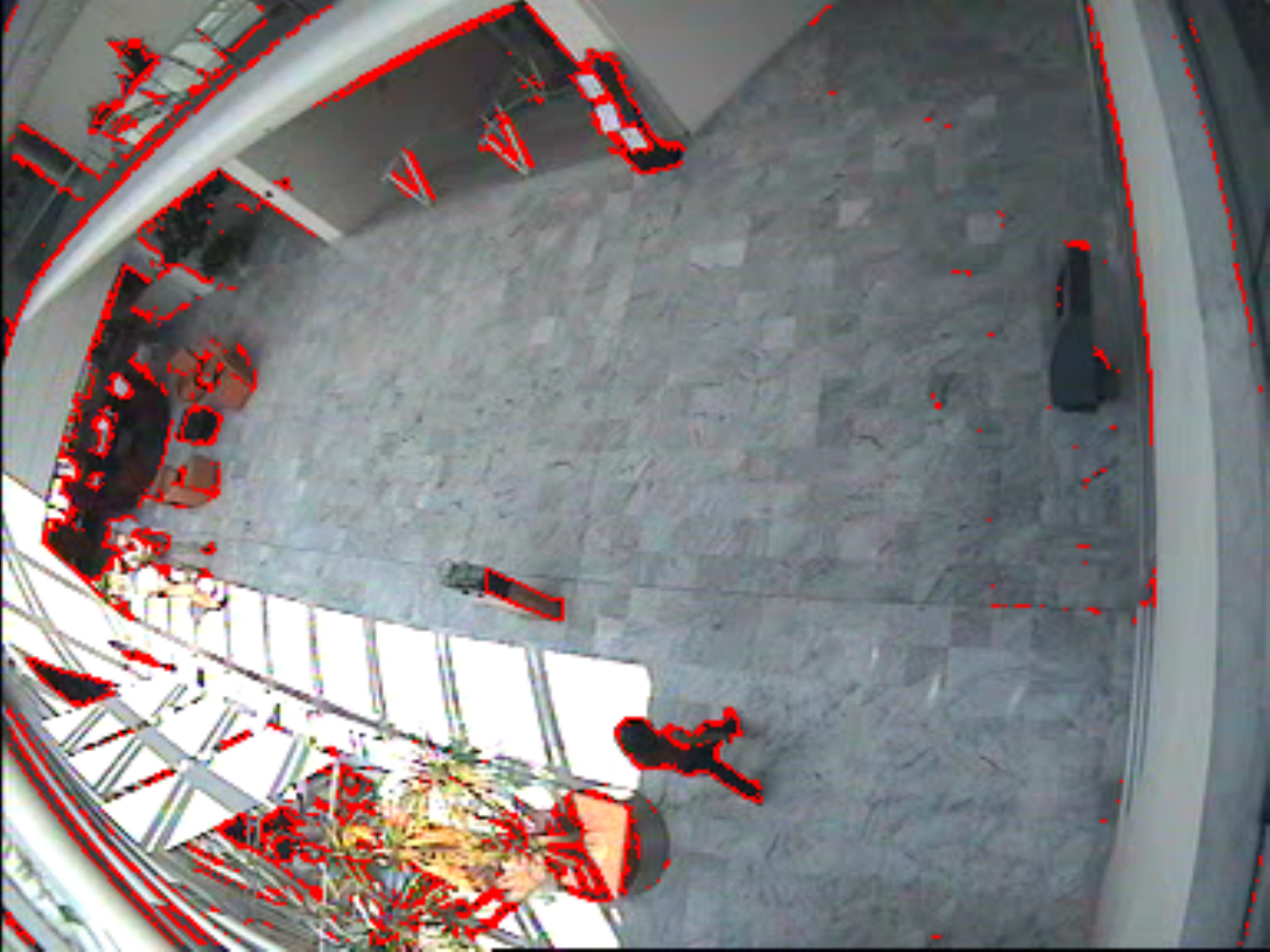}
\includegraphics[width=4 cm]{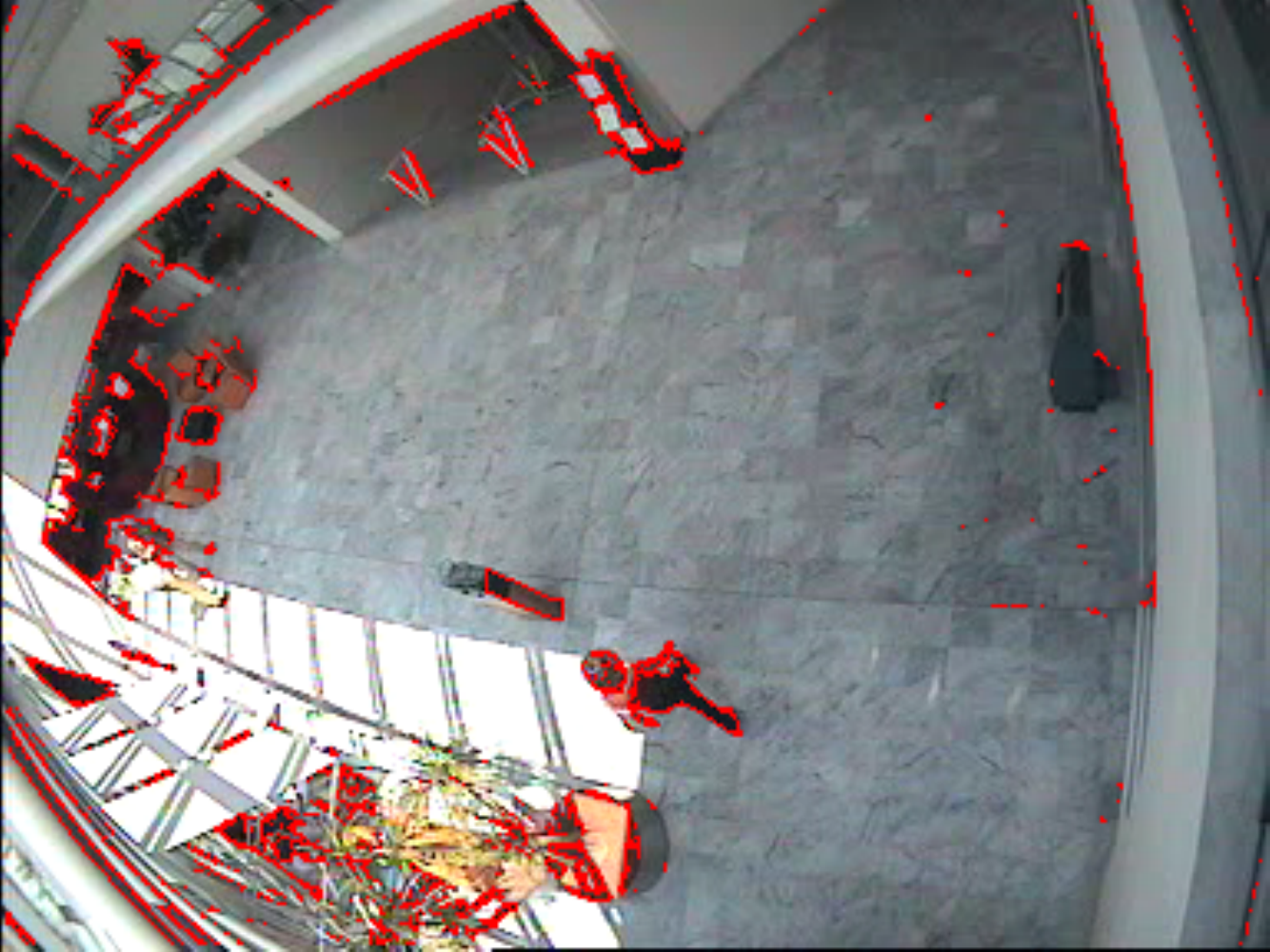}
\caption{Contours located for four frames of CAVIAR video}\label{fig46n}
\end{figure}

Despite this limitation, hand drawn contours as reference may be used e.g.~for the tracking
of a moving object. In this case, it is sufficient to locate the strong foreground outline
of the required object. Figure \ref{fig46n} illustrates some frames
of CAVIAR video where a human is detected in each frame. The hand drawn image used as training image
is taken from the same video set but not the same image.

\subsection{Using pairs of artificial patterns as training images}

\begin{figure}[ht!]
\includegraphics[width=4 cm]{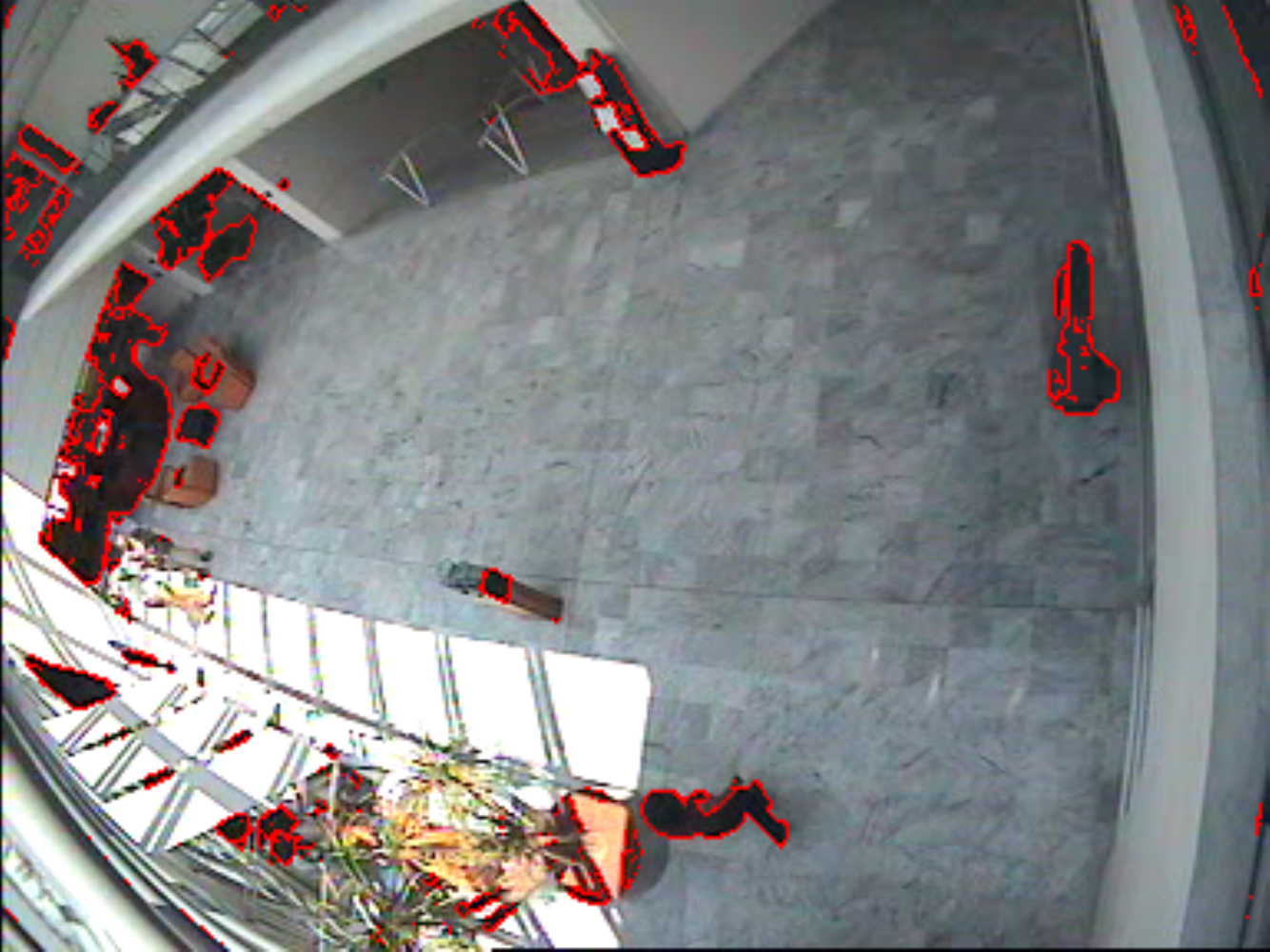}
\includegraphics[width=4 cm]{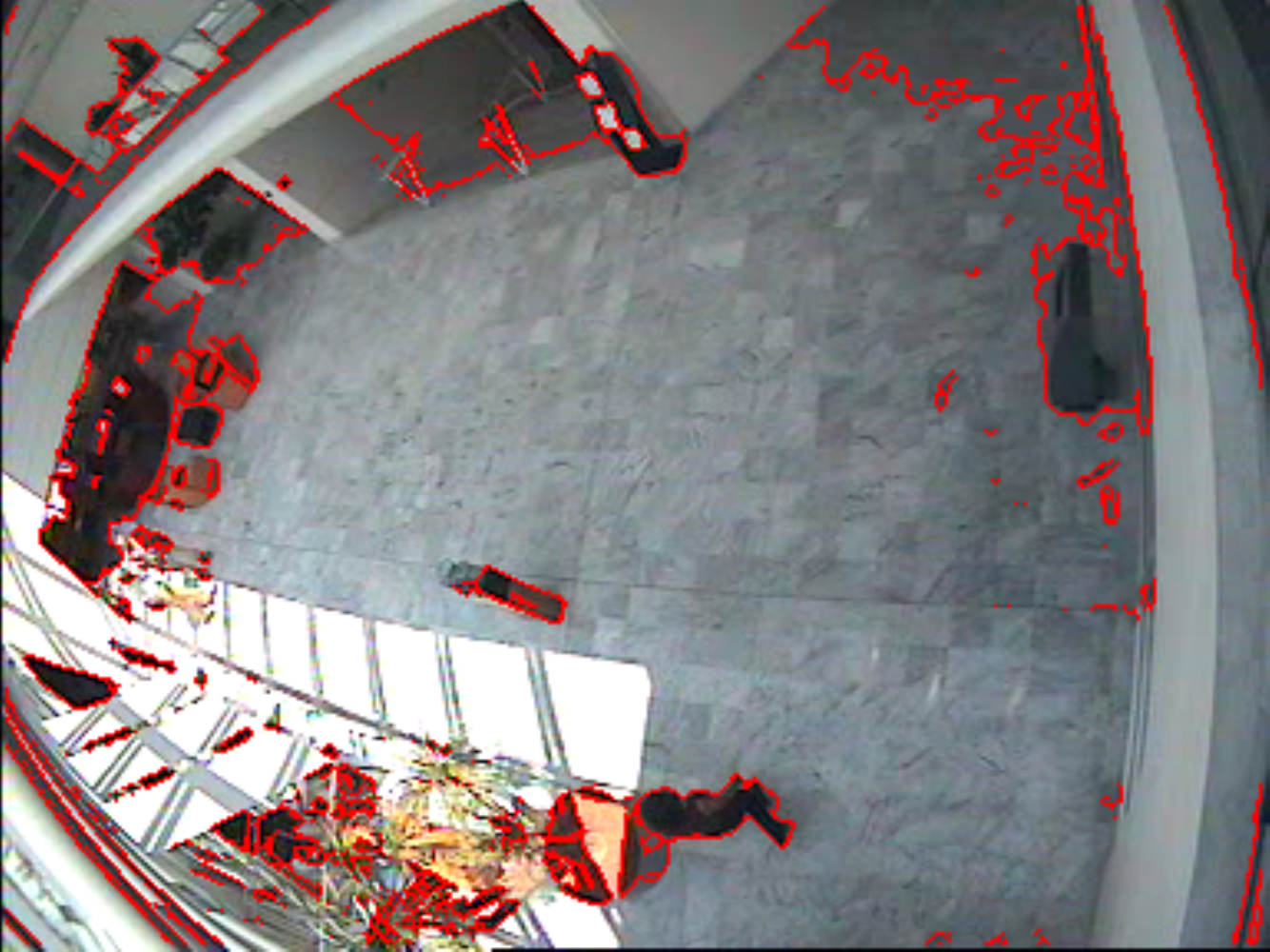}
\includegraphics[width=4 cm]{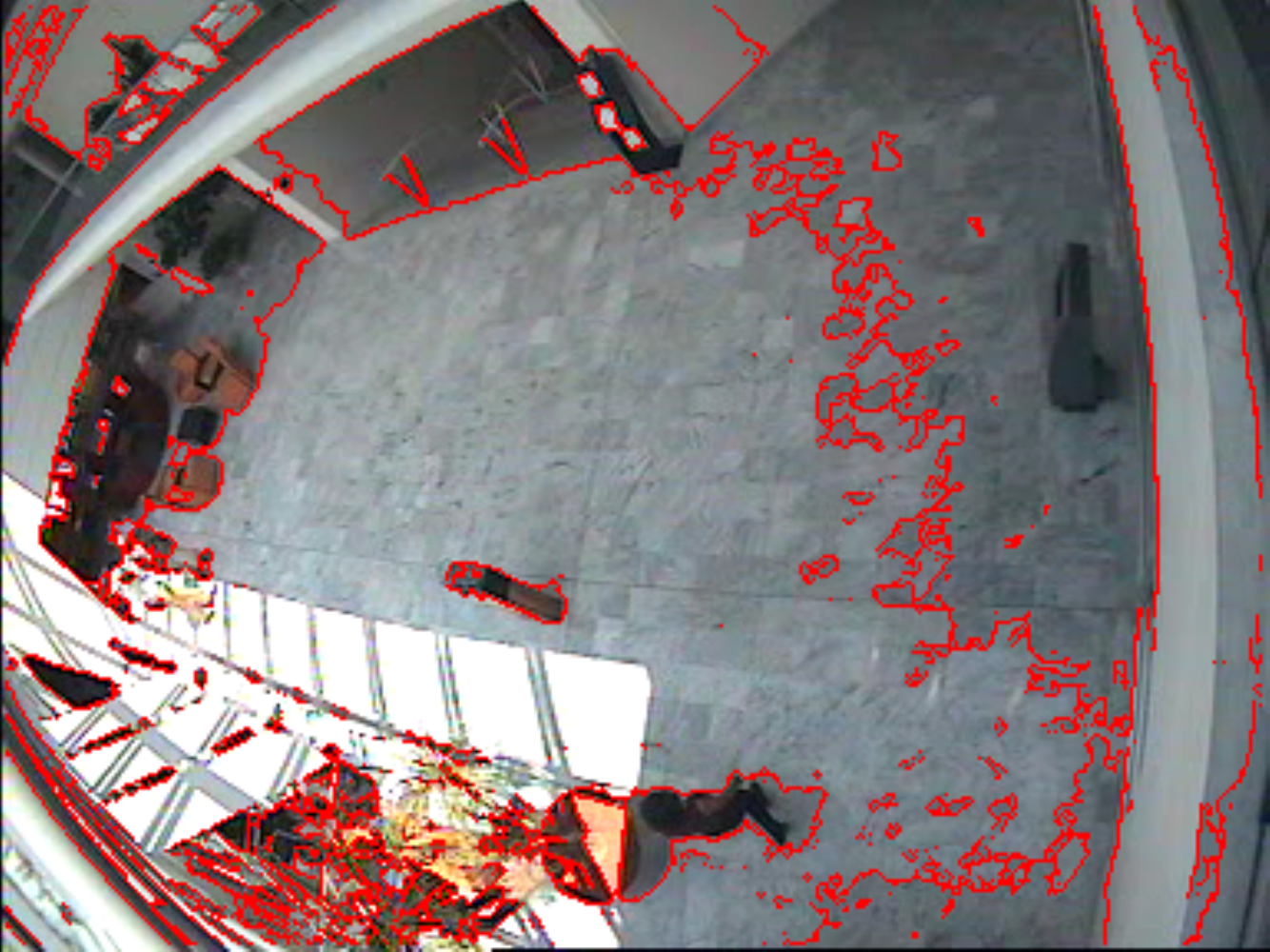}\\
\includegraphics[width=4 cm]{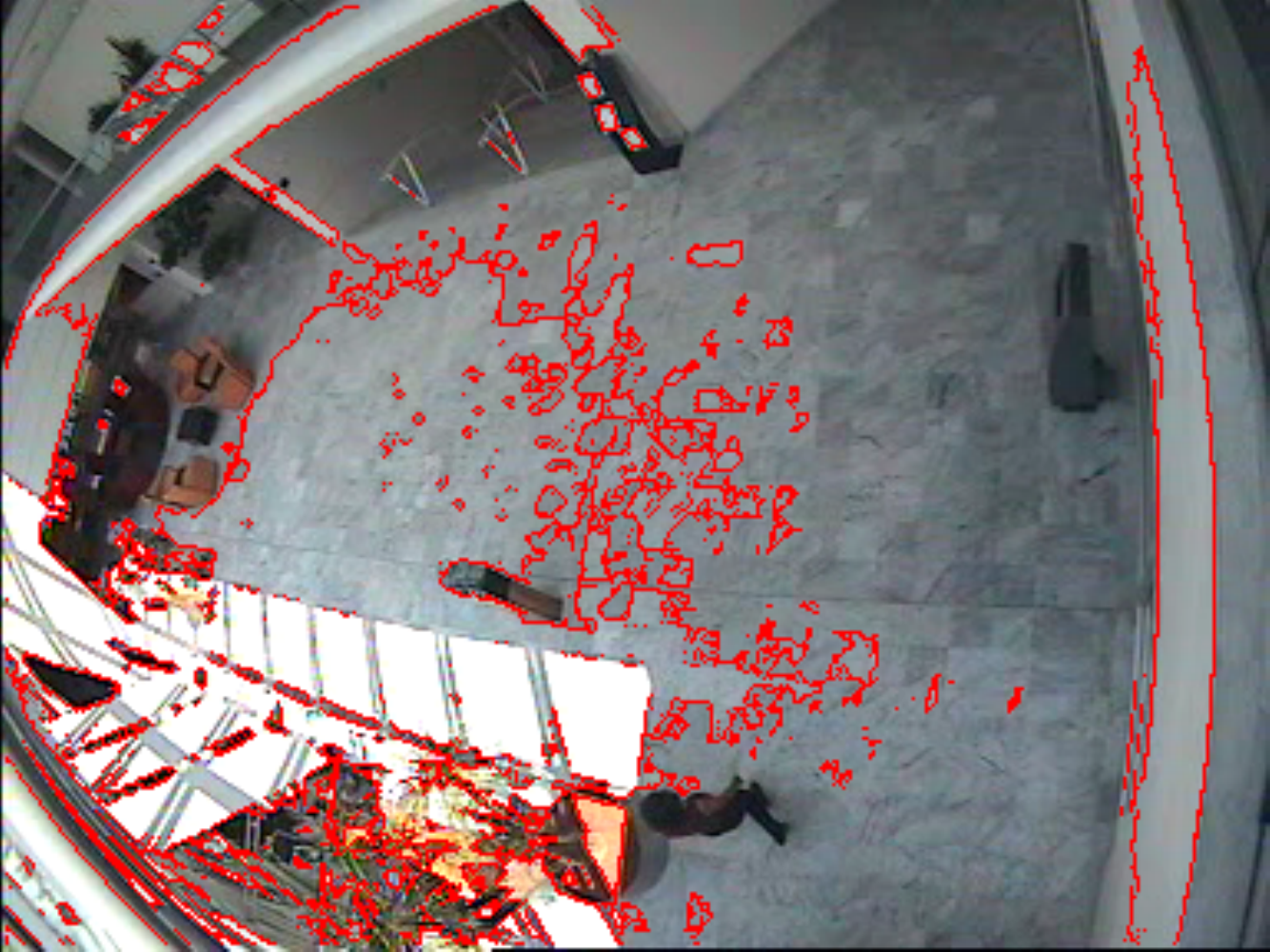}
\includegraphics[width=4 cm]{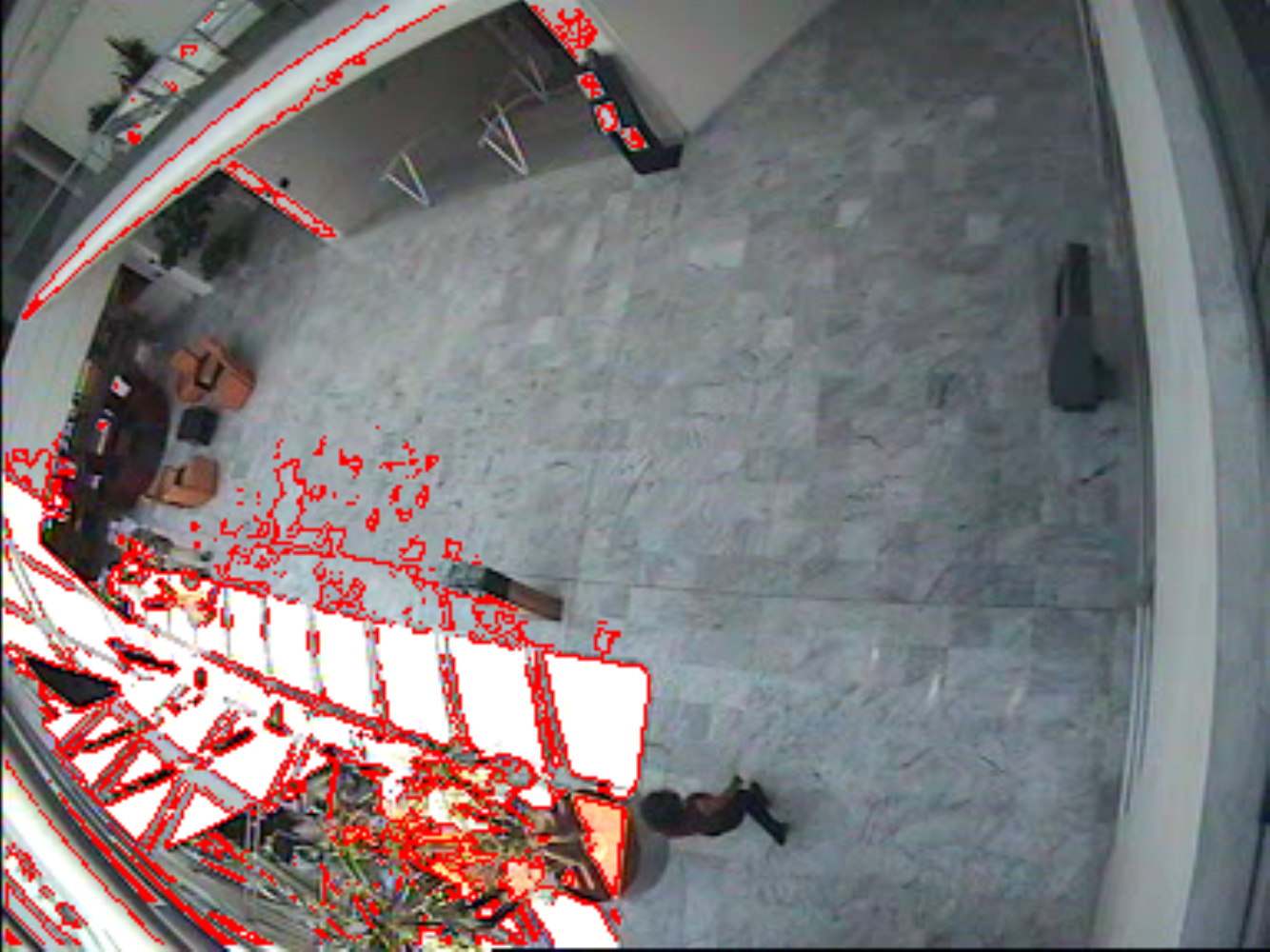}
\includegraphics[width=4 cm]{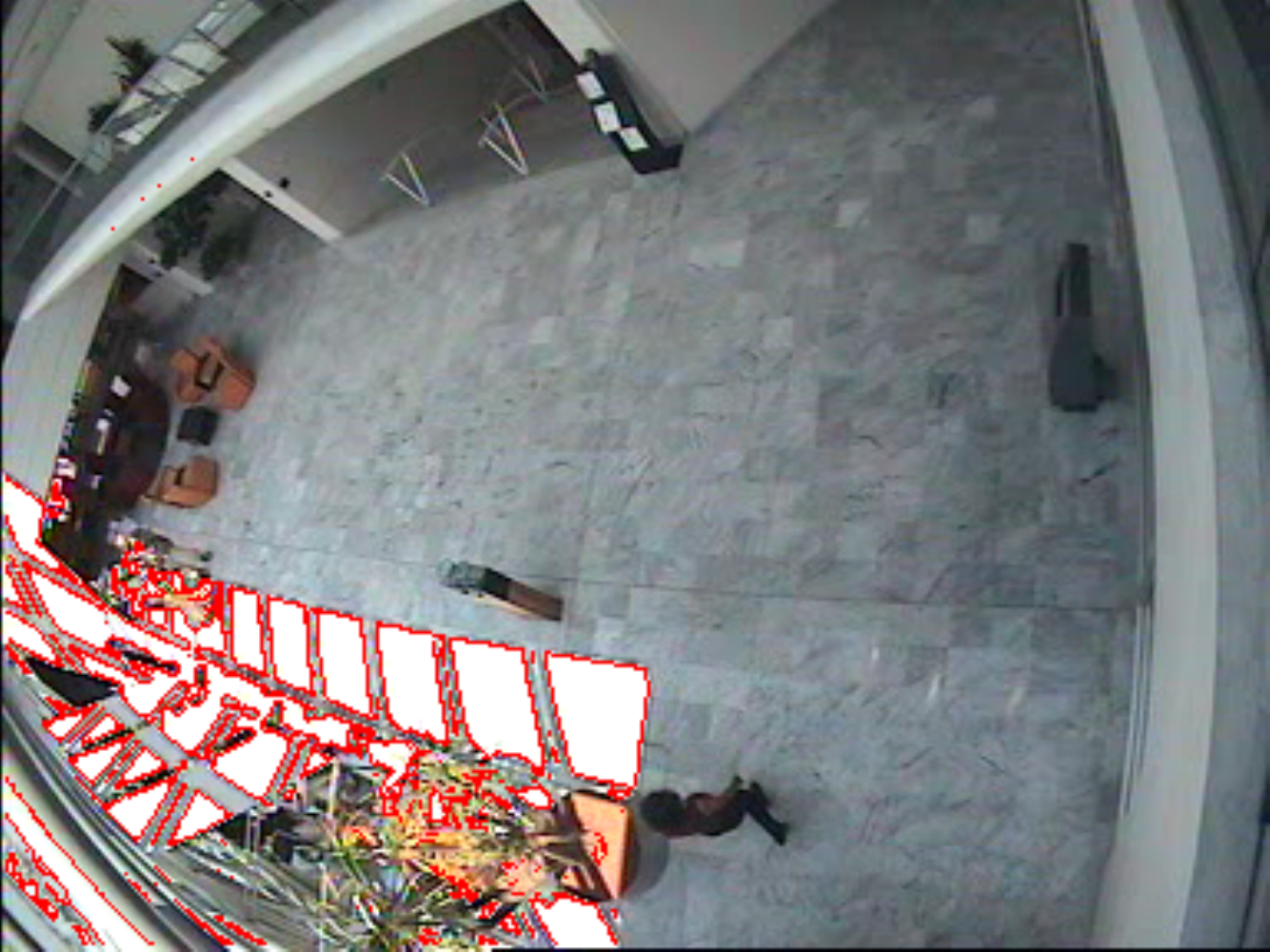}
\caption{Contours computed using the patterns $P_{1,3}$, $P_{1,5}$, $P_{1,7}$, $P_{1,9}$, $P_{1,11}$, $P_{1,14}$}\label{fig52n}
\end{figure}

Each one of the $14$  pairs artificial patterns $(A, A')$ enables us to detect a
specific level of contour depending on the intensities of the neighboring regions to the border.
Figure \ref{fig52n} illustrates contours located on the same frame of CAVIAR video using some
patterns of the set of  $14$ patterns. We can see that the outline is moving in image as
explained in subsection \ref{LevelsOfContours} from the darkest region (applying the pattern $P_{1,1}$)
to the clearest one (applying the pattern $P_{1,14}$).

\begin{figure}[ht!]
\includegraphics[width=4 cm]{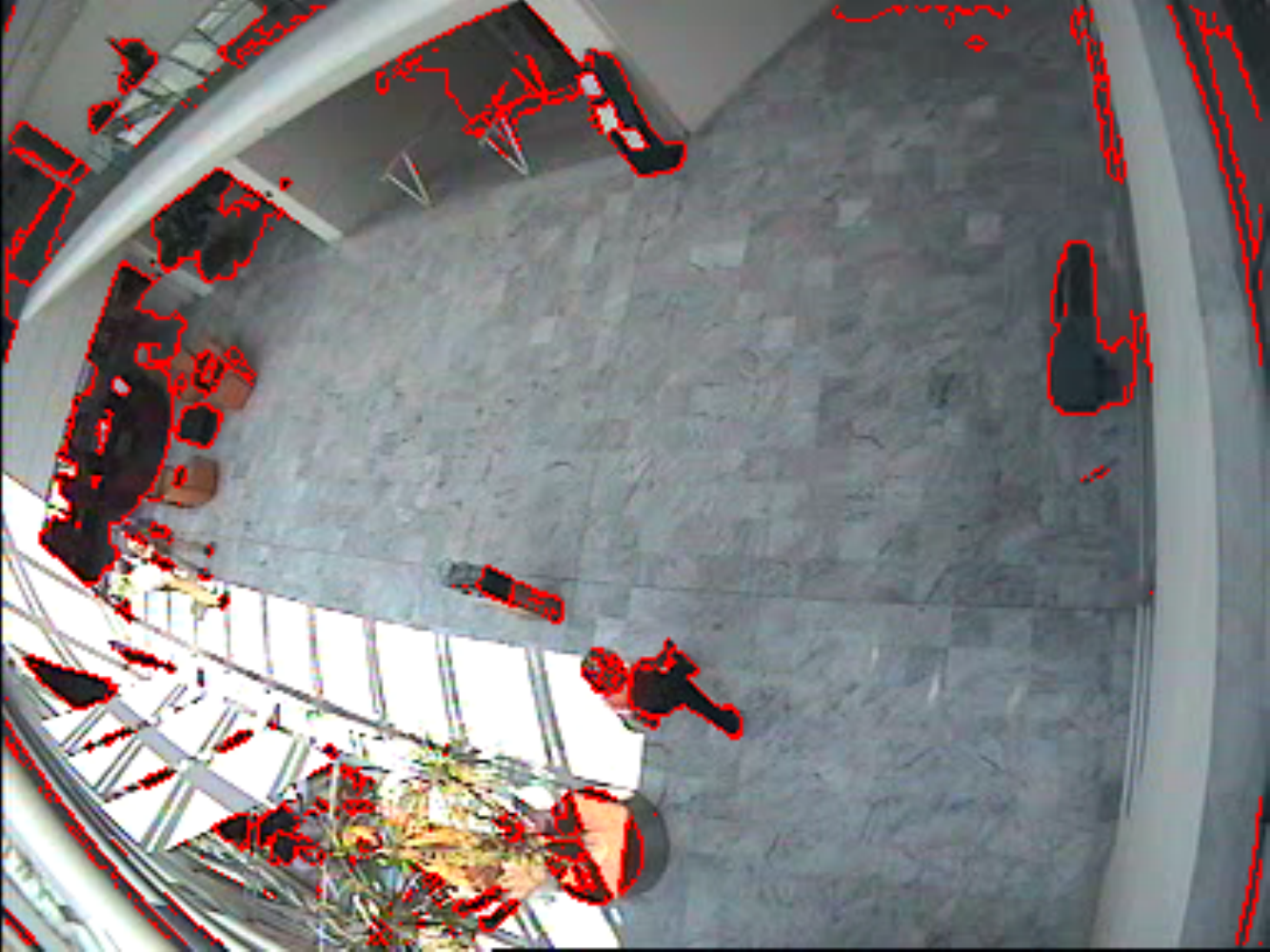}
\includegraphics[width=4 cm]{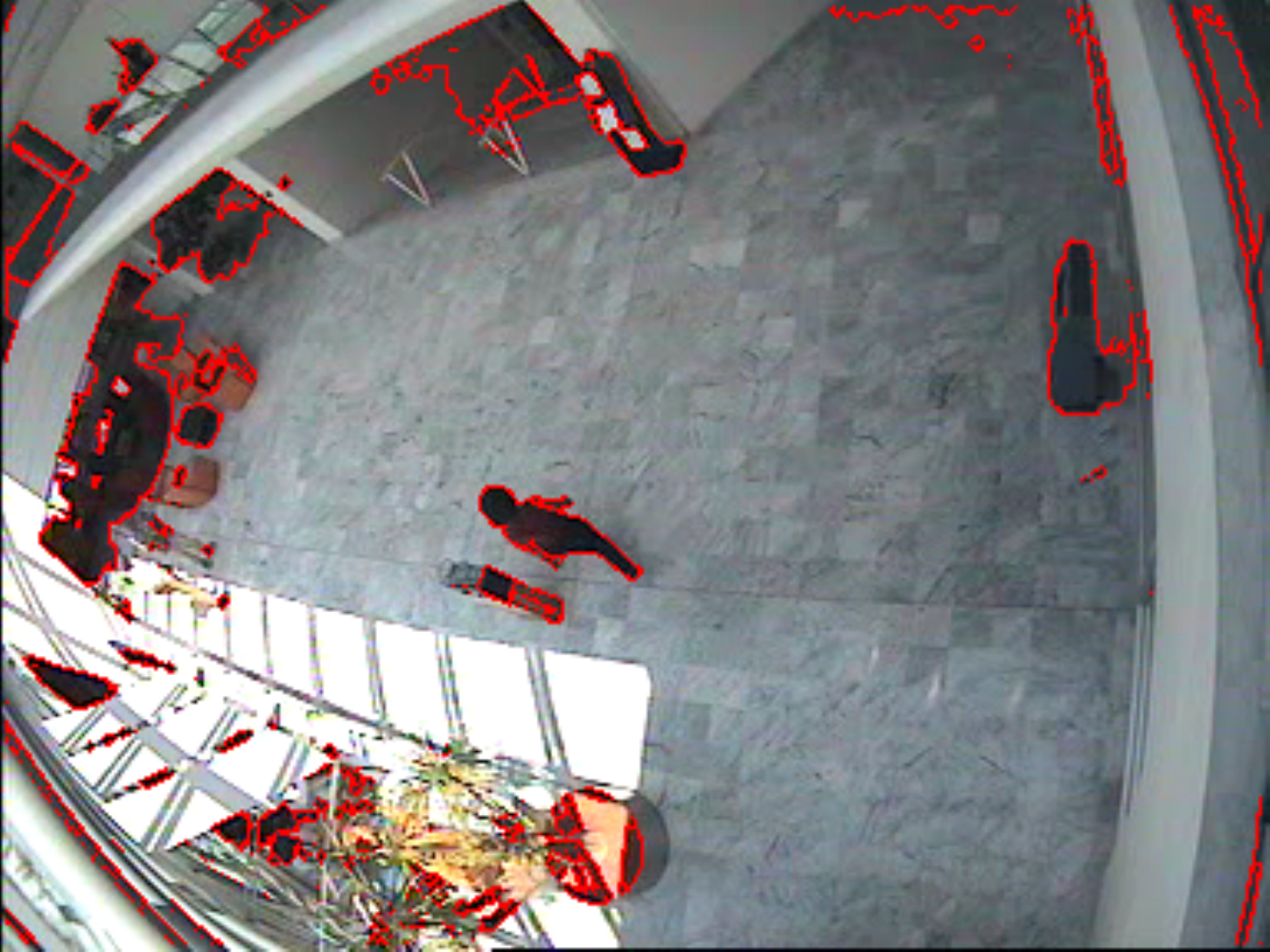}
\includegraphics[width=4 cm]{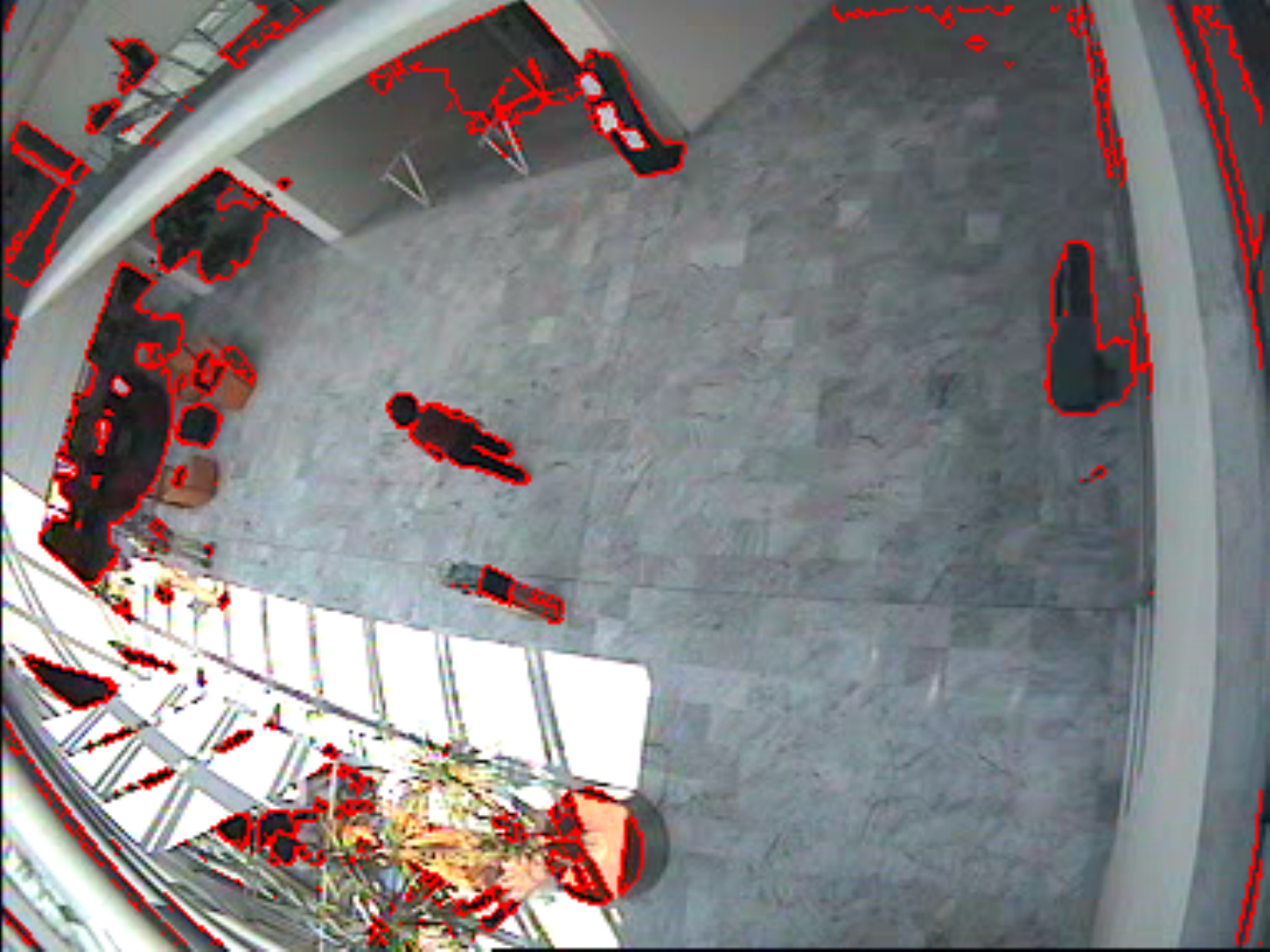}
\caption{Contours located using the pattern $P_{1,4}$ for a first, a second frame, and a third frame of the CAVIAR dataset}\label{fig55n}
\end{figure}

The application of one level over a set of frames from the CAVIAR video produced the results shown by
figure \ref{fig55n}. In this case the suitable pair of patterns have been
chosen in order to locate the moving human. Indeed, if another pair of patterns is applied, the human
outline will be (or partially) not located depending on how well it is represented in the training image.

\subsection{Contour detection at different levels of resolution}

The proposed method as described in the  previous section allows to detect contours
in image specifying the required level. This level is related to the intensity variation between
pixels at the sides of the outline contour.
The level $l$ is associated to the contour which is located by successive $l$ patterns:
$P^1_j, P^1_{j+1}, ..., P^1_{j+l-1}$.
Then the position of the correspondent value $I^f_B$ for such contour is located by the recovering
of the intervals $]I^b_{B,j+k}, 255], k=0, l-1$ as seen in subsection \ref{LevelsOfContours}.

\begin{figure}[ht!]
\centering
\includegraphics[width=4 cm]{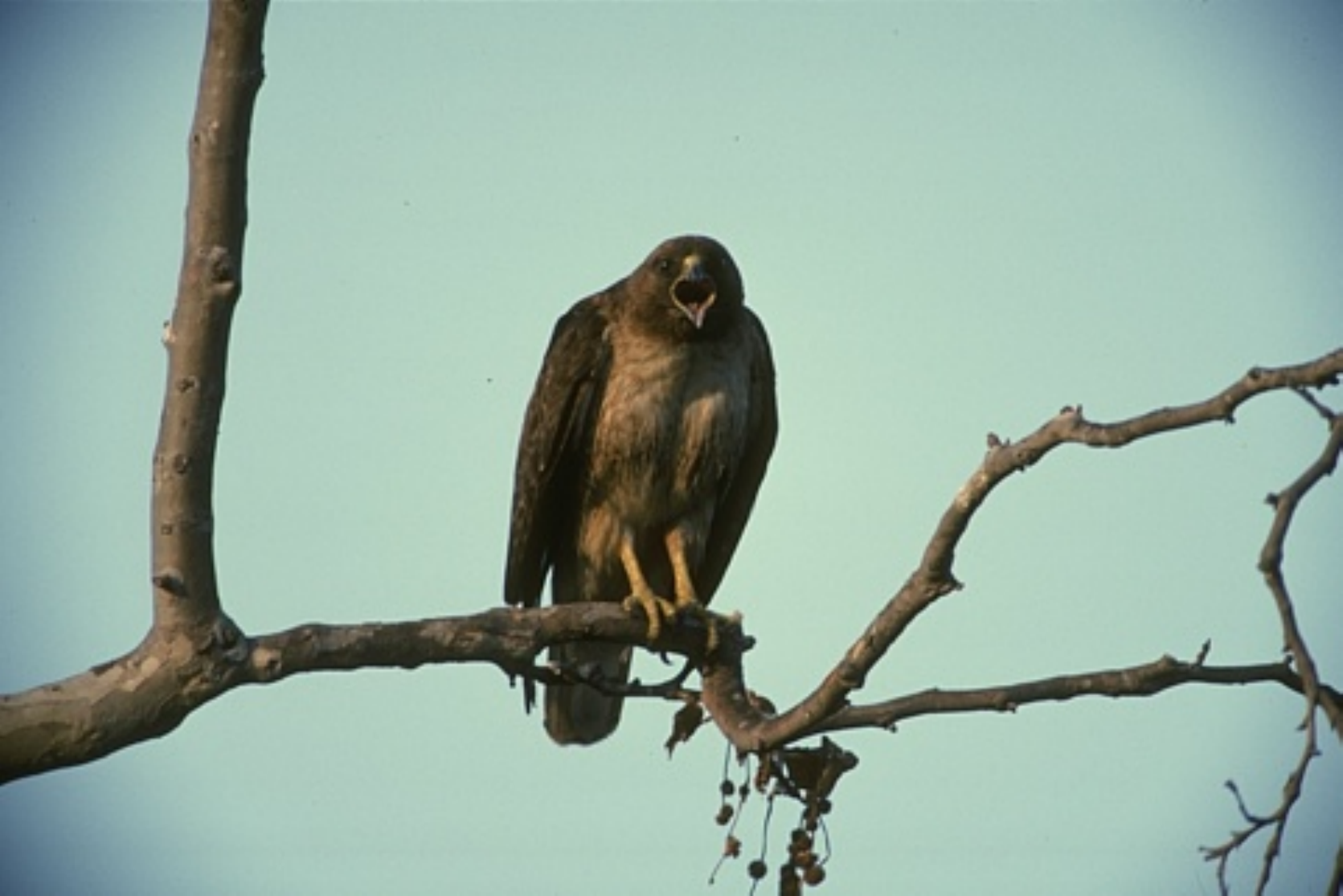}
\includegraphics[width=4 cm]{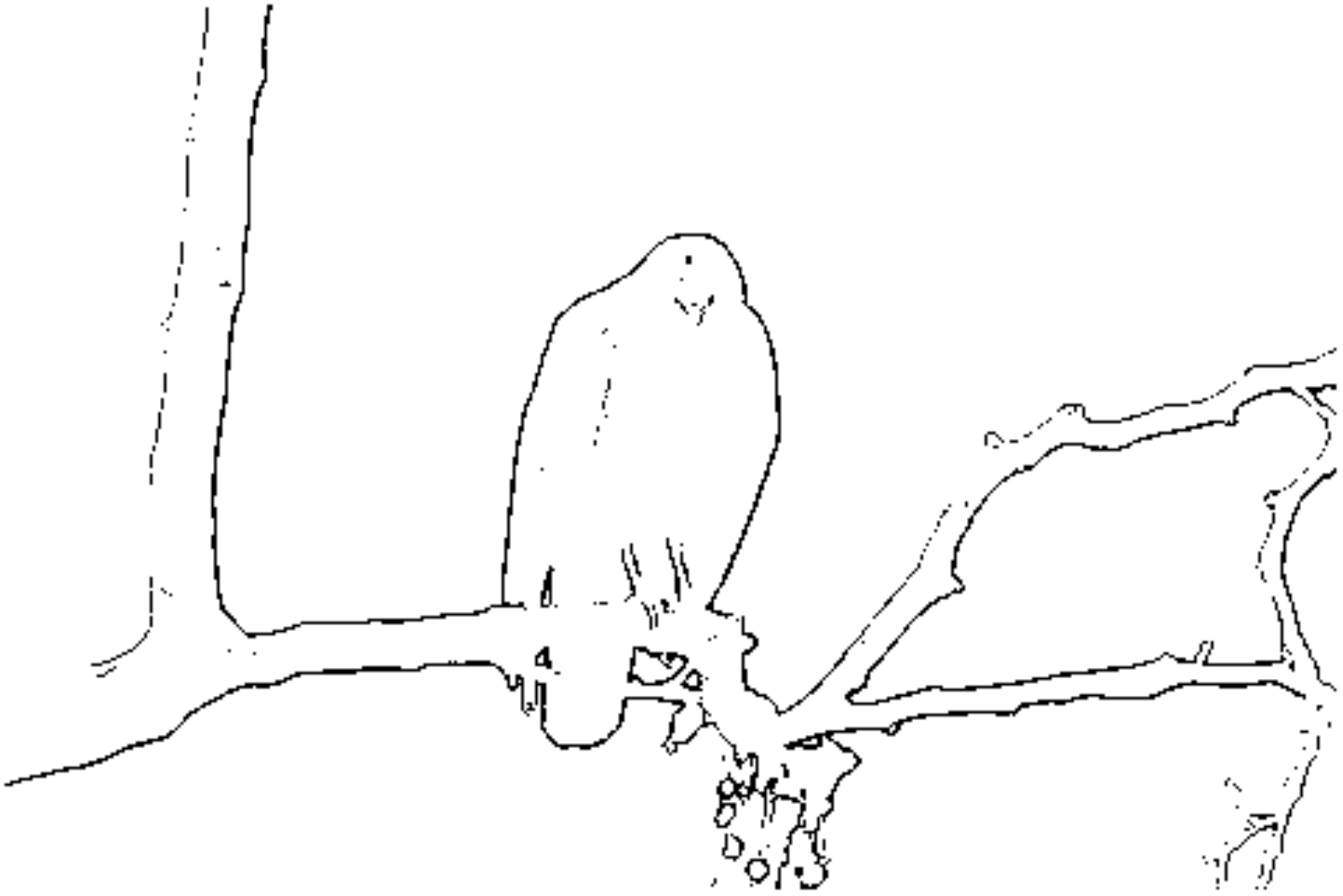}
\includegraphics[width=4 cm]{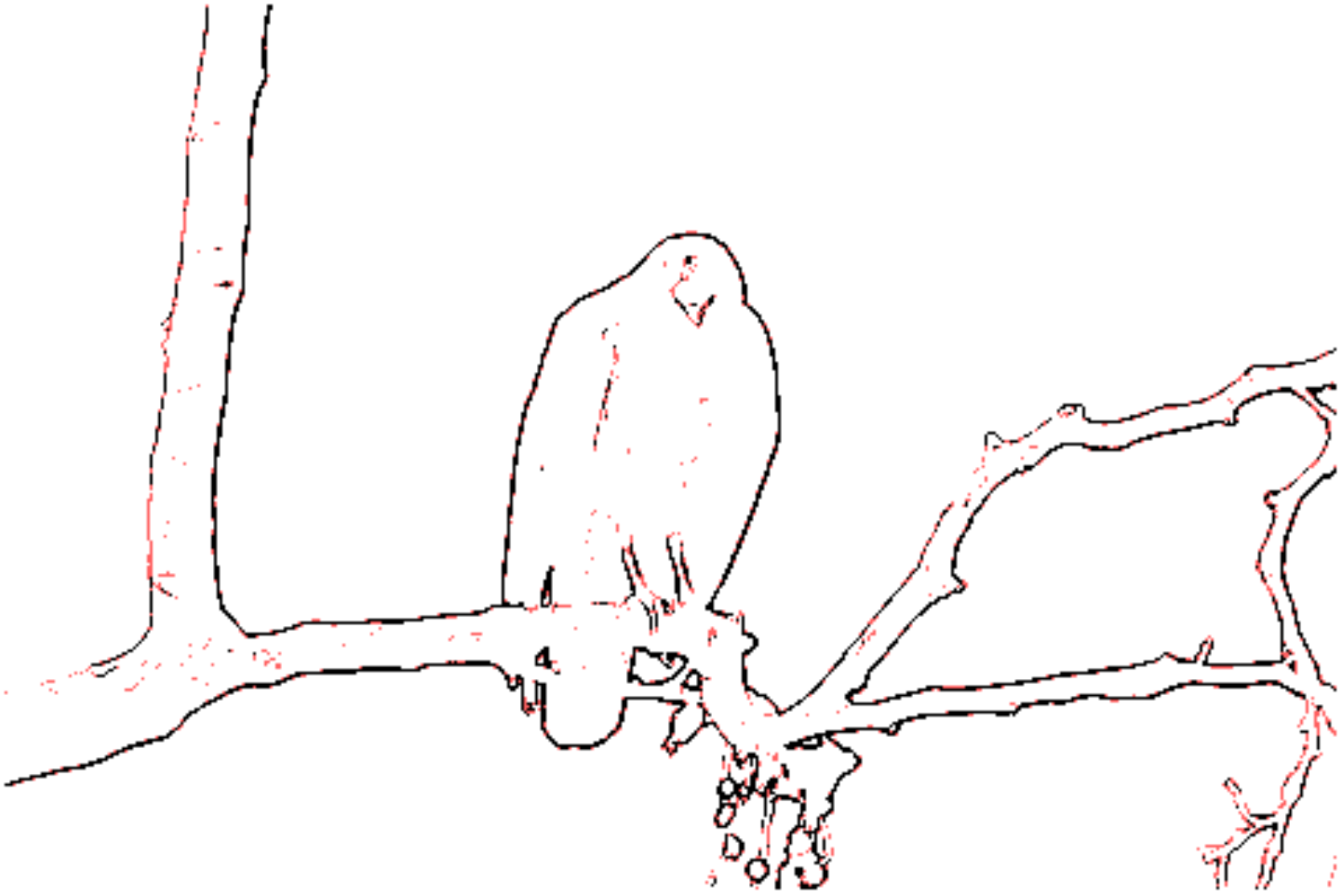}
\includegraphics[width=4 cm]{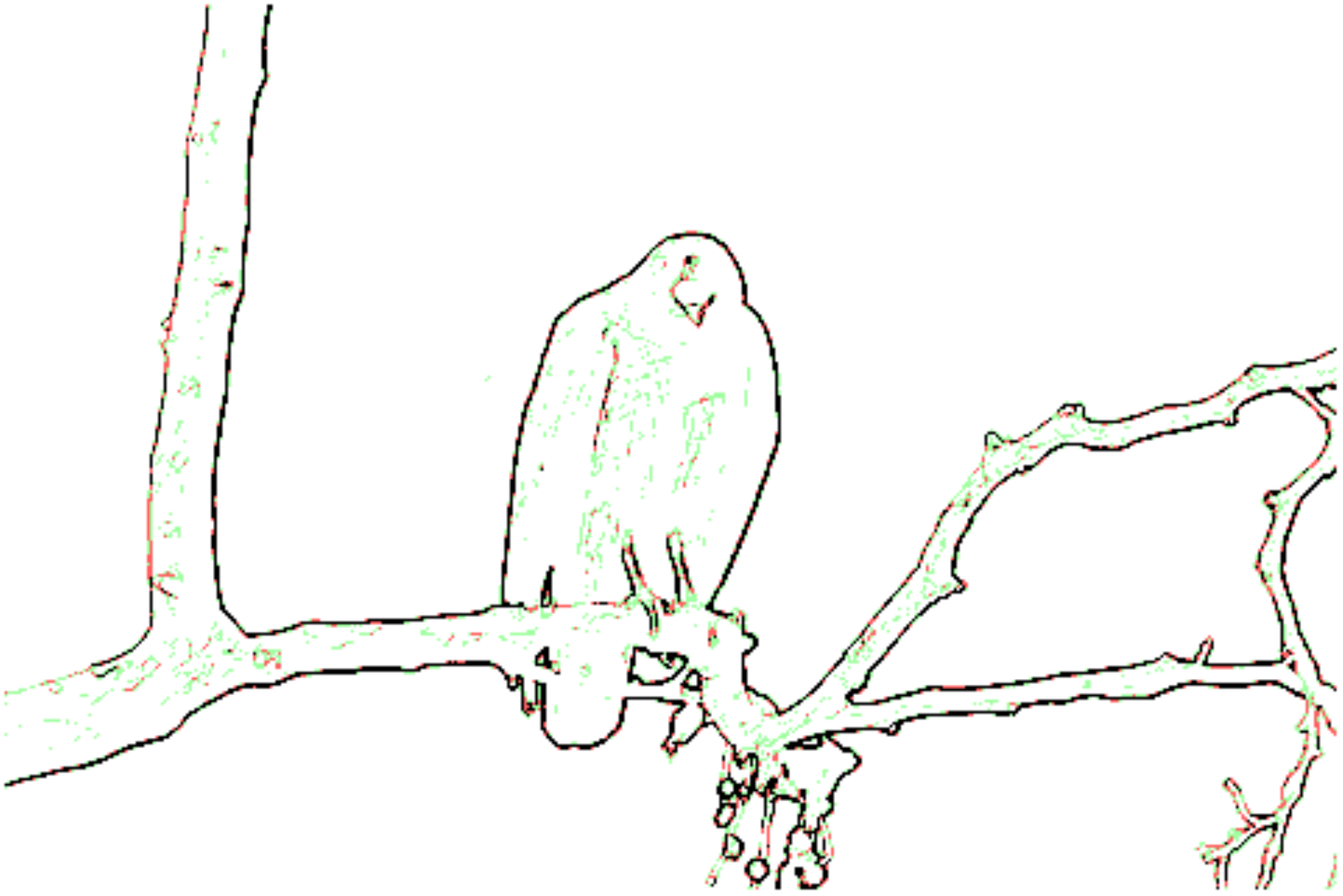}
\includegraphics[width=4 cm]{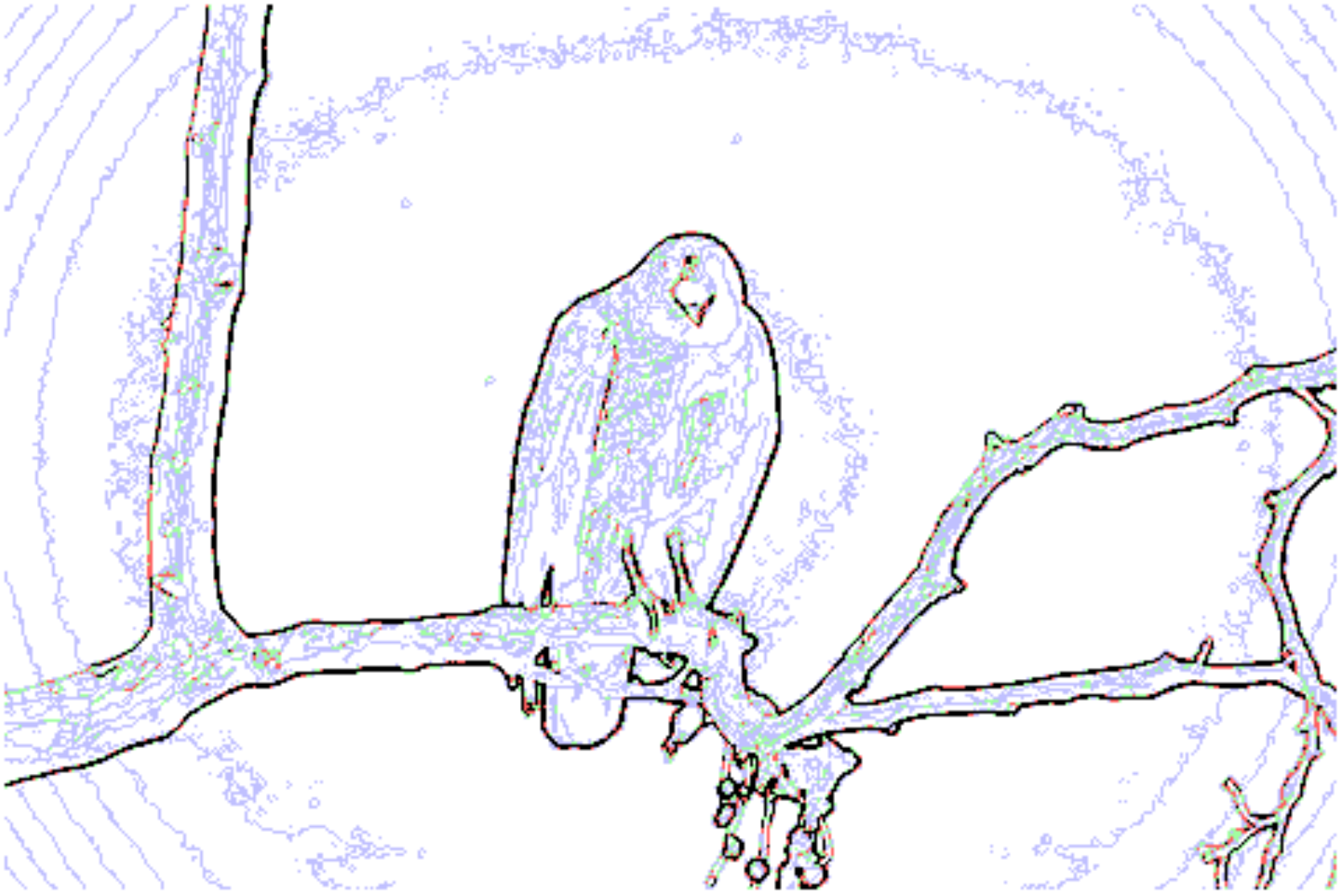}
\caption{Original image, Contours of level $4, 3, 2 $ using the $14$ patterns}\label{fig60n}
\end{figure}

Figure \ref{fig60n} shows an example of located contours on image of BSD dataset
using the set of $14$ pairs of artificial patterns. We can see the low frequency outlines which
correspond to the value of $l=4$ and the high frequency outlines which correspond value of $l=1$.
The color black is associated to contours of level $4$ and colors red, green and blue are used
to distinguish the new contours detected for the considered levels $3$, $2$ and $1$. The use
of the set of $28$ instead of $14$ pairs of patterns for the same image produces contours
at higher resolution for the same level. In summary, more the level
is greater, more outline contours are found and thus contours corresponding to high difference
of intensity between regions will remain in the computed image.


\subsection{Affine transformation Invariance}

\begin{figure}[ht]
\centering
\includegraphics[width=4 cm]{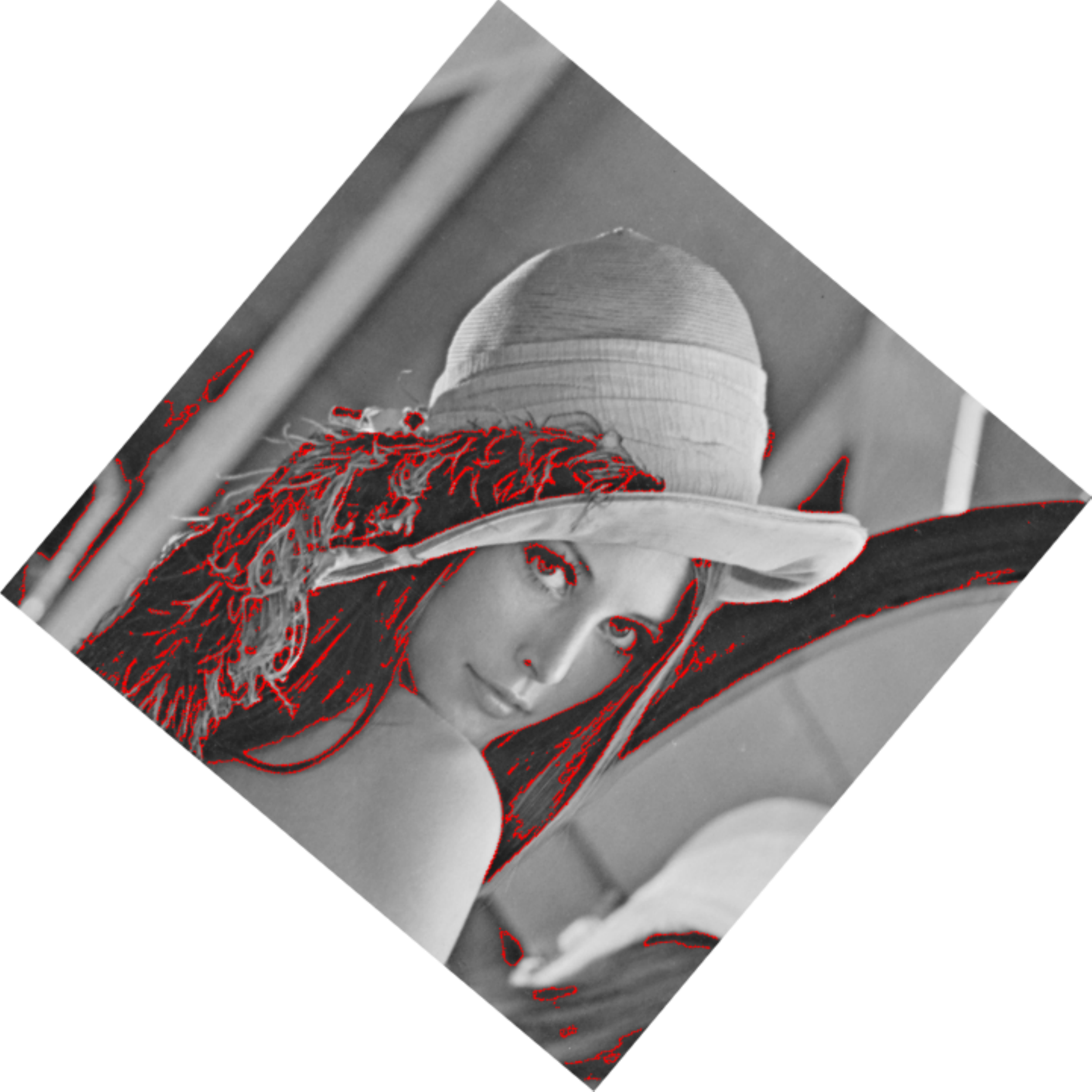}
\includegraphics[width=4 cm]{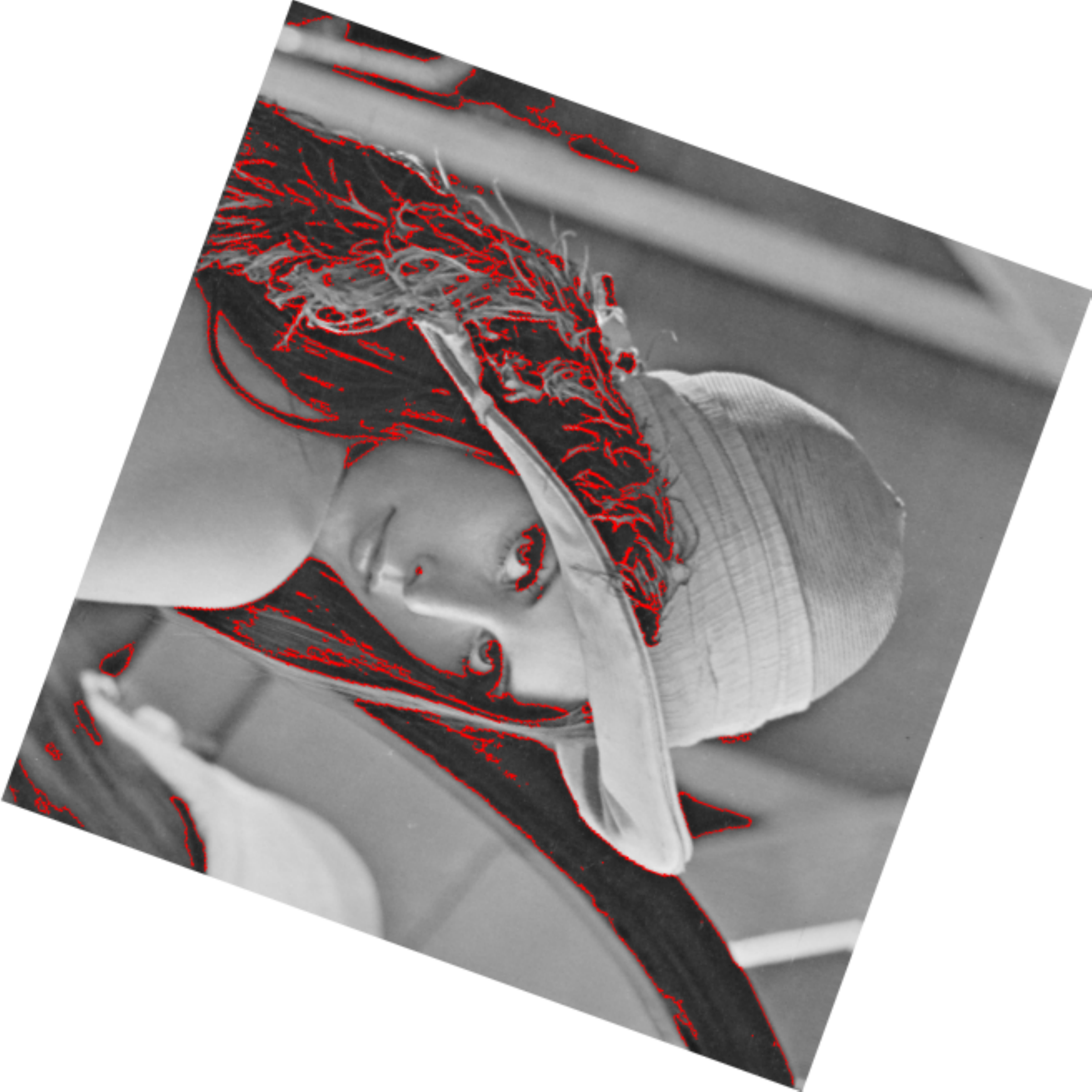}
\includegraphics[width=4 cm]{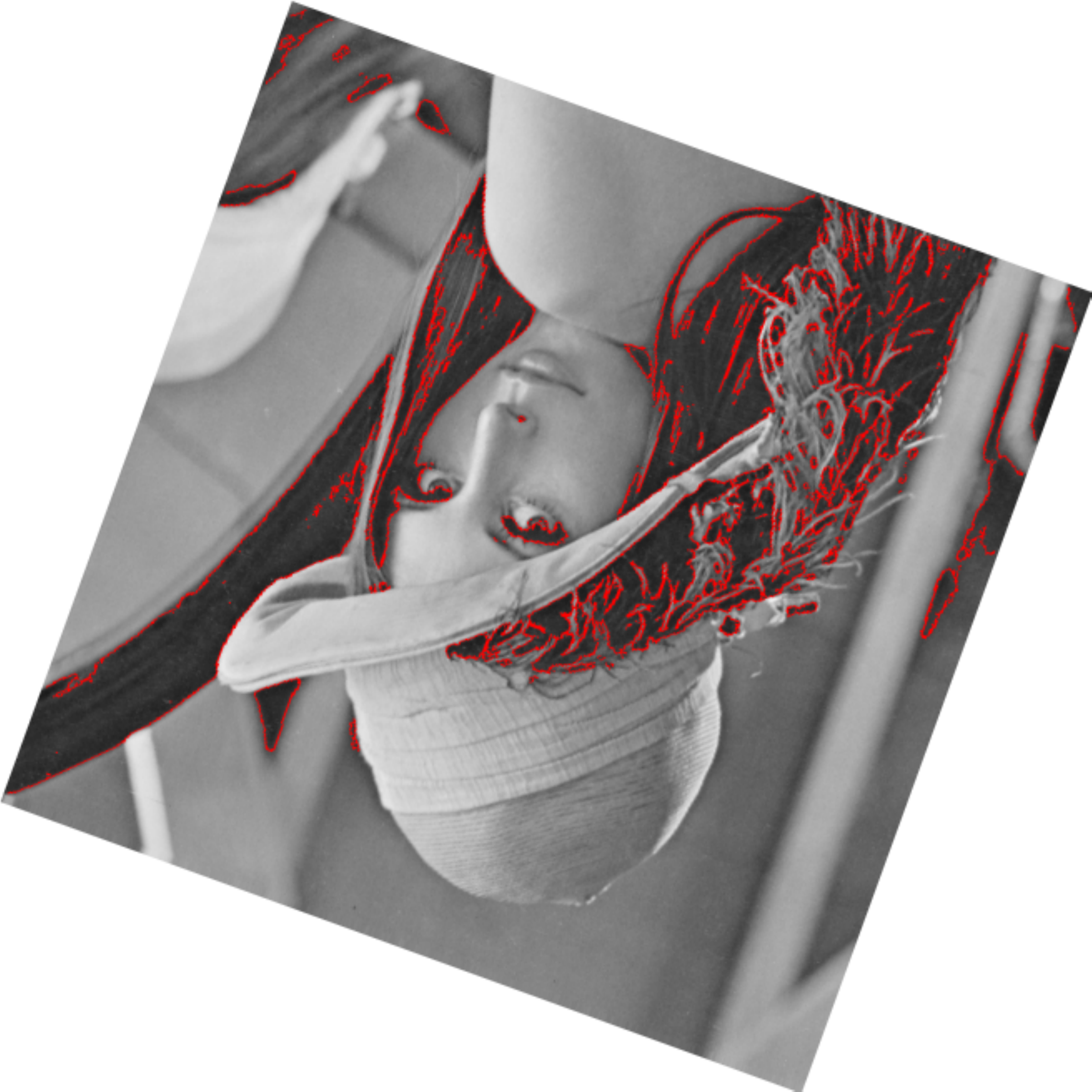}
\includegraphics[width=4 cm]{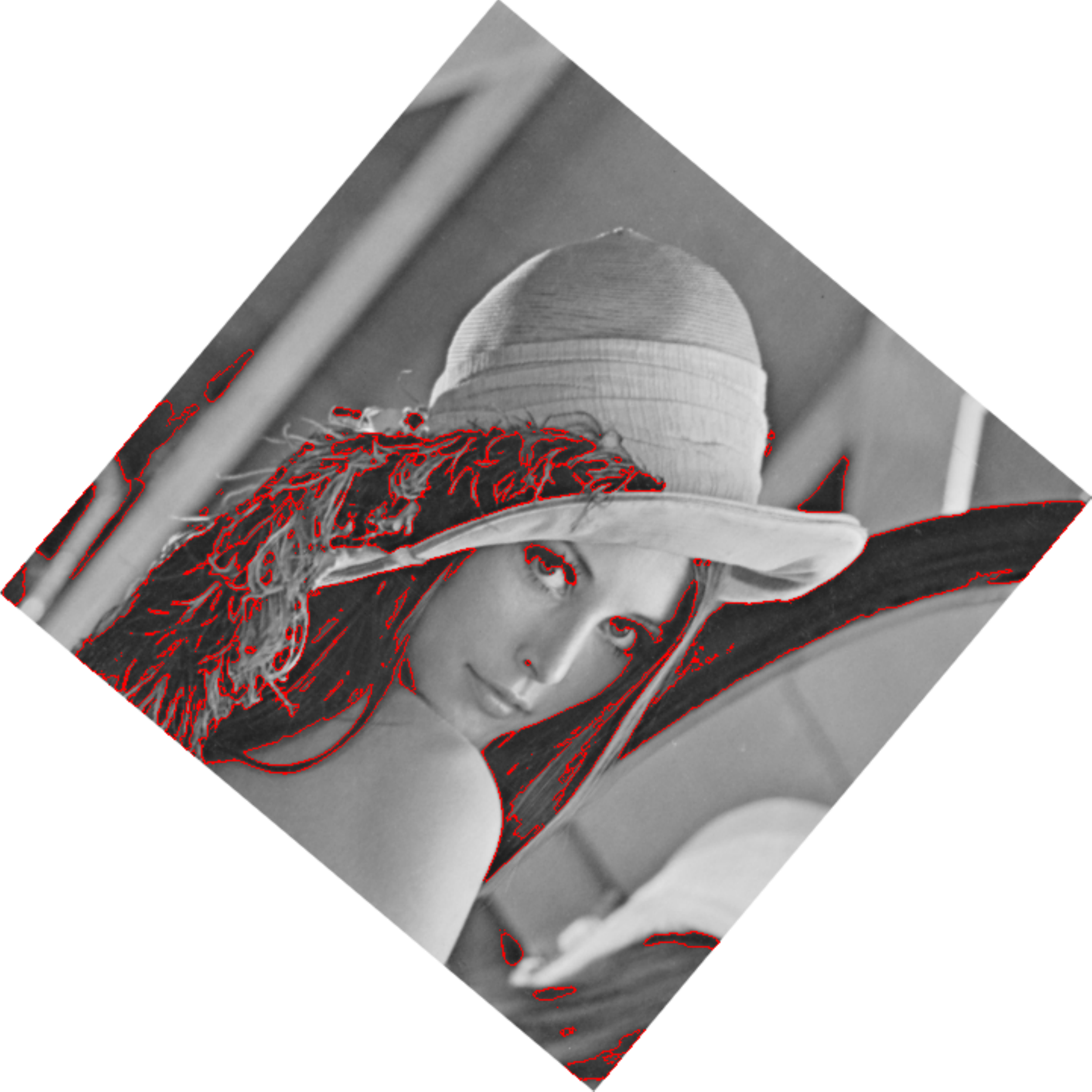}
\includegraphics[width=4 cm]{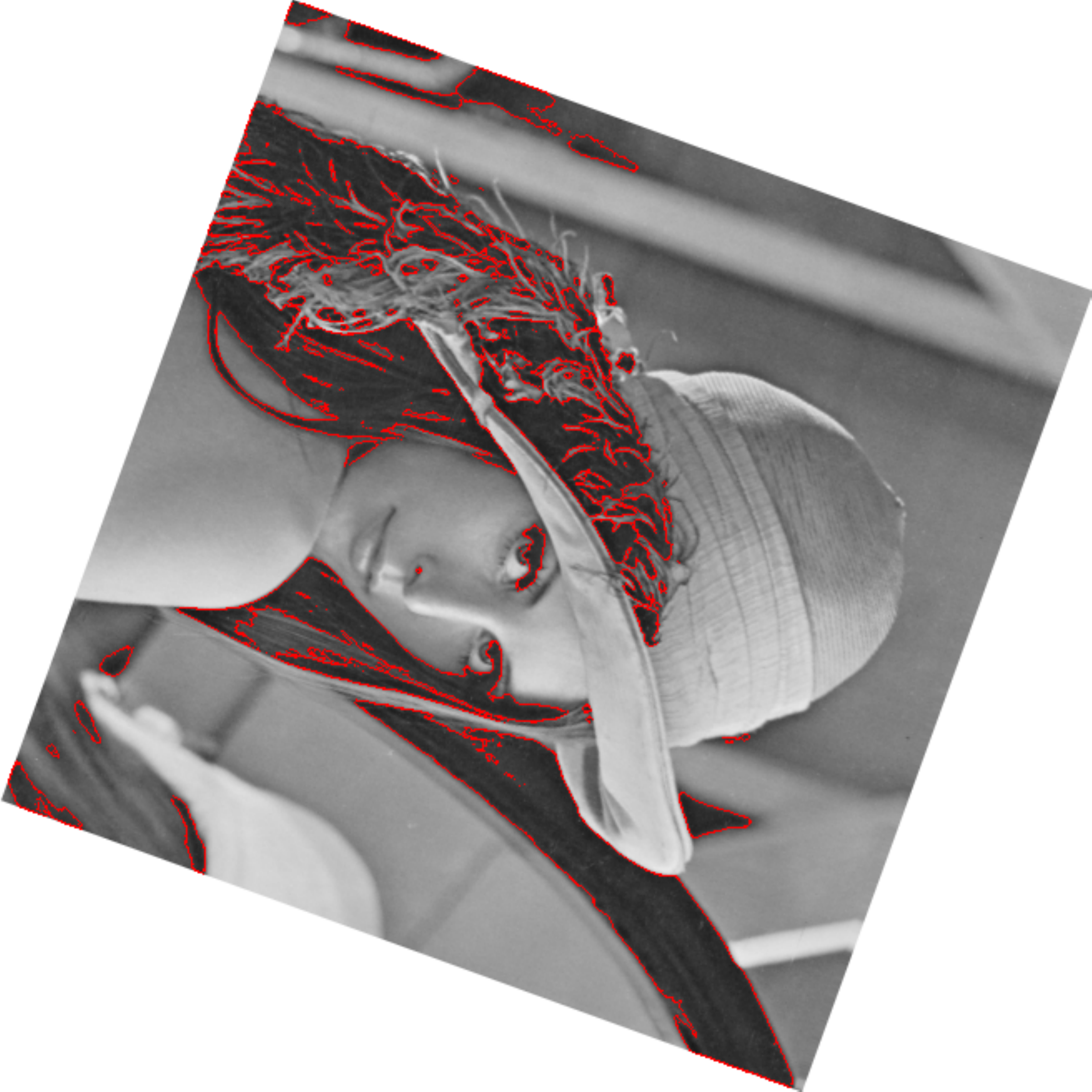}
\includegraphics[width=4 cm]{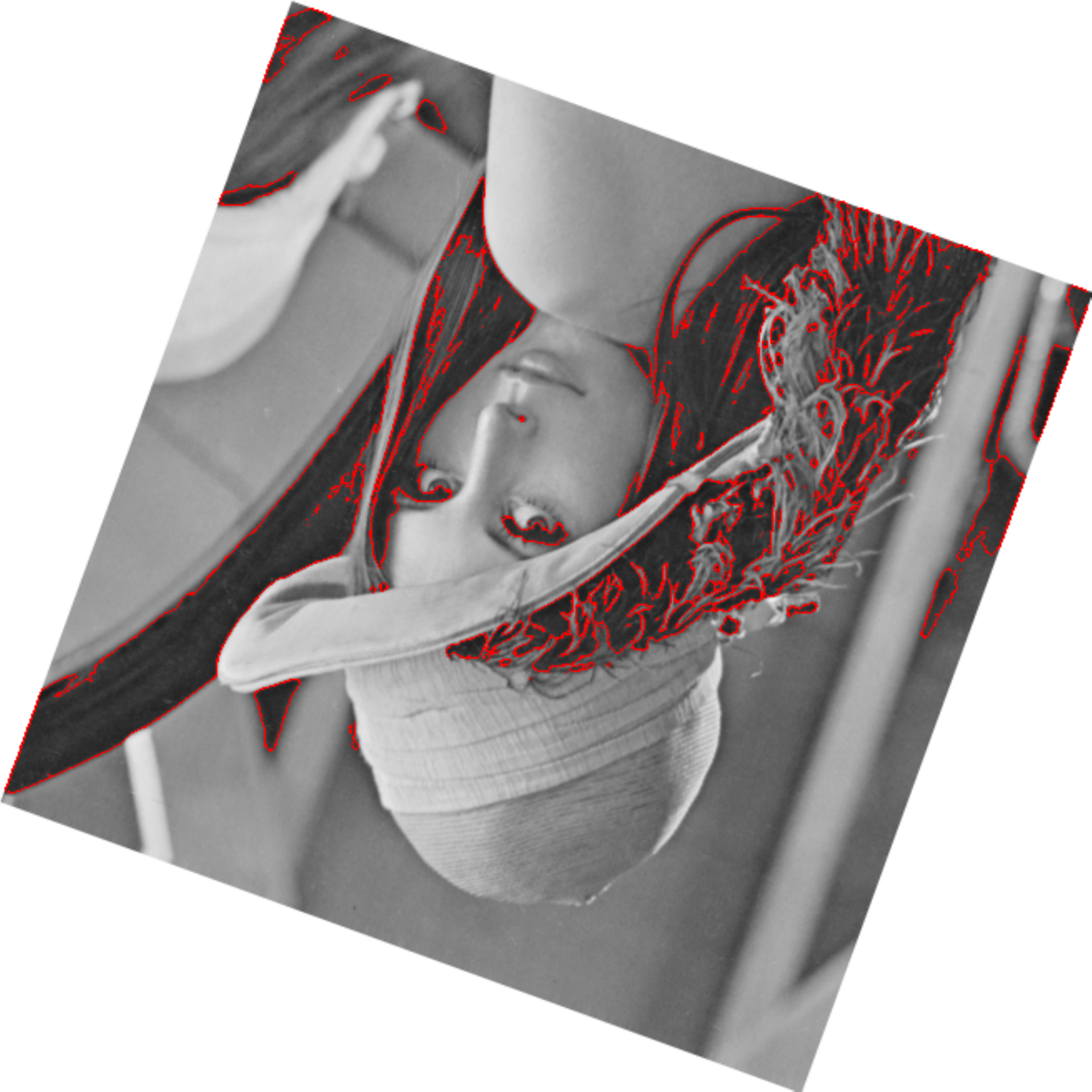}
\caption{From top to bottom, left to right: Rotated images of contours located using the pattern $P_4$ and the angle $40\,^{\circ}$, $110\,^{\circ}$, $200\,^{\circ}$.
Contours of rotated images located using the pattern $P_4$ and the angle $40\,^{\circ}$, $110\,^{\circ}$,
$200\,^{\circ}$}
\label{fig66n}
\end{figure}

We studied the invariance of our method to affine transformation such as rotation, scale change.
We compared the rotated contours computed for initial images and the contours located after image rotation.
We considered for this all contour images obtained using the set of $14$ pairs of patterns.
We measured the ratio of contour pixels that haven't been located after image rotation and
computed the average of this ratio for different images with different angles of rotation
(lena, BSD500 data set). This ratio is around $0.02$ for any rotation angle.
Figure \ref{fig66n} illustrates images obtained for "lena" image and the computed contours shown with red color.

\begin{figure}[ht!]
\centering
\includegraphics[width=4 cm]{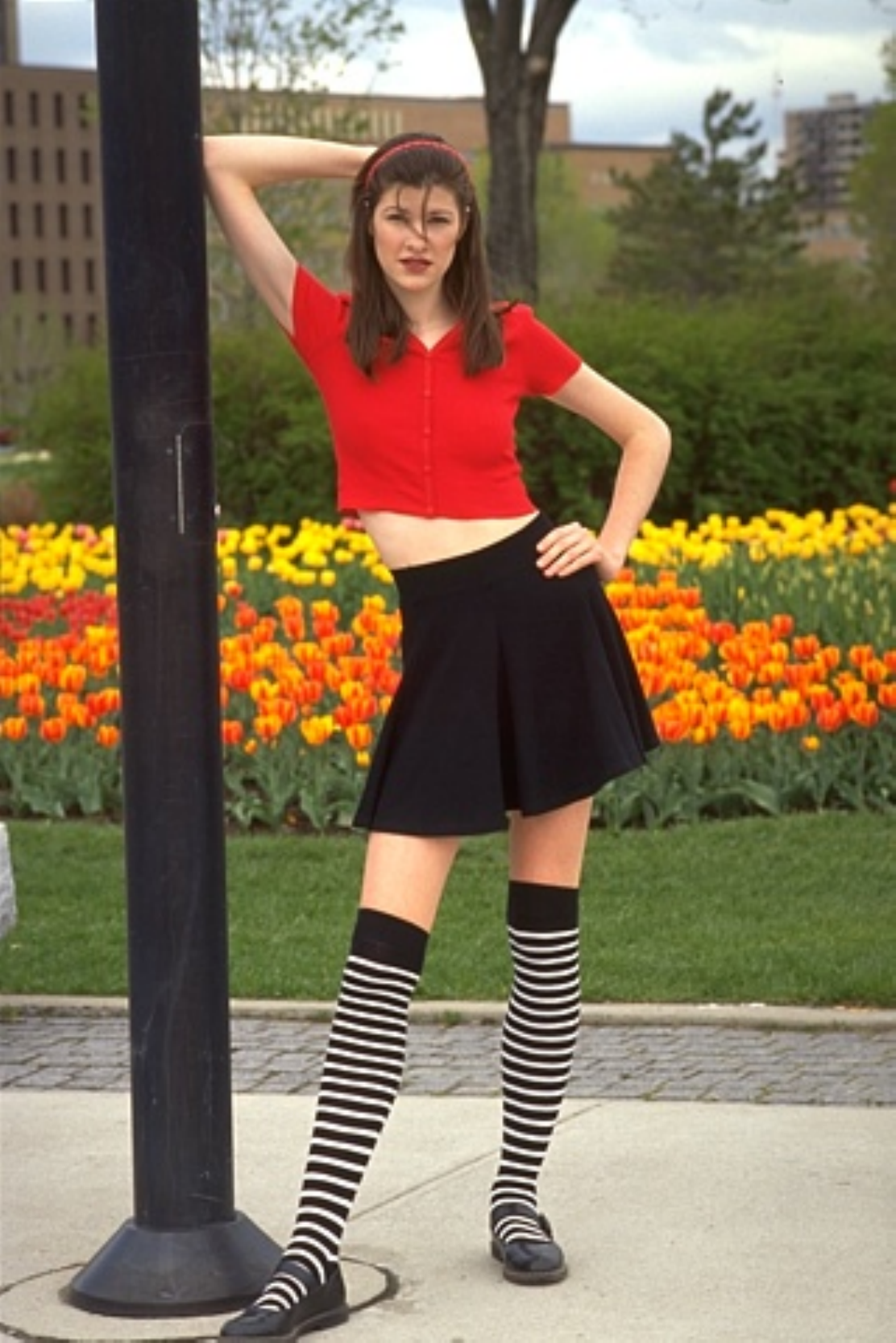}
\includegraphics[width=2.8 cm]{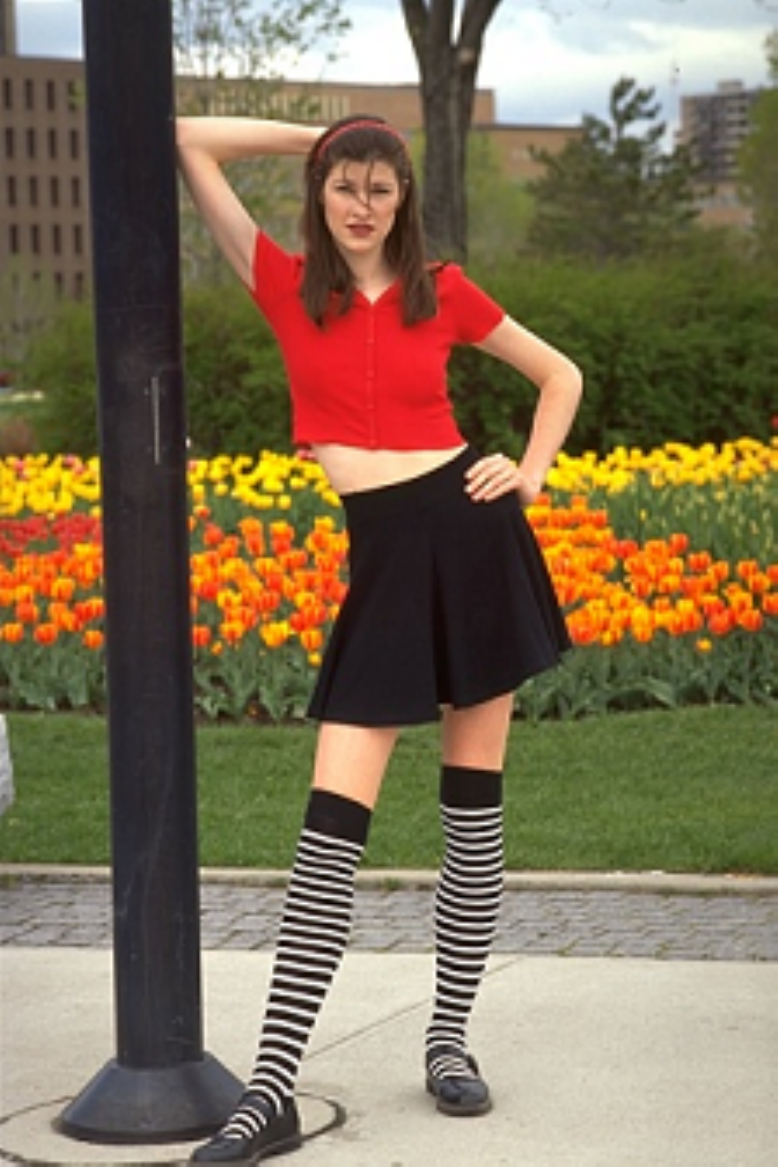}
\includegraphics[width=1.6 cm]{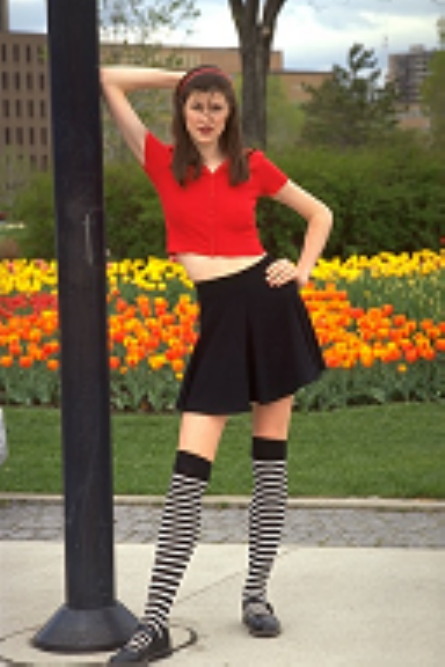}
\includegraphics[width=1 cm]{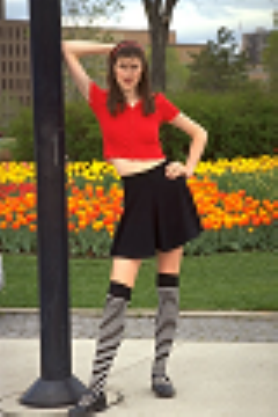}
\caption{Original image, Scaled image to 70\%, Scaled image to 40\%, Scaled image to 25\%}\label{fig70n}
\end{figure}

\begin{figure}[ht!]
\centering
\includegraphics[width=4 cm]{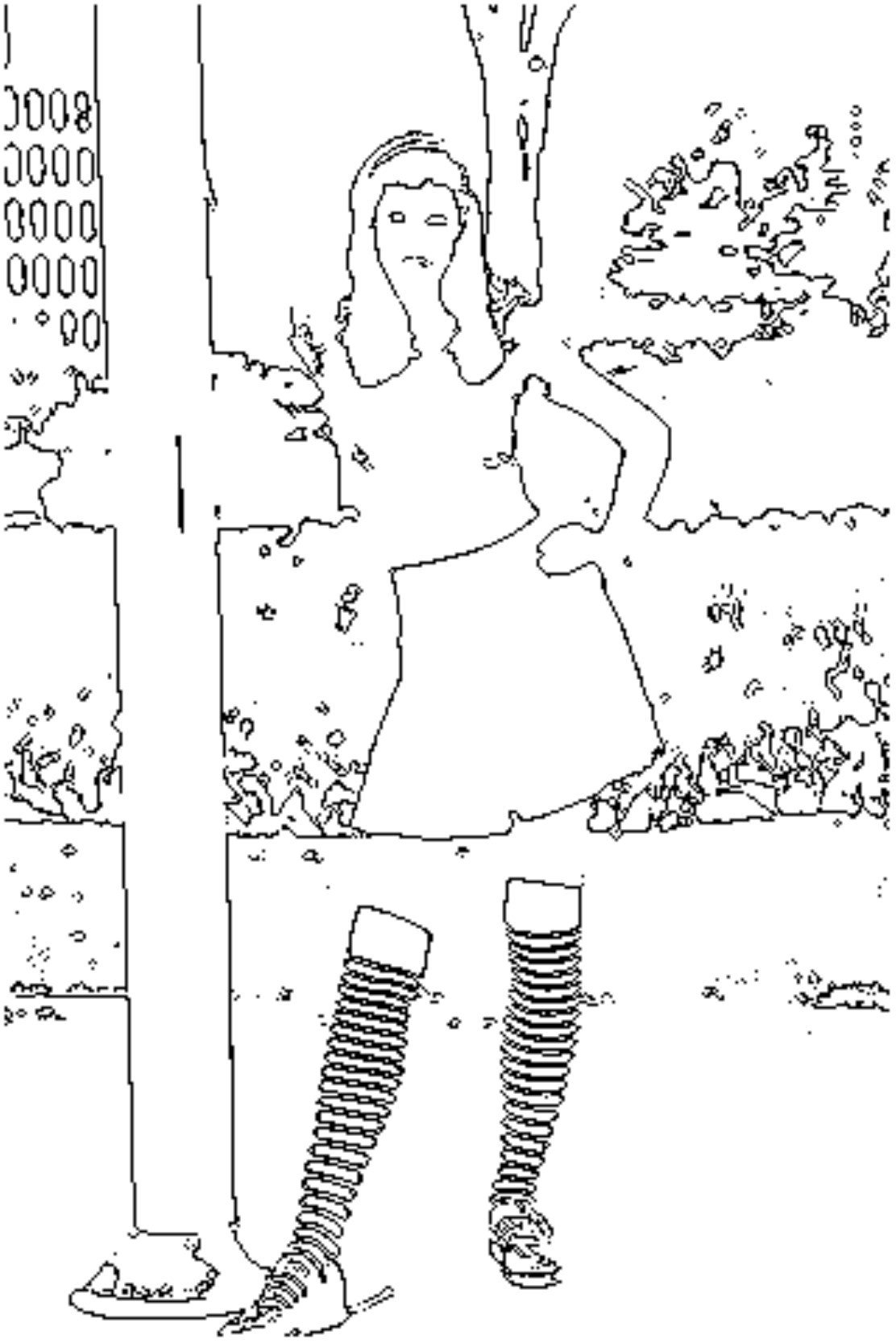}
\includegraphics[width=2.8 cm]{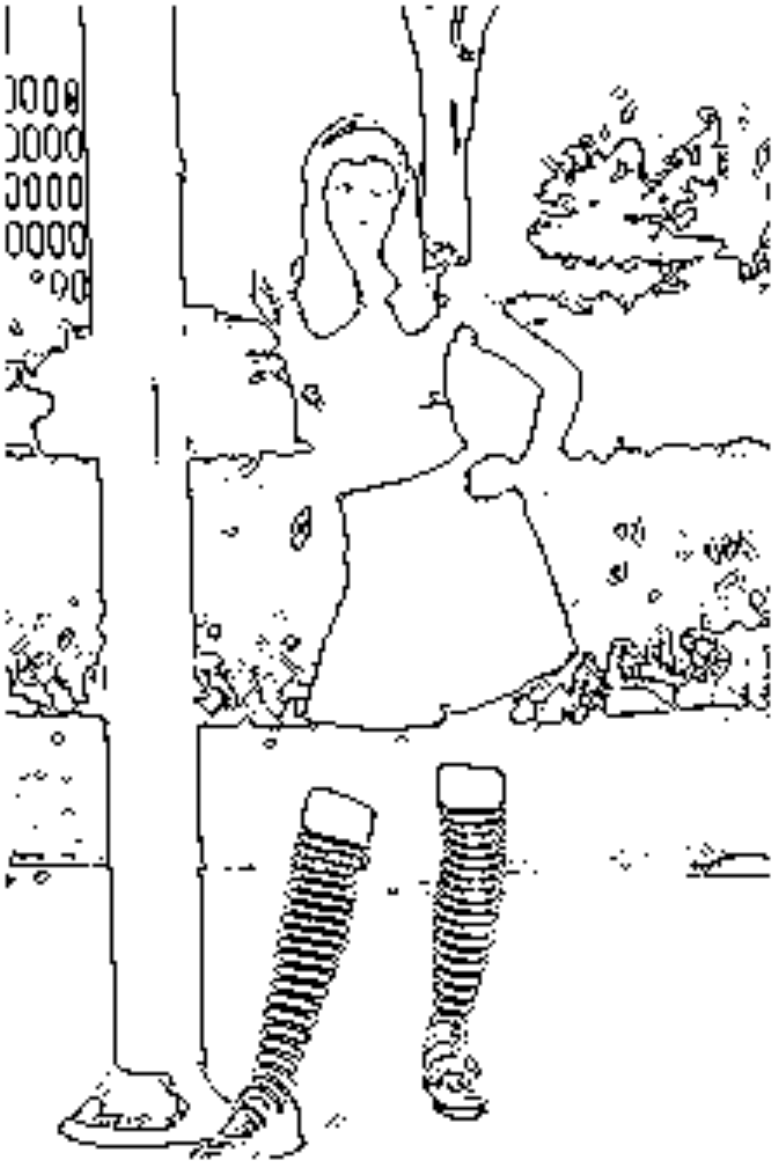}
\includegraphics[width=1.6 cm]{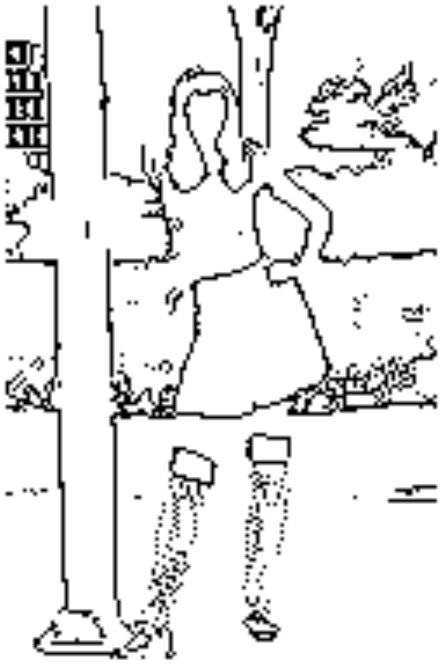}
\includegraphics[width=1 cm]{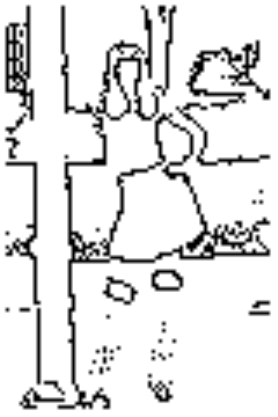}
\caption{Computed contours using the pattern $P_5$ for the scaled images}\label{fig74n}
\end{figure}

\begin{figure}[ht!]
\centering
\includegraphics[width=4 cm]{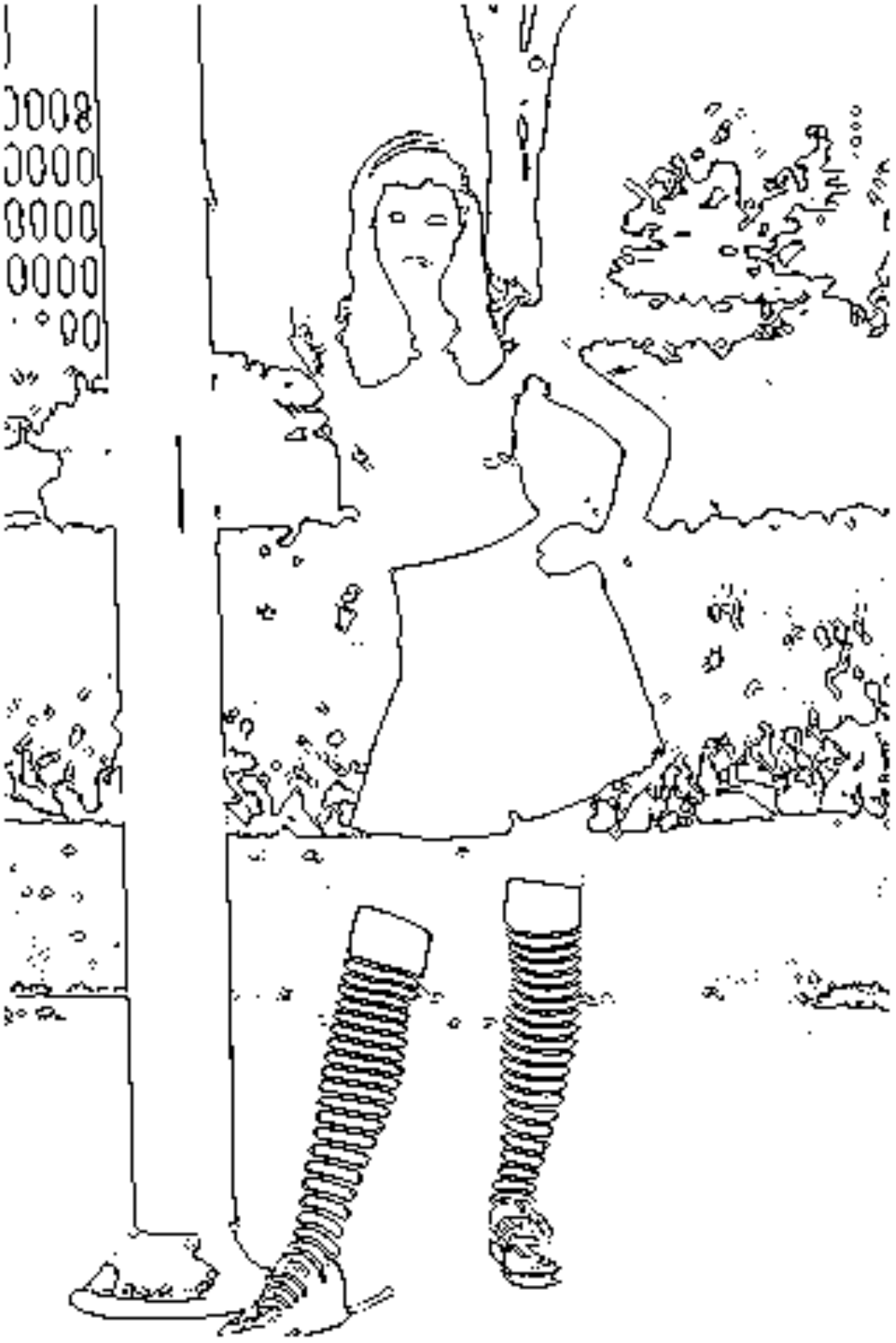}
\includegraphics[width=2.8 cm]{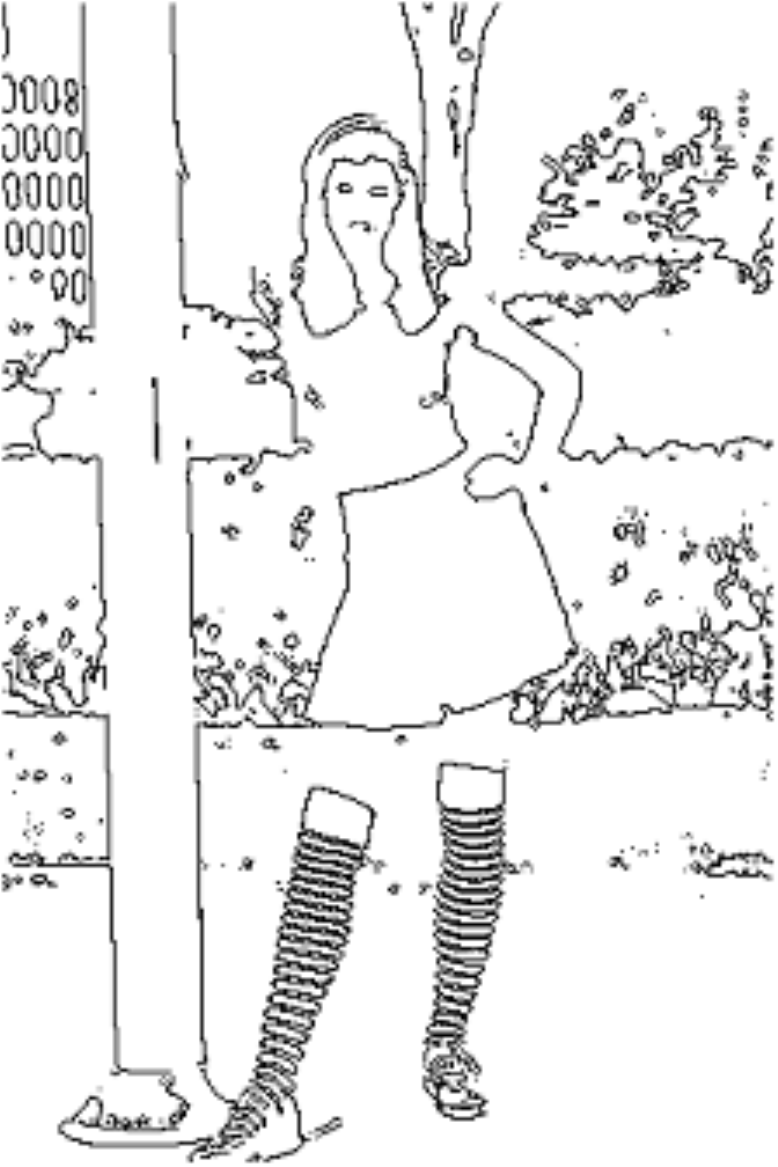}
\includegraphics[width=1.6 cm]{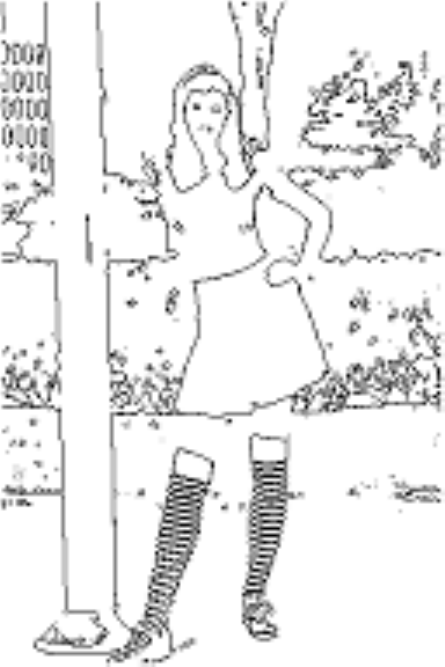}
\includegraphics[width=1 cm]{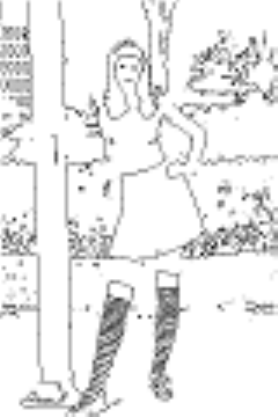}
\caption{From left to right: Computed contours using the pattern $P_5$, Scaled contours of the original image to 70\%, 40\%, 25\%}\label{fig78n}
\end{figure}

For scaling invariance study, we applied contour detection after scaling images to $90\%$,
$80\%$, $70\%$, $50\%$, $40\%$, $25\%$, $12.5\%$ of its original size (see figures \ref{fig70n}, \ref{fig74n}, \ref{fig78n}).

The scaling has been done using "paint" software and we can see that under 50 the quality of
image is degraded.
We measured the ratio of contour pixels that haven't been located after image scaling related
to the scaled contours of original image. Experiments have been conducted on BSD500 data set
and the average of this ratio computed from all images obtained using the $14$ patterns on BSD
dataset images gave:  $0.04, 0.05, 0.09, 0.17, 0.26, 0.33$ corresponding to scaled original
image to $80\%, 70\%, 60\%, 50\%, 25\%, 12\%$.

Despite that the used images are scaled with software, we can see that above $50\%$, our method
is invariant to scaling. Below of $50\%$, images are very degraded and thus locating contours
will produces a missing of original pixels contours due to missing of information.



\section{Evaluation and Discussion}
\label{Evaluation}

Many evaluation measures have been proposed for boundary quality and for all them there is a
necessity to have the ground-truth data represented by the correct pixels contours drawn by hand.
Estrada et al \cite{Estrada and Jepson 2006} considered this measure as related to the quality of the
segmentation of the image induced by the computed contour where the best boundary neatly separates two
visually distinct regions of the image. The error measure based on the average distance between boundary
pixels from two contours is used. The inconvenience of this measure is that is necessary to match all
computed outlines and all the reference ones.

Other measure proposed by Martin et al \cite{Martin et al 2004,Arbelaez et al 2011} considers
the ratios - precision, recall - computed using the numbers of pixels found in the automatic contours
vs the correct (hand-drawn) ones.
This measure may do a good job of estimating the quality of found contours only if two conditions are
verified otherwise it gives a false estimation of the quality. The first condition is that both the computed
and reference contours must be of the same resolution which means that if the method locates only a specific
resolution of contours (levels in our scheme), the used hand drawn contour must also have the contours at
that resolution. The second condition is that the task of reference contours drawing must be done
with high accuracy and all possible contours must be located for the ground truth data.

As example, we cite the data set BSD500 \cite{Arbelaez et al 2011} where for each image, five hand drawn
contours are available. Our opinion is that such reference
images must be used as benchmark for segmentation algorithms rather than contour detection algorithms
because only shapes are segmented and some true contours are not located even if they separate
two visually distinct regions.
In our experiments, we built our reference images for the BSD 500 dataset. 

\subsection{Visual estimating of the quality of contour detection: some samples}

We give in this subsection some results obtained by our method for visual comparison with the results obtained
by the algorithms of Berkeley \cite{Arbelaez et al 2011} and Canny \cite{Canny 1986} performed on
BSD500 data set.

The reader can easily  identify on the illustrated results which is the best computed image of contours.
The strong advantage of our method is that the computed contour pixels are located only when there
is a transition between two regions. Depending of the resolution used (high, intermediated or low),
the precision of located contours by our method is better than the precision of  located by Berkeley
\cite{Arbelaez et al 2011} and Canny \cite{Canny 1986}.
Indeed, Canny's algorithm depends on the used thresholds and then when there are more pixels of contours,
there will be consequently more and more false candidates.
Concerning the method of Berkeley \cite{Arbelaez et al 2011}, many good candidates are missing and then is
 more suitable for image segmentation. Indeed, the making of BSD data set has been done by subjects
 satisfying this criterion. Only the outlines of objects are drawn. Consequently, this algorithm
 has outperformed all contours detection algorithms on this data set.

\begin{figure}[ht!]
\centering
\includegraphics[width=5cm]{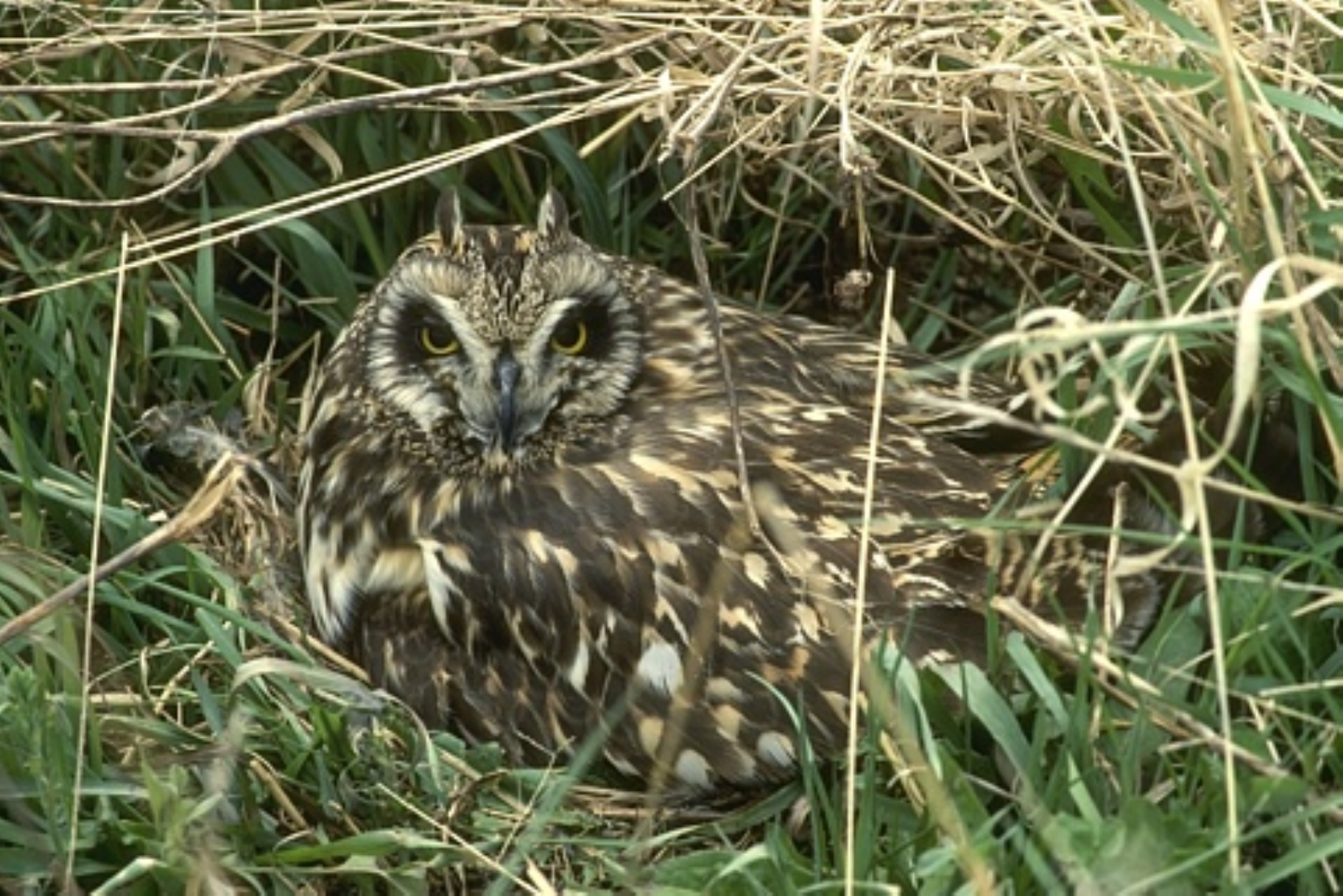}
\includegraphics[width=5cm]{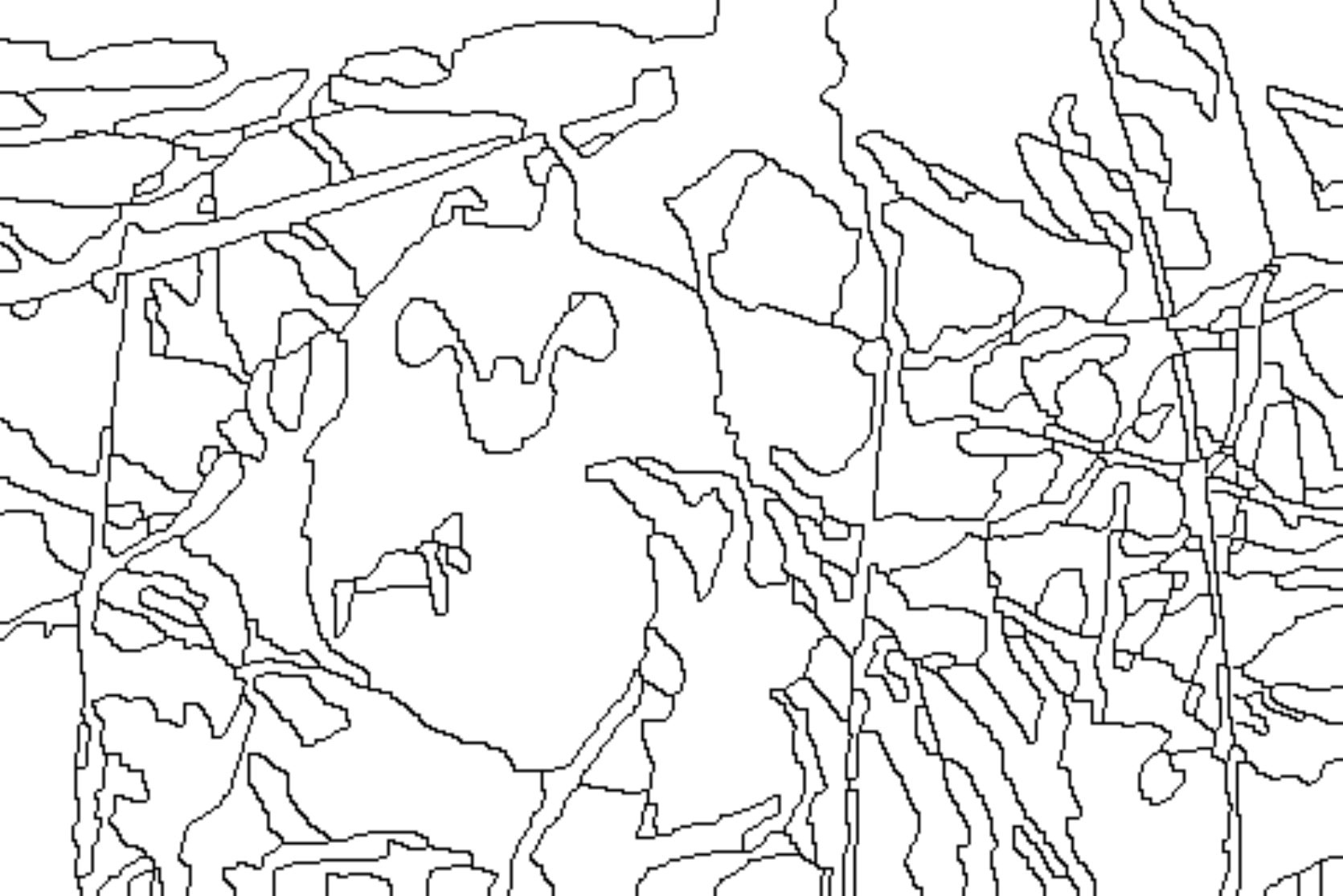}
\caption{Image from BSD500 data set, Located contours by Arbelaez algorithm}\label{fig99n}
\end{figure}

\begin{figure}[ht!]
\centering
\includegraphics[width=5 cm]{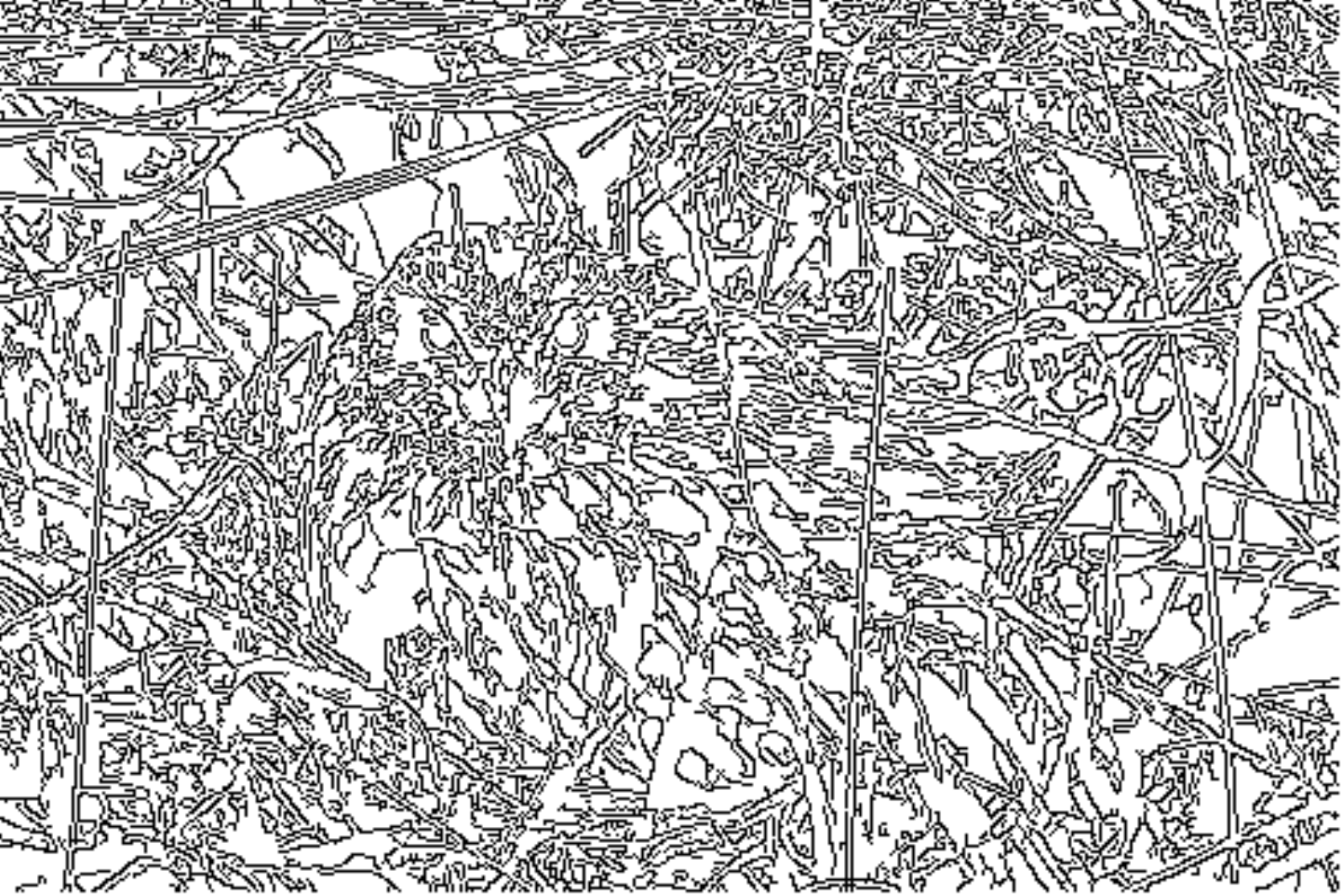}
\includegraphics[width=5cm]{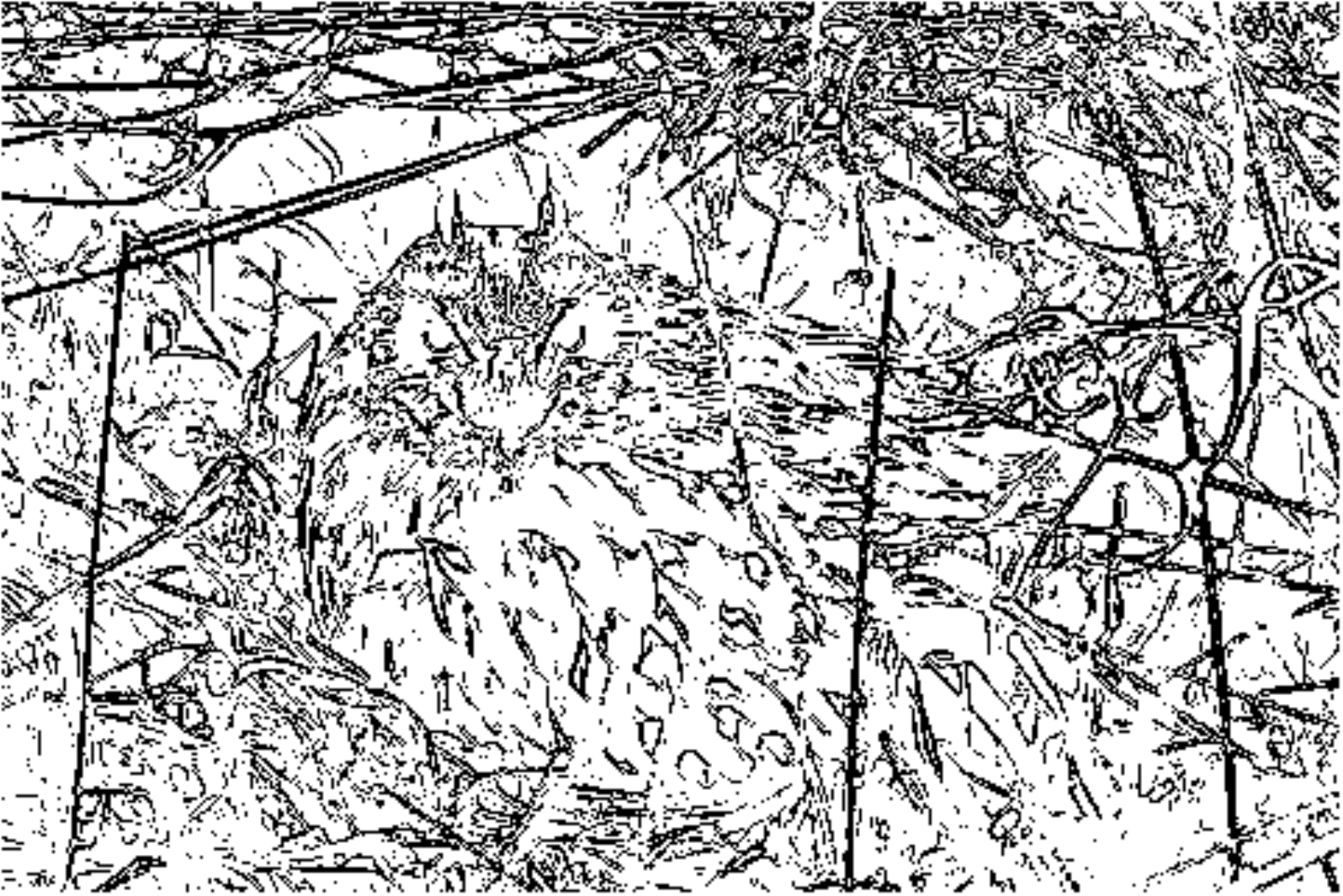}
\caption{Located contours by Canny algorithm with high threshold equals to $300$,
Located contours by our method with intermediate resolution}\label{fig101n}
\end{figure}

For example, seeing to figures \ref{fig99n}, \ref{fig101n}, the owl is clearly more recognizable
from our contours than perceiving other results. On the canny result, there are many false candidates
and the image of contours looks like a random dot stereogram. For Berkeley result, many contours are missing.

\begin{figure}[ht!]
\centering
\includegraphics[width=5cm]{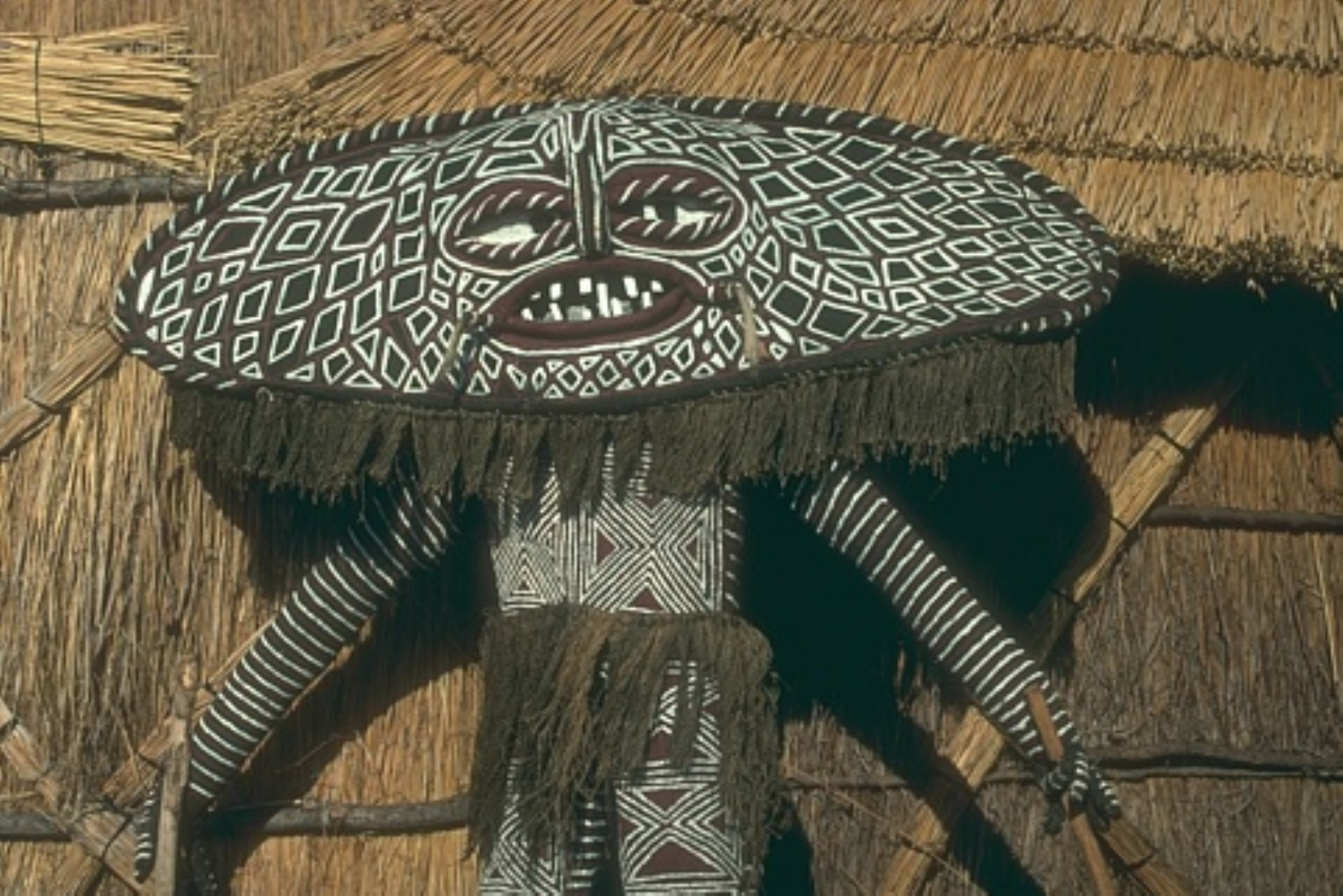}
\includegraphics[width=5cm]{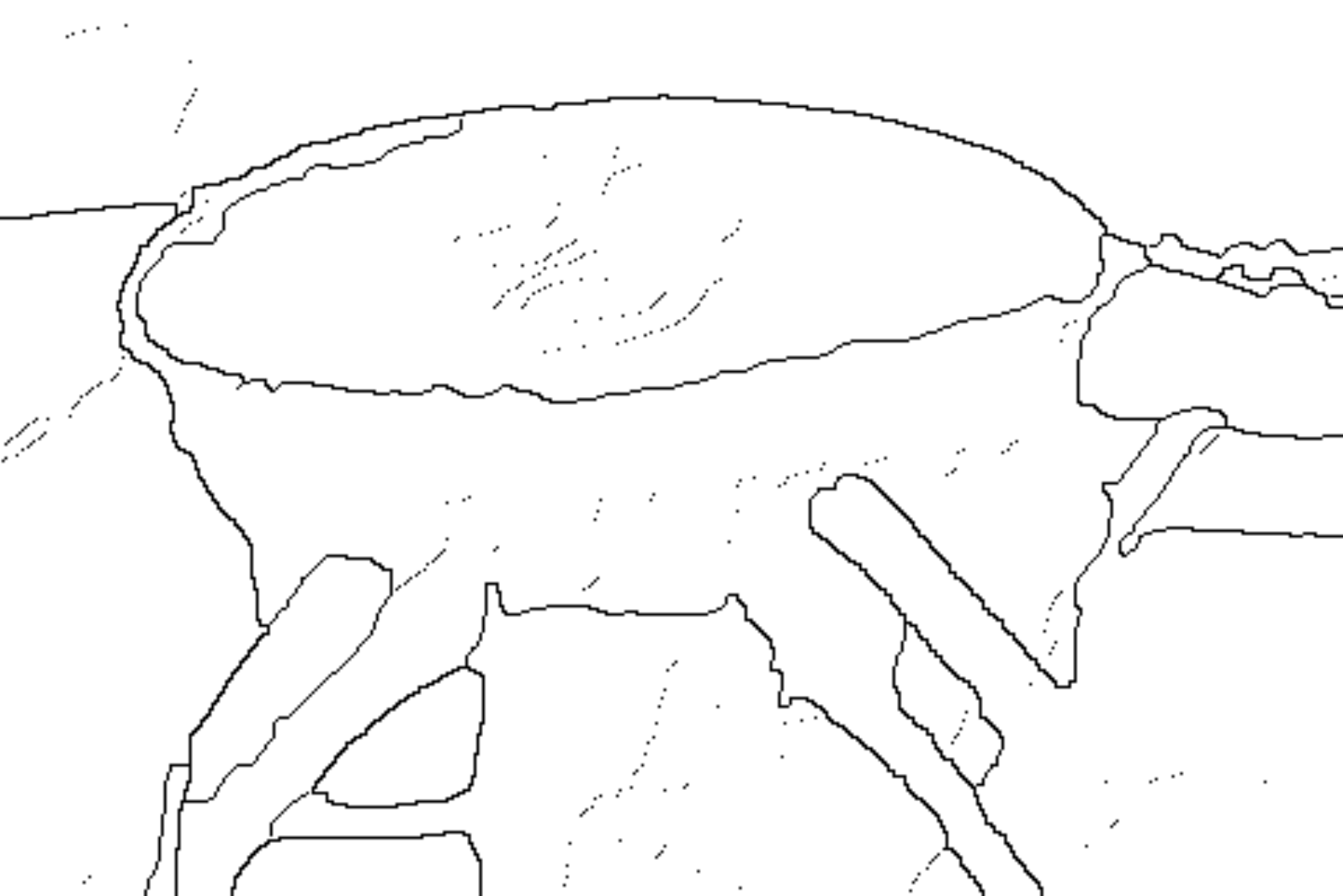}
\caption{Image from BSD500 data set, Located contours by Arbelaez algorithm}\label{fig91n}
\end{figure}

\begin{figure}[ht!]
\centering
\includegraphics[width=5 cm]{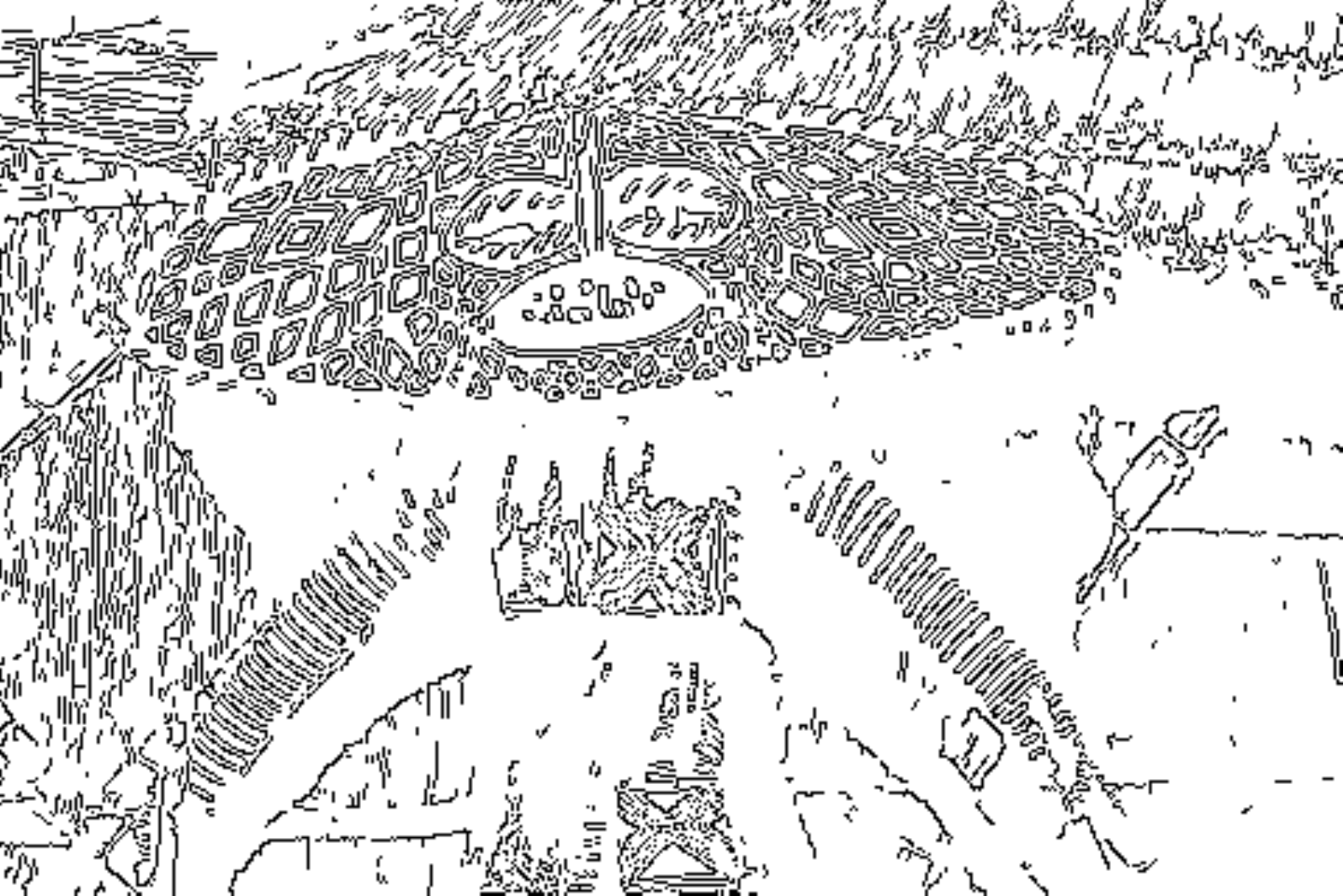}
\includegraphics[width=5cm]{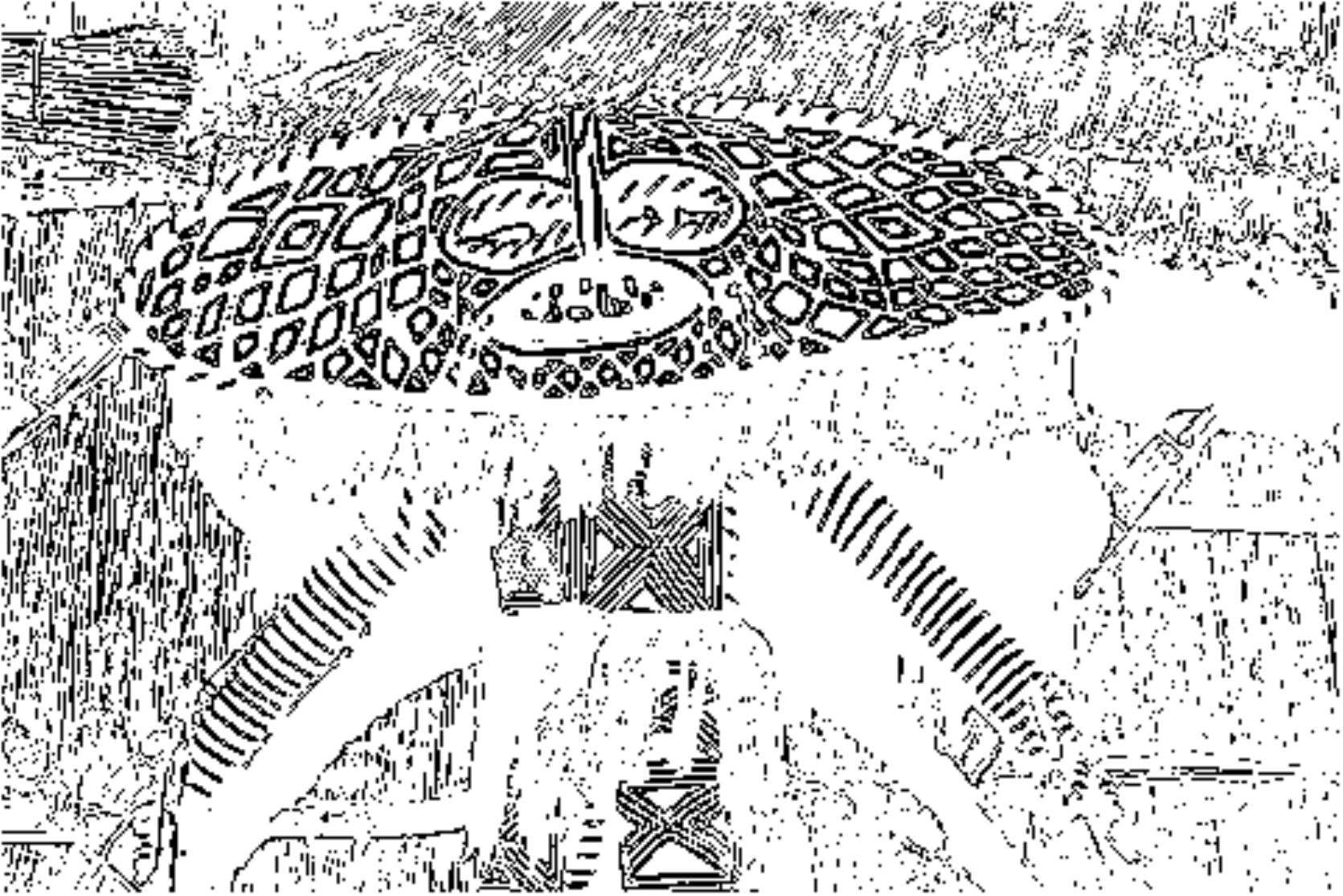}
\caption{Located contours by Canny algorithm with high threshold equals to $300$,
Located contours by our method with intermediate resolution}\label{fig93n}
\end{figure}

\begin{figure}[ht!]
\centering
\includegraphics[width=5cm]{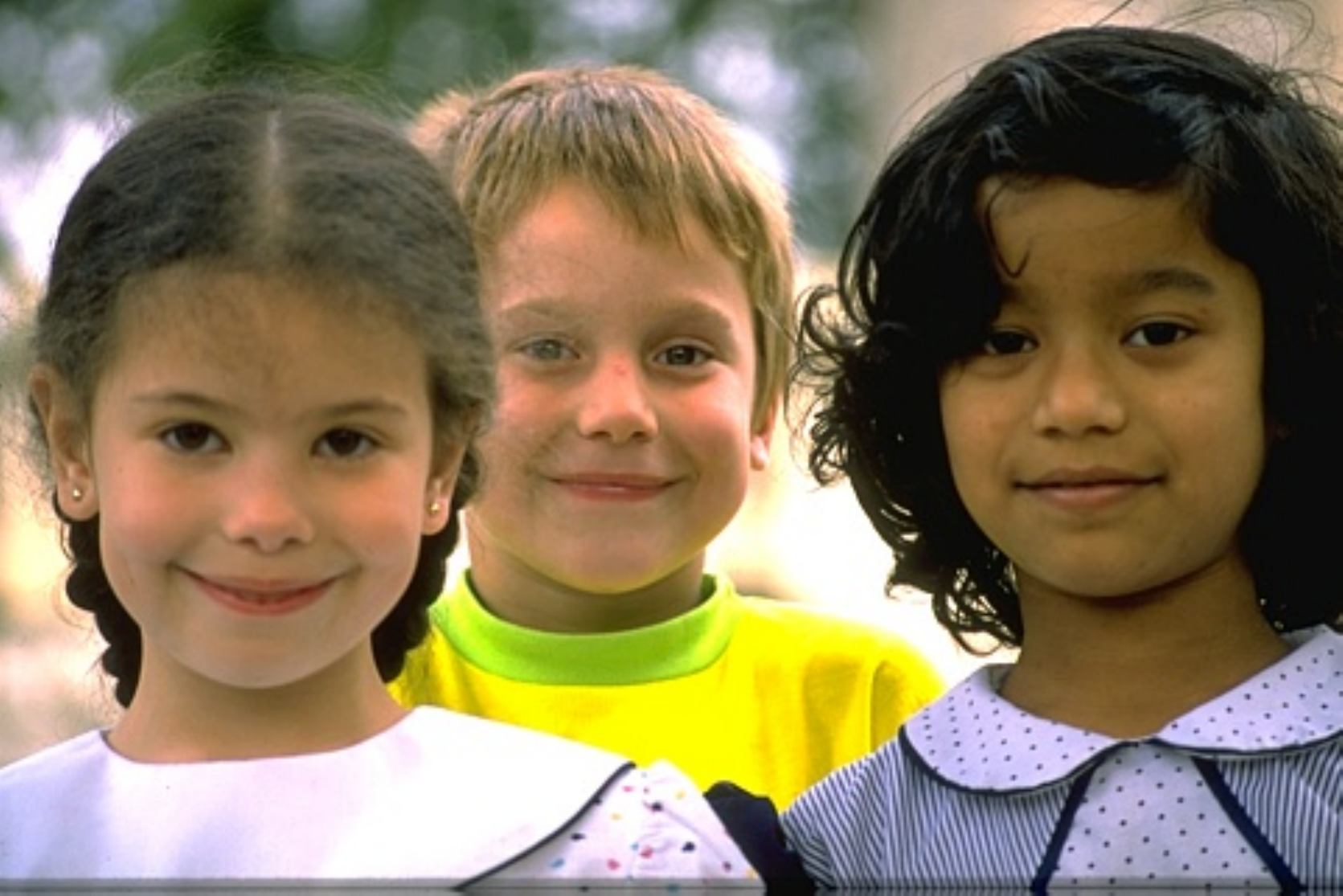}
\includegraphics[width=5cm]{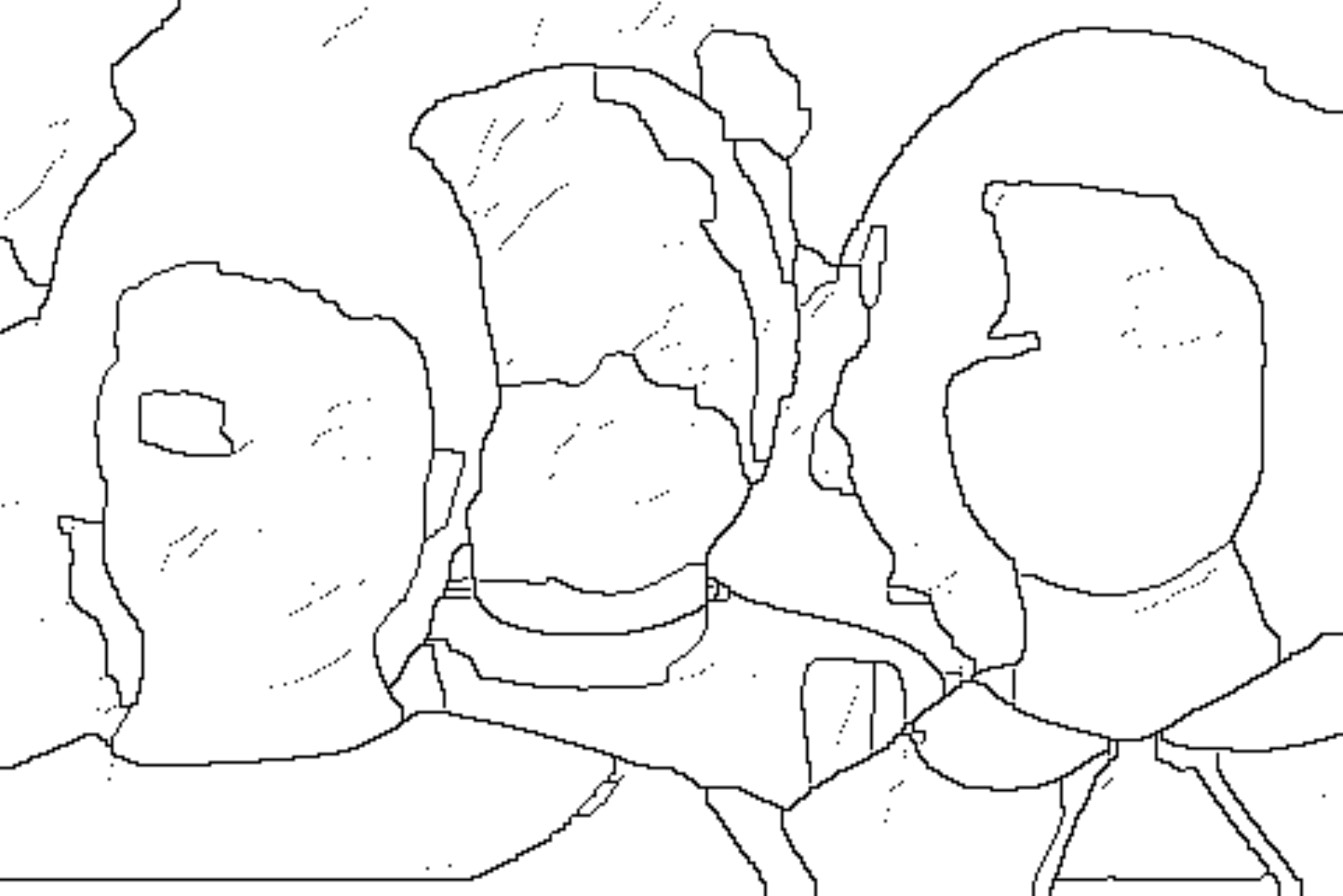}
\caption{Image from BSD500 data set, Located contours by Arbelaez algorithm}\label{fig95n}
\end{figure}

\begin{figure}[ht!]
\centering
\includegraphics[width=5 cm]{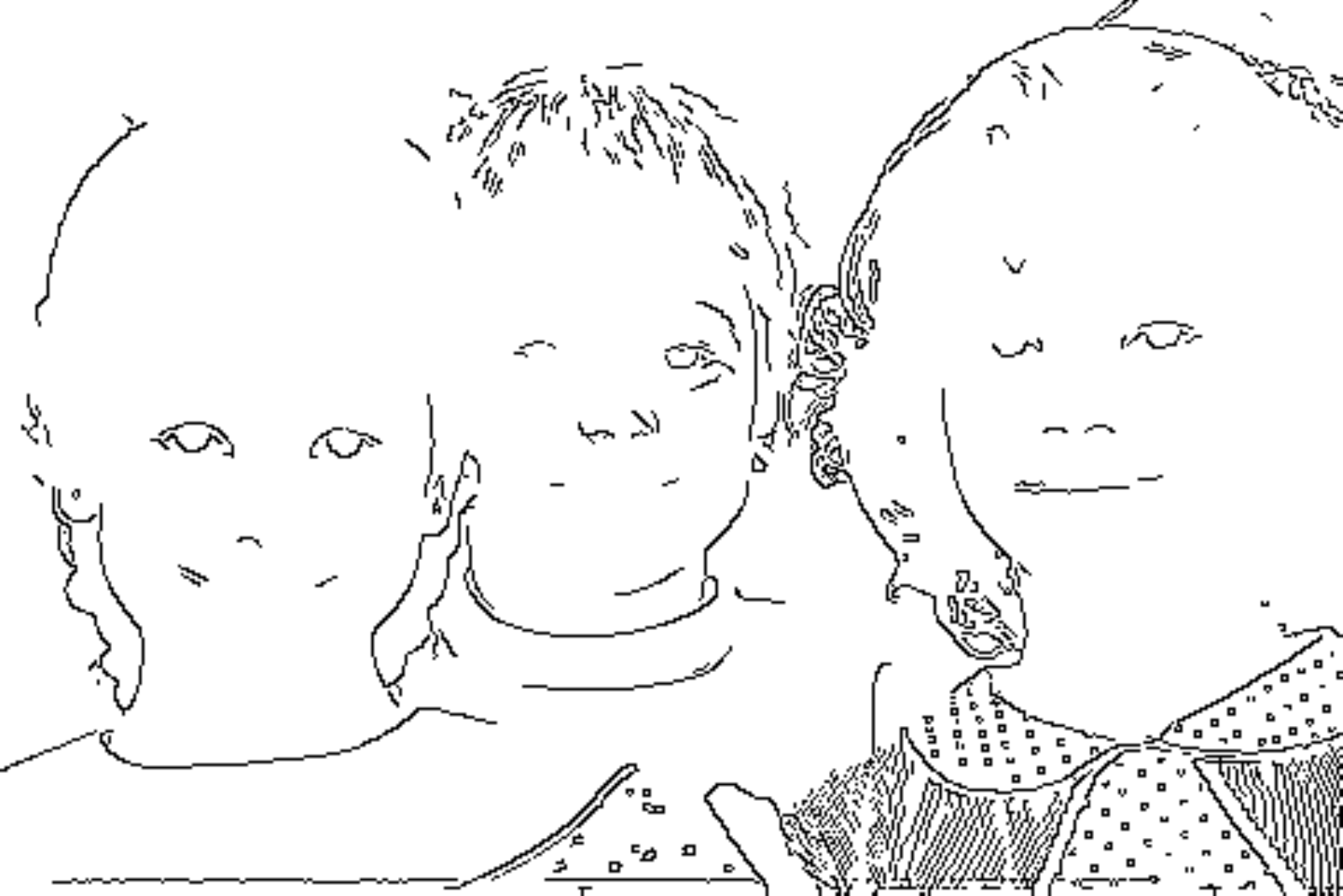}
\includegraphics[width=5cm]{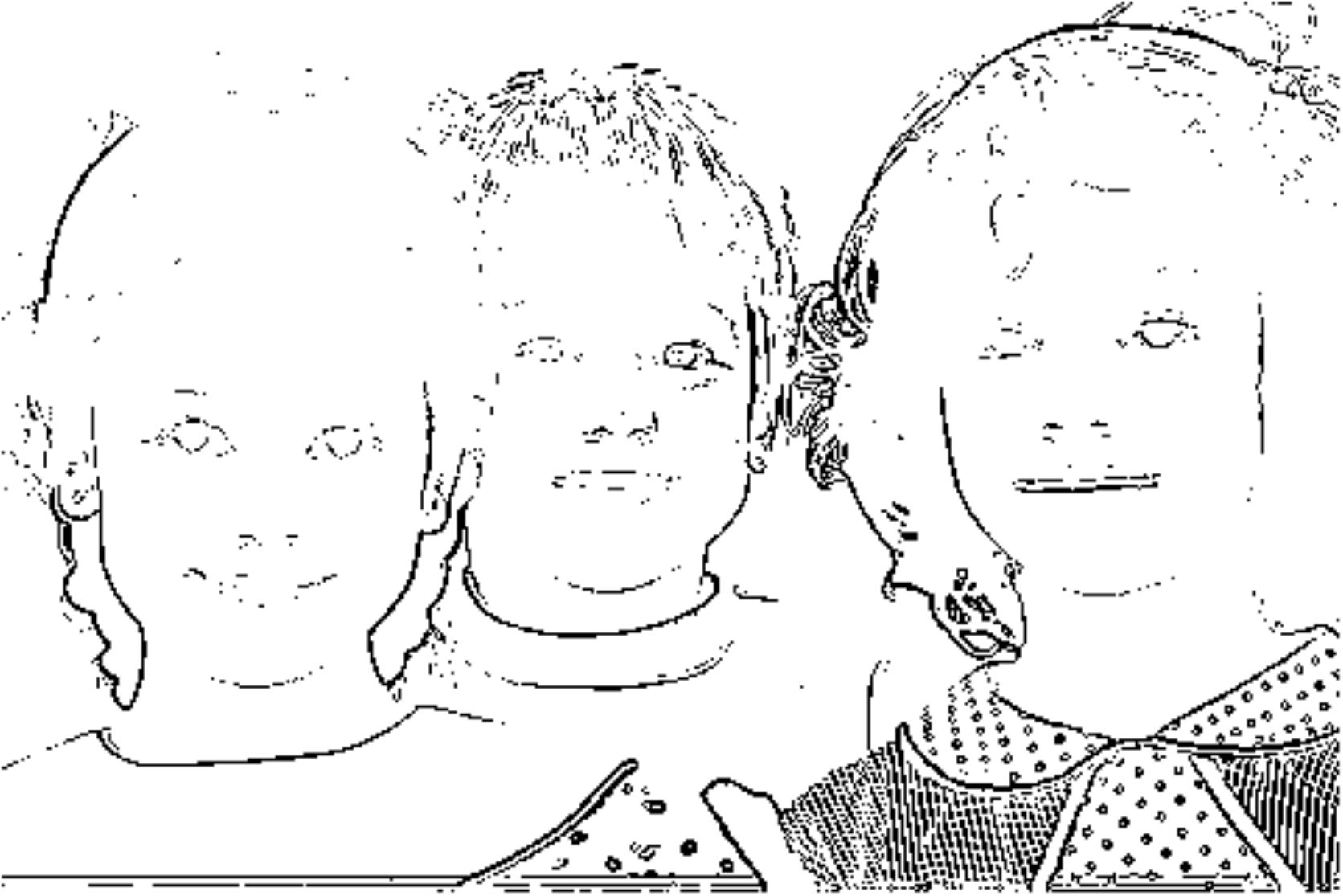}
\caption{Located contours by Canny algorithm with high threshold equals to $300$,
Located contours by our method with intermediate resolution}\label{fig97n}
\end{figure}

The canny method fails to detect texture for some images of BSD 500. In figures \ref{fig91n}-\ref{fig93n},
our method localizes the texture on the center of the image with high precision, however Canny fails
completely. The same case is repeated for another image as illustrated by figures \ref{fig91n} to \ref{fig97n}
where Canny encounters a problem for locating the texture at low left of the image.
For the two images, Berkekey does not locate all textures.

\subsection{Evaluating contour detection quality on BSD500 data set}

We compared our results to those obtained by Canny's and Berkeley's methods. Canny is selected
because it uses only the low level feature (intensity), and Berkeley is selected because it
uses some mid-level features (patch descriptor, texture, histogram). Similar results to those of
P. Arbelaez \emph{et al.} \cite{Arbelaez et al 2011} are obtained by some algorithms which have used mid-level and high level
information in order to locate the outlines of objects such as
\cite{Borenstein et al 2008, Donoser et al 2010, Zhang et al 2013} and \cite{Payet and Todorovic 2013}
which is contour-based method starts from located contours, locates boundary and performs their grouping.
The results of these methods are then have not been considered.

For the BSD500 data set, we computed the Precision and Recall and we obtained best results than those
of gPb method of Arbelaez \emph{et al.} \cite{Arbelaez et al 2011} and Canny method \cite{Canny 1986}.

\begin{figure}[ht!]
\centering
\includegraphics[width=6cm]{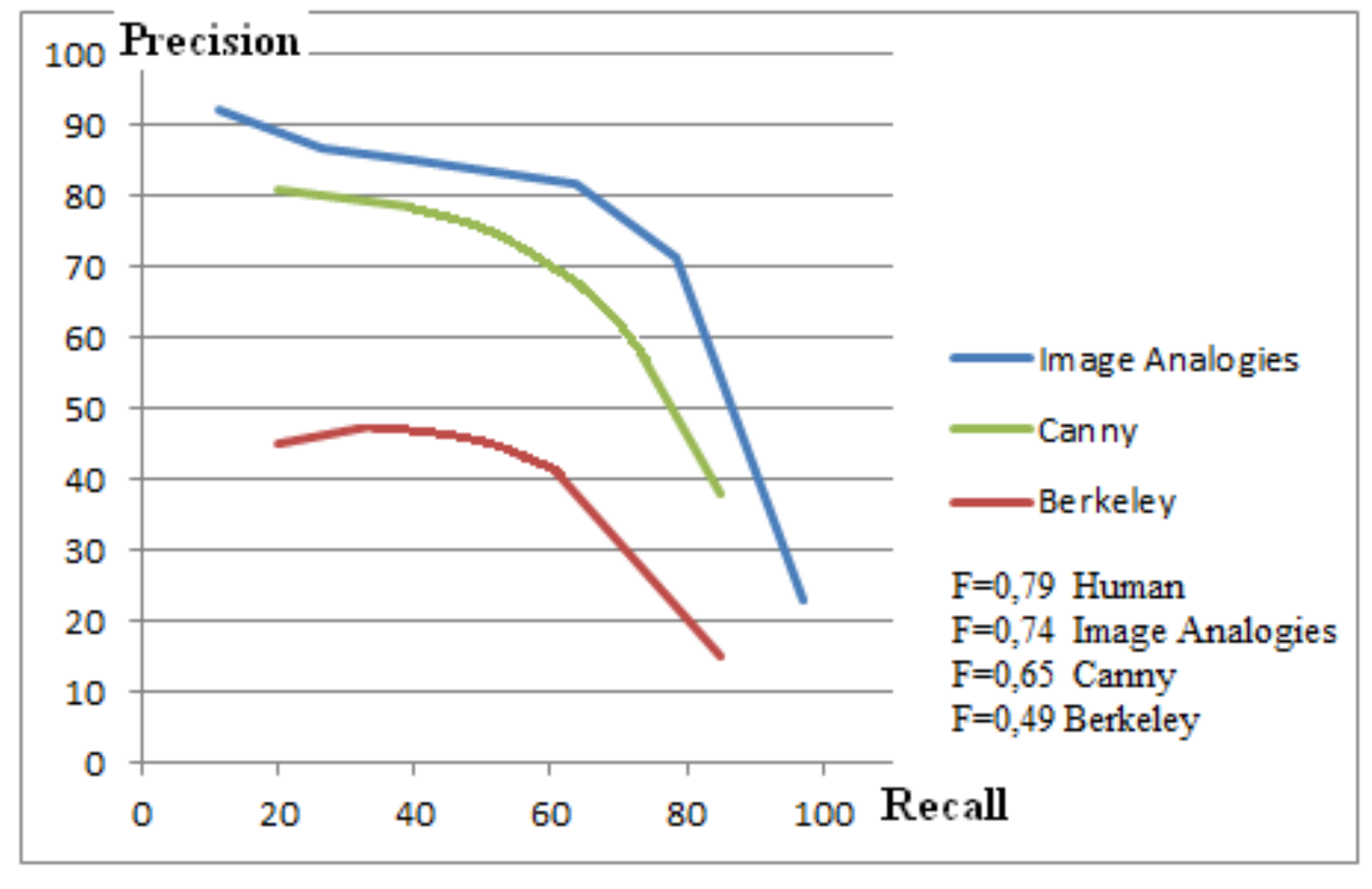}
\caption{Recall and Precision of contour detection approaches on BSD500 dataset with respect to human ground-truth boundaries}
\label{fig86n}
\end{figure}

The maximum of $F-measure=2.Precision.Recall/(Precision+Racall)$ for our method is equal to $0.74$
which is greater than Canny with $0.65$ and Berkeley with $0.49$.
\cite{Arbelaez et al 2011} (see figure \ref{fig86n}).
This superiority is due to two factors. The first one is that our method locates all pixels of contours
such as do Canny, however, we outperform Canny which depends on the thresholds. If more contours are
located, the precision decreases, with the increase of Recall. If less contours are located, Recall
decreases and Precision increases.
The second one is the used ground truth made by drawing all contours. This new data set cannot produces
good results for the methods which located outlines than contours.


\subsection{Evaluating contour detection quality on Weizmann Horses data set}

\begin{figure}[ht!]
\centering
\includegraphics[width=6cm]{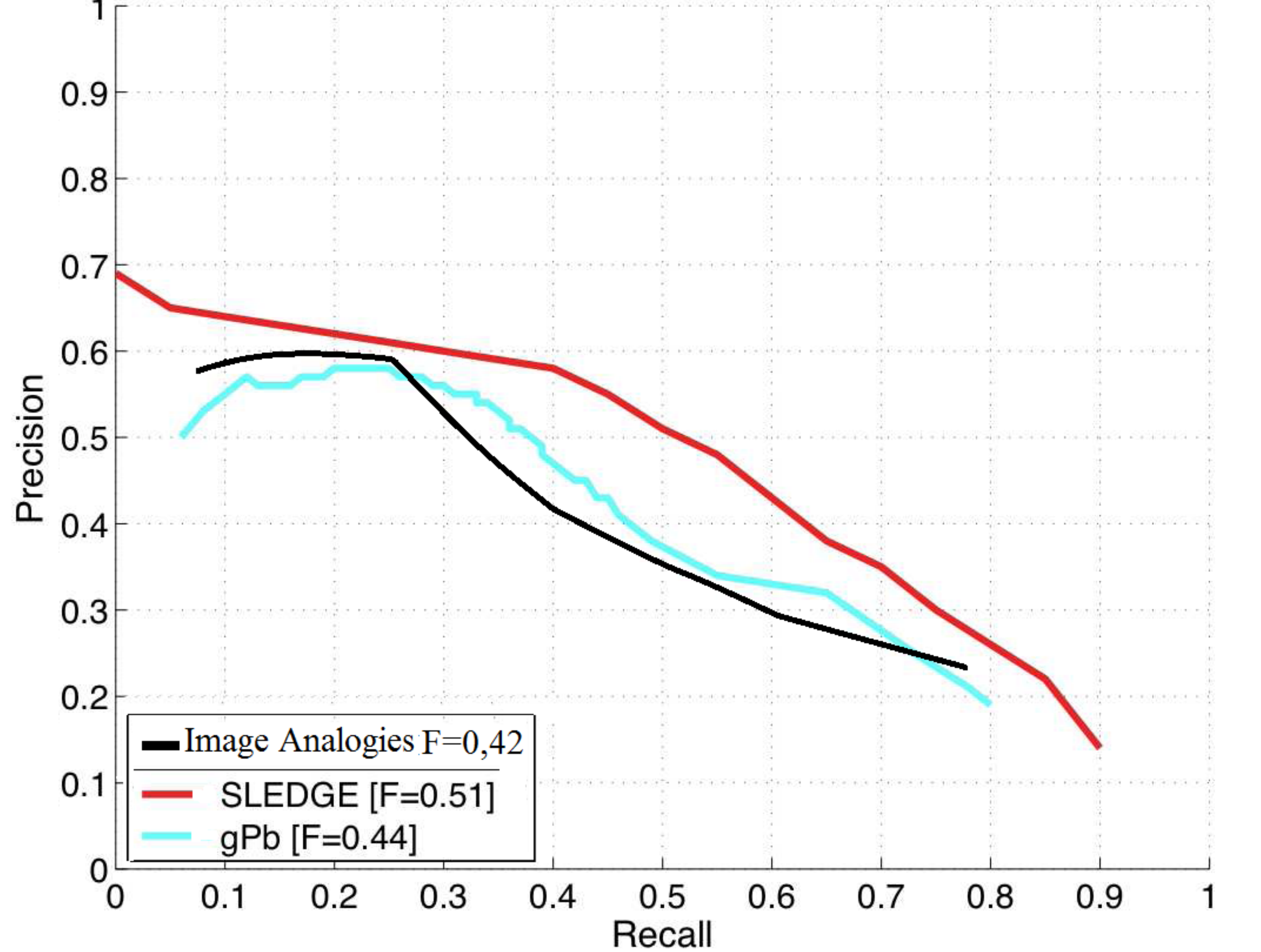}
\caption{Recall and Precision scores obtained for Weizmann data set}
\label{fig87n}
\end{figure}

We repeated the same experiments for Weizmann Horses data set \cite{Borenstein and Ullman 2002} and our results are compared to
the results of Payet and Todorovic \cite{Payet and Todorovic 2013} and gPb method \cite{Maire et al 2008}
(see figure \ref{fig87n}). We note that for the ground truth data only the horses are located.
Consequently, methods which locates outlines than contours are favourites to obtain best scores
of (Recall, Precision). The obtained results will be better if all contours are located in the
ground truth data.

\begin{figure}[ht!]
\centering
\includegraphics[width=4cm]{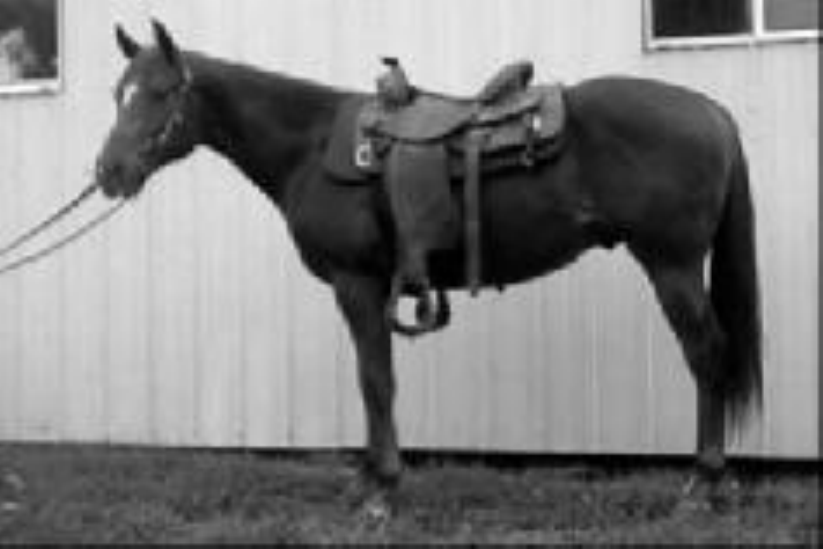}
\includegraphics[width=4cm]{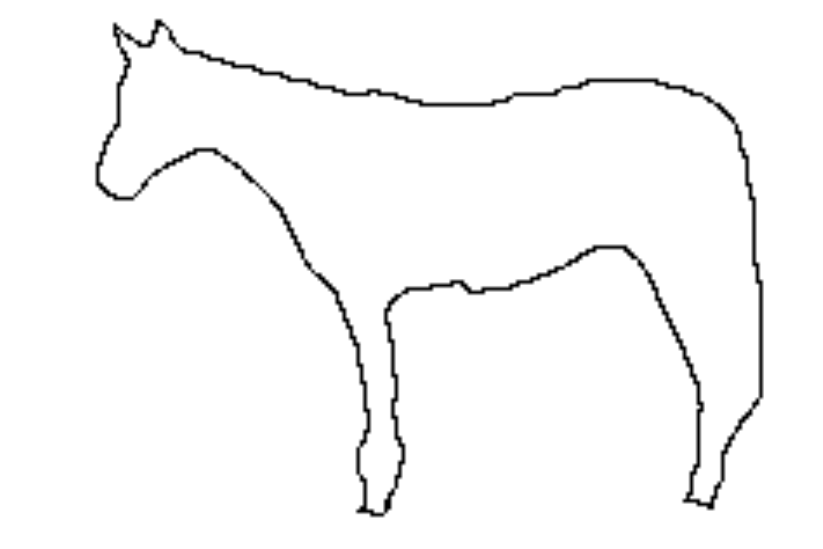}
\includegraphics[width=4cm]{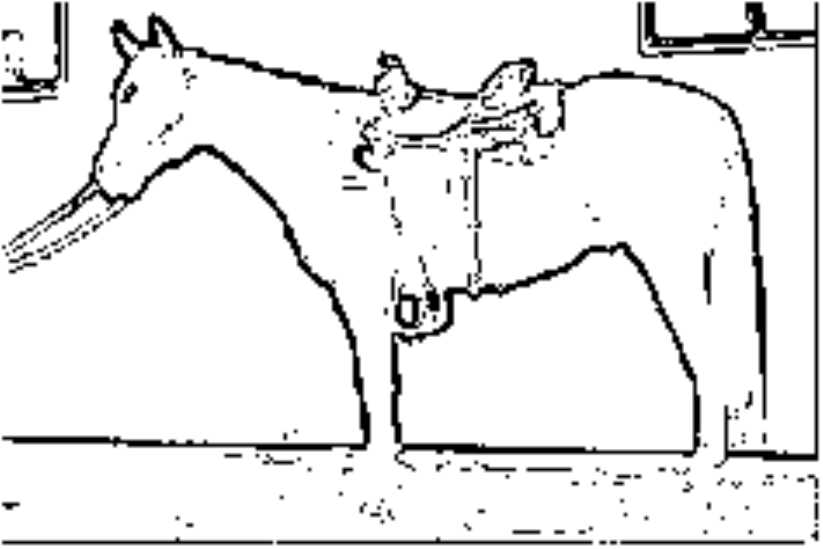}
\caption{First image from Weizmann Horses data set, Ground truth data, Our computed contours}\label{fig110n}
\end{figure}

\begin{figure}[ht!]
\centering
\includegraphics[width=4cm]{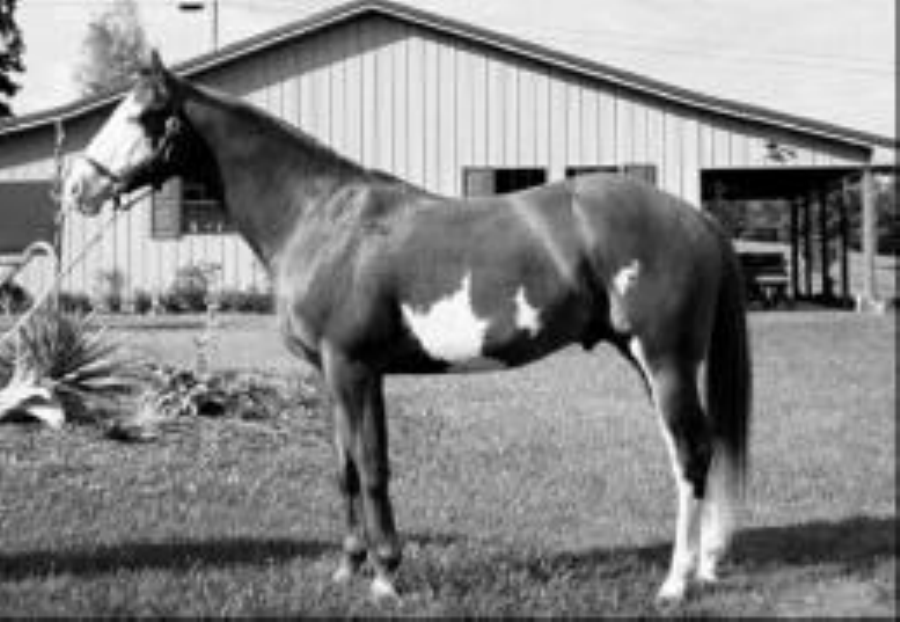}
\includegraphics[width=4cm]{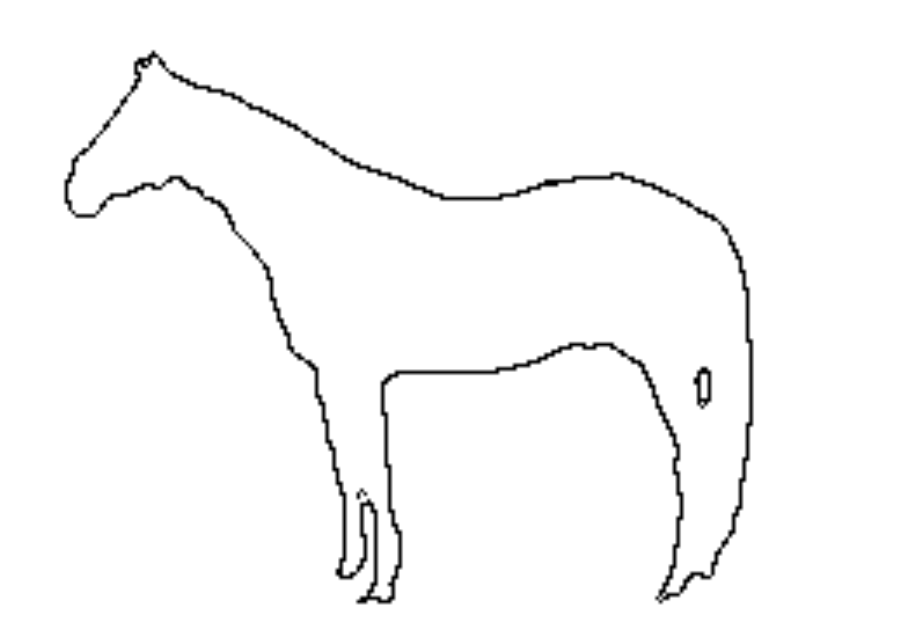}
\includegraphics[width=4cm]{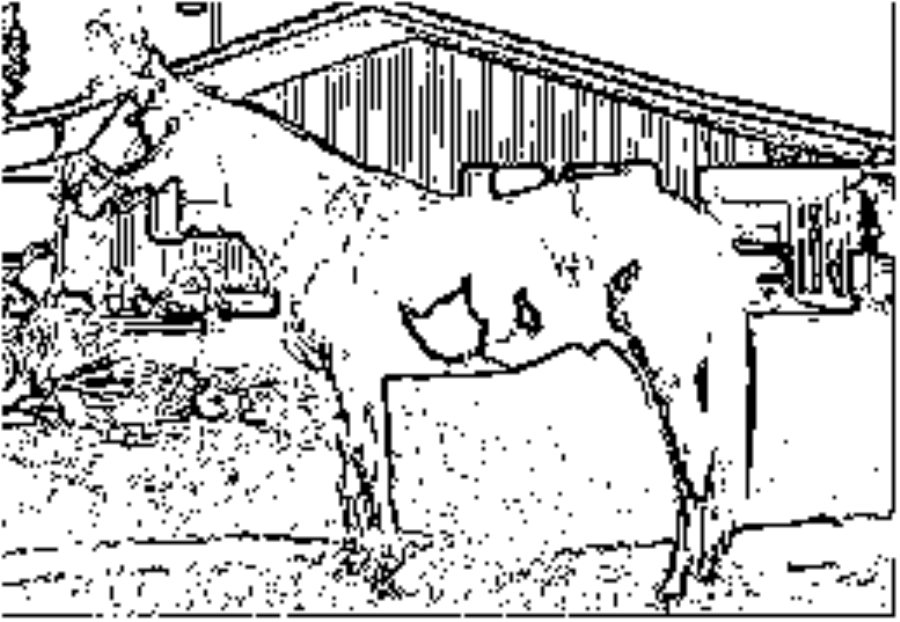}
\caption{Second image from Weizmann Horses data set, Ground truth data, Our computed contours}\label{fig113n}
\end{figure}

Figure \ref{fig110n} illustrates a sample of images from this data set and the
reference contours are made by locating only horses from background. Our computed contours are
precise. However, as all pixels are counted in the evaluation of the quality measure, most of the values
obtained (Recall, Precision) for all images  are under to those obtained by SLEDGE and gPb methods.
This low of performance is due to numerous contours located of the background in addition to internal
ones of the horse (see figures \ref{fig110n}, \ref{fig113n}).

\subsection{Discussion}

The resolution level of contours must be defined in order to make suitable benchmarks for comparison
of contour algorithms. In the step of reference image making, the human completes some parts of outline
shapes that are not neatly visible because he uses his/her prior knowledge about the shape geometry and
locates the perceived contours without specifying their levels. That is, it is a cognitive process based
around shape extraction, not contours \emph{per se}.

In this work the level of contour (as stated above) is related to the difference of
intensity between neighboring regions. In our comparison, we used reference images where contours are hand
drawn by human subjects without specifying the level of contours. In this case, we computed the contours for
level $1$, $2$, $3$ and $4$ using $14$ artificial patterns for the computation of Recall and Precision.

Our method cannot localize contours when the human himself cannot (note we are not talking about inferring
the presence of a shape, but an actual contour). This occurs when the part of shape and the background have
the same color or intensity. Such cases have decreased the performance of our method. Even if all these
contours may be localized at the level $1$, they cannot be exploited due to the high number of contour
pixels found for this level.



\section{Conclusion}

We proposed in this paper a new method for contour detection based on image analogies.
In the first part, we studied the possibility to use of hand-drawn contours as reference
images for the detection of new contour pixels by analogy in the query image.
We found that only pixels that have the same conditions as those of the reference image may be located
which is perhaps to be expected in such a data driven technique. This implies that numerous reference images
are needed to locate all possible new contour pixels which implies in the hard and time-consuming task of
hand drawing reference contours, and thus increasing of the algorithm complexity.

Instead to apply directly image analogies, we investigated in this work, how can we avoid this
constraint in order to guaranty that all contour pixels
will be located for any query image. Fourteen derived patterns are sufficient to be used as training images
(instead of real images alone) to locate contour location at different scales independently of the light
conditions present in the real images.

To avoid this constraint and to locate all contour pixels whatever the image query happens to be, we
proposed a set of $14$ artificial pairs of patterns as reference images of low size and containing
the required information to locate contours of different levels of resolution where levels are related to the
difference of intensity between neighboring regions.

The proposed method has been applied to different types of images: the ``natural'' BSD dataset, Horses
of Weizmann data set. Compared to the reference images, our
method demonstrates a very good recall, precision and finds all visible contours of gray level images.

\subsection{Further work}

We note that, as the technique is based only on intensity attributes, some contours separating colored
regions that are visible are not located due to the close values of region intensity. Because our approach
uses only intensity, we are confident that including color attributes in the similarity measure will increase
the performance of the method and decrease the failure modes.

Also, an interesting work will be the exploitation of the set of contours computed with the artificial
patterns for image segmentation. Indeed, seeing the computed contours from one pattern to another, we
can notice that there is a slow motion of contours around the region boundaries.

 it will be interesting the modeling what human do in similar way.


\section{Appendix: Constraints required for images training}\label{annex}

We study, given the proposed similarity measure, what are the required constraints in the training
images in order to locate all and only contour pixels.

We consider the general case where intensity in image is not uniform around the boundary
and we will take into account complicating factors such as gradually changing shading or texture.

We distinguish the two cases where the pixel $q$ in the query image B is a contour pixel or not .

\textbf{case (1)}: \textbf{$q$ isn't a contour pixel}\\

Under this hypothesis, we study if the selected pixel $p^*$ could be a contour pixel?

\begin{figure}[ht]
\centering
  \includegraphics[width=10cm]{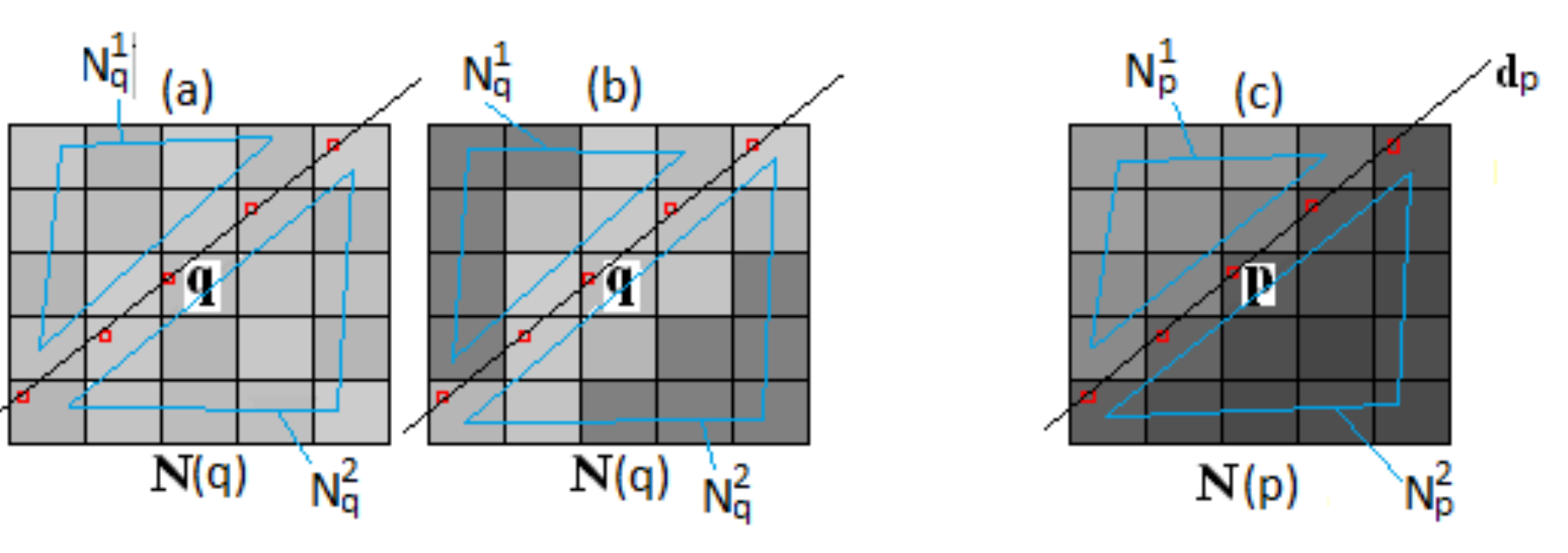}
  \caption{(a) $q$ isn't a contour pixel and $N(q)$ doesn't contain contour pixels
  (b) $q$ isn't a contour pixel and $N(q)$ contains some contour pixels, (c) $p$ is a contour pixel}
  \label{fig19n}
\end{figure}

Let $d_p$ be the line of contour pixels in $N(p)$) defining two regions $N_p^1, N_p^2$. $N(q)$ is also
divided into two regions $N_q^1, N_q^2$ following the same direction of the line $d_p$ (see figure \ref{fig19n}).

The similarity measure $S(q,p)$ between $N(q)$ and $N(p)$ is then given by the equation \ref{eq5}:\\
{ \footnotesize
\begin{equation}\label{eq5}
  S(q,p)=\sum_{i}\sum_{j}(N_q^1(i,j)-N_p^1(i,j))^2+\sum_{i}\sum_{j}(N_q^2(i,j)-N_p^2(i,j))^2
\end{equation}
}

Where $N_p^1(i,j), N_p^2(i,j)$ (respectively $N_q^1(i,j), N_q^2(i,j)$  are the intensities
of pixels (i,j) of  regions $N_p^1, N_p^2$ (resp. $N_q^1, N_q^2$).

As the pixel $p$ is a boundary pixel, there is a difference of intensity between the pixels
of regions $N_p^1, N_p^2$.
It is sufficient to find in $A$ the neighbor $N(p)$, such as $p$ is not a contour pixel, which produces
a similarity measure less than $S(q,p)$ obtained with $N(p)$ having $p$ as contour pixel.
Indeed, the value of $S(q,p)$ is greater than $S^1(q,p)$ obtained when $N(p)$ is taken entirely from a
region like $N_p^1$ ($N_p^2=N_p^1$) and either $N_q^2(i,j)<N_p^1(i,j)<N_p^2(i,j)$ or $N_q^2(i,j)>N_p^1(i,j)>N_p^2(i,j)$.
Also, the value of $S(q,p)$ is greater than $S^2(q,p)$ obtained when $N(p)$ is taken entirely from a
region like $N_p^2$ ($N_p^1=N_p^2$) and either $N_q^1(i,j)<N_p^2(i,j)<N_p^1(i,j)$ or $N_q^1(i,j)>N_p^2(i,j)>N_p^1(i,j)$.

Consequently, the $q$ pixel of $B$ can't be classified as a contour.\\


\textbf{case (2)}: \textbf{$q$ is a contour pixel}\\

In this case, we will study if the similarity measure $S(q,p)$ is minimal when the
pixel $p$ is a contour pixel. Otherwise, we must find the required conditions such that
a contour pixel $p$ will be selected as the best match for $q$.

As $q$ is assumed as a contour pixel, let $d_q$ the direction of the boundary in $N(q)$.
$N(p)$ is assumed to have any structure and the direction
$d_p$ of its boundary may be different from the direction $d_q$ of $N(q)$, the central
pixel $p$ could not be a contour pixel (see figure \ref{fig20n}).

For each pixel $q$, the best match $p^*$ is computed exploring all $N(p)$
in $A$ and chosen so as the similarity measure $S(q,p^*)$ is minimal.

In order to determine the required conditions such that the computed best match $p^*$ of $q$
will be a contour pixel, we study the variation of $S(q,p)$ related to $N(p)$ structure and the
intensities of pixels inside of $N(p)$ and $N(p)$. We note that the central pixel $p$ may be
or no an outline pixel.\\

Let:\\
- $S(q,p)$ be the value of the similarity measure such that $N(q), N(p)$ have the same structure: $q,p$ are
both contour pixels, and the directions of the boundaries inside $N(q), N(p)$ are identical $d_q=d_p$.\\
- $S(q,p')$ be the values of the similarity measure such that $N(q), N(p')$ have different structures
and $p'$ may be or no a contour pixel.

$S(q,p)$, $S(q,p')$ are given by the equations \ref{eq8} and \ref{eq7} where:\\
- $N_p^1(i,j), N_p^2(i,j),
N_p^3(i,j), N_p^4(i,j)$, $N_q^1(i,j), N_q^2(i,j), N_q^3(i,j), N_q^4(i,j)$ are pixels of regions
$N_p^1, N_p^2, N_p^3, N_p^4$, $N_q^1, N_q^2, N_q^3, N_q^4$,\\
- $N_{p'}^1(i,j), N_{p'}^2(i,j),N_{p'}^3(i,j), N_{p'}^4(i,j)$ are pixels of regions $N_{p'}^1, N_{p'}^2,
N_{p'}^3, N_{p'}^4$.

These regions are defined from the two assumed different orientations $d_q$ and $d_{p'}$ of the boundaries
in $N(q)$ and $N(p')$ (see figure \ref{fig20n}):

\begin{figure}[ht]
\centering
  \includegraphics[width=10cm]{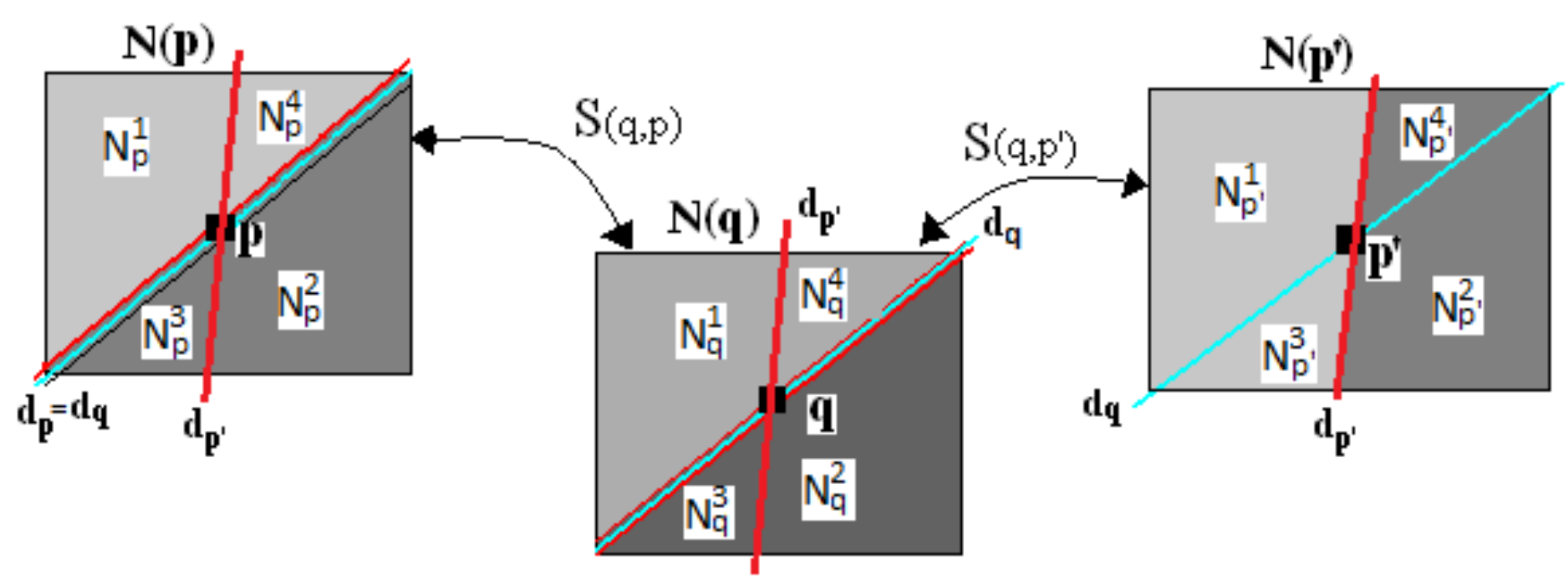}
  \caption{A same direction for the boundaries inside $N(q)$ and $N(p)$ and different directions for the
  boundaries inside $N(q)$ and $N(p')$, $p'$ is a contour pixel}
  \label{fig20n}
\end{figure}

{ \scriptsize
\begin{equation}\label{eq8}
S(q,p)=\sum_{i}\sum_{j}(N_q^1(i,j)-N_p^1(i,j))^2
        +(N_q^2(i,j)-N_p^2(i,j))^2+(N_q^3(i,j)-N_p^3(i,j))^2
        +(N_q^4(i,j)-N_p^4(i,j))^2
\end{equation}
}

{ \scriptsize
\begin{equation}\label{eq7}
S_m(q,p')=\sum_{i}\sum_{j}(N_q^1(i,j)-N_{p'}^1(i,j))^2
        +(N_q^2(i,j)-N_{p'}^2(i,j))^2+(N_q^3(i,j)-N_{p'}^3(i,j))^2
        +(N_q^4(i,j)-N_{p'}^4(i,j))^2
\end{equation}
}
To compare between $S(q,p)$ and $S(q,p')$, we compute $\bigtriangleup S=S(q,p')-S(q,p)$ such that
$p$ and $p'$ are assumed to be pixels of the same region which implies that $N_p^1=N_{p'}^1, N_p^2= N_{p'}^2$.
We get:

{\scriptsize
\begin{equation}\label{eq9}
\bigtriangleup S=\sum_{i}\sum_{j}(N_q^3(i,j)-N_{p'}^3(i,j))^2
        +(N_q^4(i,j)-N_{p'}^4(i,j))^2
        -(N_q^3(i,j)-N_p^3(i,j))^2
        -(N_q^4(i,j)-N_p^4(i,j))^2
\end{equation}
}

\begin{figure}[ht]
\centering
  \includegraphics[width=10cm]{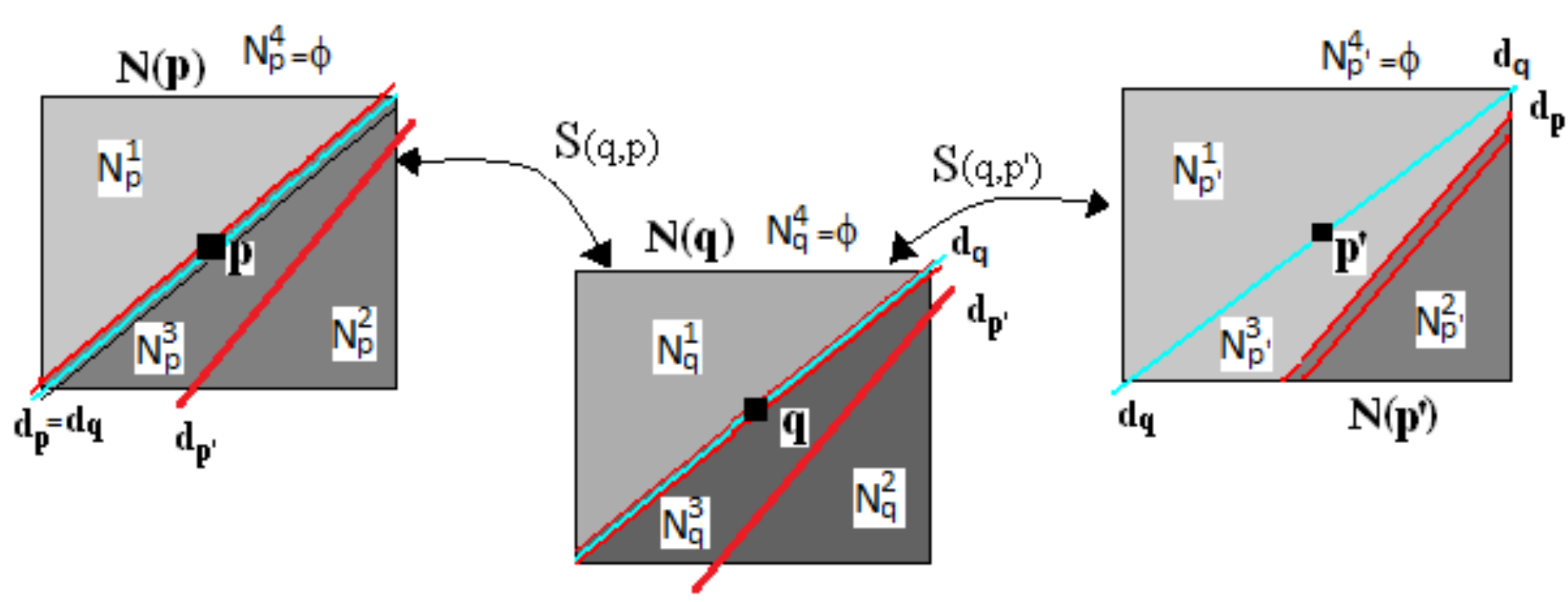}
  \caption{$N(p')$ having different structure than $N(q)$ and $p'$ isn't a contour pixel}
  \label{fig21n}
\end{figure}

\begin{figure}[ht]
\centering
  \includegraphics[width=10cm]{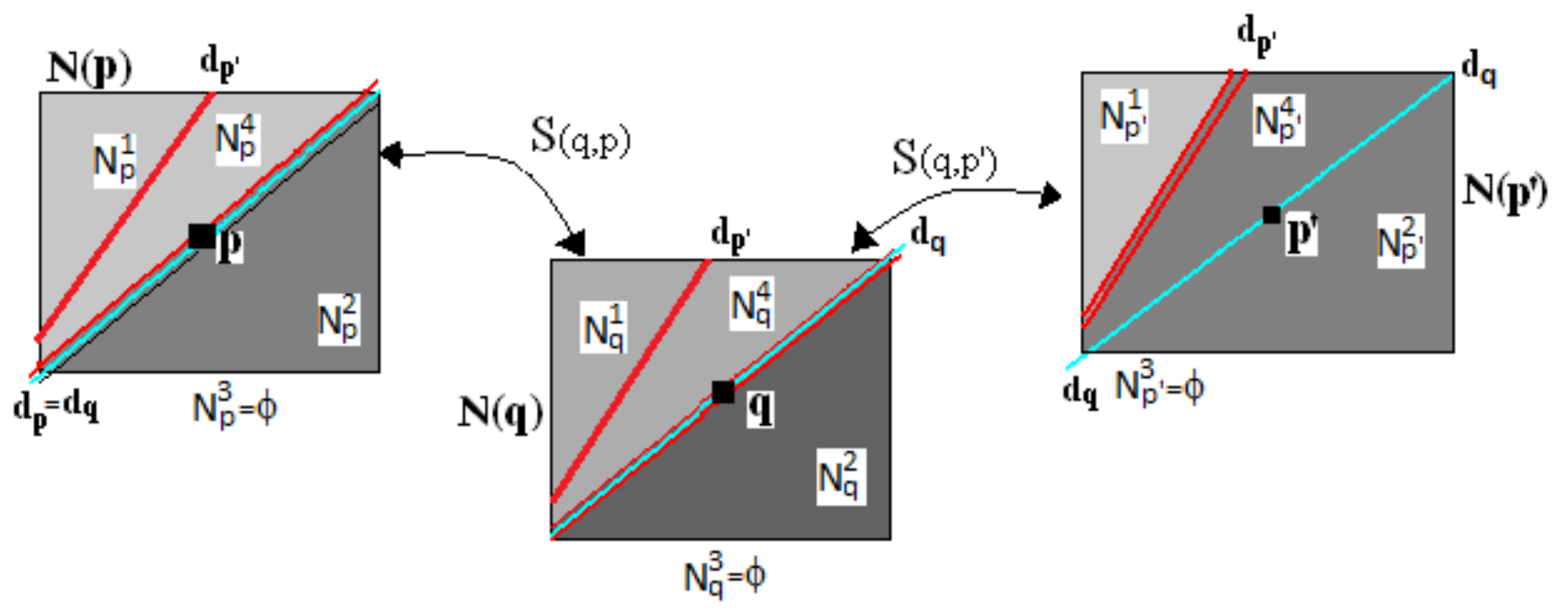}
  \caption{$N(p')$ having other different structure than $N(q)$ and $p'$ isn't a contour pixel}
  \label{fig22n}
\end{figure}

The same equation is obtained in case where $N(p')$ has a different structure than $N(p)$ and
$p'$ isn't a contour pixel (see figures \ref{fig21n}, \ref{fig22n}).\\

The neighbor $N(p)$ having the same structure as $N(q)$ will be selected if the difference
$\bigtriangleup S=S_m(q,p')-S(q,p)$ is positive.

Let $I^b_A, I^f_A$ (resp. $I^b_B, I^f_B$ ) be the average of pixels intensities of the two regions
of $N(p)$ (resp. $N(q)$). Otherwise, the average of intensities will concern all pixels of $N(p)$, $N(q)$
and will be noted $I^b_A$, $I^b_B$. Note here that the pixels of the same region in $N(p)$ and $N(q)$
are assumed have different values of intensities and then we are in complicating factors such
as gradually changing shading or texture, but there is a boundary between them such that human can draw it.

We will use the following notations (see figure \ref{fig20n}):\\

- $I^b_A, I'^b_A, I''^b_A$ be the average, minimal, maximal intensity of all $N_p^1(i,j),
N_p^4(i,j), N_{p'}^1(i,j), N_{p'}^3(i,j)$ pixels,\\
- $I^f_A, I'^f_A, I''^f_A$ be the average, minimal, maximal intensity of all $N_p^2(i,j),
N_p^3(i,j), N_p'^2(i,j), N_p'^4(i,j)$ pixels,\\
- $I^b_B, I'^b_B$, $I''^b_B$ be the average, minimal, maximal intensity of all $N_q^1(i,j),
N_q^4(i,j)$ pixels , \\
- $I^f_B, I'^f_B, I''^f_B$ be the average, minimal, maximal intensity of all $N_q^2(i,j),
N_q^3(i,j)$ pixels.\\

\begin{figure}[ht]
\centering
  \includegraphics[width=10cm]{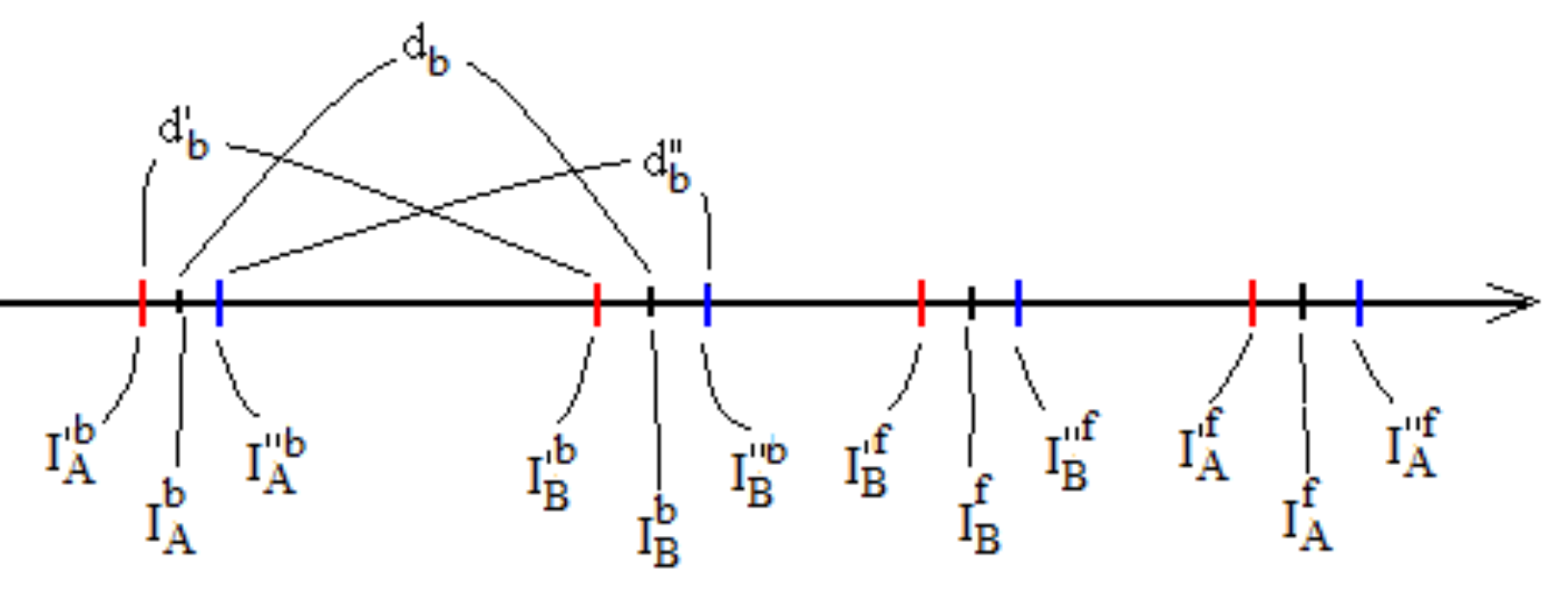}
  \caption{Example of possible values of $I^b_A, I^b_B, I^f_B, I^f_A$}
  \label{fig23n}
\end{figure}

Firstly, we assume that $I^b_A<I^b_B<I^f_B<I^f_A$ (see figure \ref{fig23n}), the terms of the
equation \ref{eq9} can be written as follow:

{ \small
\begin{equation}\label{eq10}
  \sum_{i}\sum_{j}(N_q^3(i,j)-N_{p'}^3(i,j))^2> n_1(I'^f_B-I''^b_A)^2
\end{equation}
}
{ \small
\begin{equation}\label{eq11}
 \sum_{i}\sum_{j}(N_q^4(i,j)-N_{p'}^4(i,j))^2> n_2(I''^b_B-I'^f_A)^2
\end{equation}
}
{ \small
\begin{equation}\label{eq12}
-\sum_{i}\sum_{j}(N_q^3(i,j)-N_p^3(i,j))^2 > -n_1(I'^f_B-I''^f_A)^2
\end{equation}
}
{ \small
\begin{equation}\label{eq13}
  -\sum_{i}\sum_{j}(N_q^4(i,j)-N_p^4(i,j))^2 > -n_2(I''^b_B-I'^b_A)^2
\end{equation}
}

where $n_1$ and $n_2$ are respectively the number of pixels of the regions $N_p^3, N_{p'}^3, N_q^3$
and $N_p^4, N_{p'}^4, N_q^4$ (see figure \ref{fig20n}). The value of $\bigtriangleup S$ is then given by the equation \ref{eq16}.\\

{ \small
\begin{equation}\label{eq16}
\bigtriangleup S > n_1(I'^f_B-I''^b_A)^2+n_2(I''^b_B-I'^f_A)^2-n_1(I'^f_B-I''^f_A)^2
-n_2(I''^b_B-I'^b_A)^2
\end{equation}
}

We will note:\\
- $d_A=I^f_A-I^b_A$, $d'_A=I'^f_A-I'^b_A$, $d''_A=I''^f_A-I''^b_A$\\
- $d_B=I^f_B-I^b_B$, $d'_B=I'^f_B-I'^b_B$, $d''_B=I''^f_B-I''^b_B$, \\
- $d_b=I^b_B-I^b_A$, $d'_b=I'^b_B-I'^b_A$, $d''_b=I''^b_B-I''^b_A$, \\
-$\bigtriangleup I^f_B=I''^f_B-I'^f_B$, $\bigtriangleup I^f_A=I''^f_A-I'^f_A$,
-$\bigtriangleup I^b_B=I''^b_B-I'^b_B$, $\bigtriangleup I^b_A=I''^b_A-I'^b_A$\\

As:\\

$n_1(I'^f_B-I''^b_A)^2-n_1(I'^f_B-I''^f_A)^2$= $n_1(I'^f_B-I''^b_A+I'^f_B-I''^f_A)(I'^f_B-I''^b_A-I'^f_B+I''^f_A)$\\
=$n_1(I'^f_B-I''^b_A+I'^f_B-I''^f_A)(d''_A)$

$I'^f_B-I''^b_A+I'^f_B-I''^f_A=2(I'^f_B-I'^b_B)+2(I'^b_B-I'^b_A)+2(I'^b_A-I''^b_A)-(I''^f_A-I''^b_A)$
=$2d'_B+2d'_b-2\bigtriangleup I^b_A-d''_A$\\
and\\
$n_2(I''^b_B-I'^f_A)^2-n_2(I''^b_B-I'^b_A)^2$= $n_2(I''^b_B-I'^f_A+I''^b_B-I'^b_A)(I''^b_B-I'^f_A-I''^b_B+I'^b_A)$=
$n_2(I''^b_B-I'^f_A+I''^b_B-I'^b_A)(-I'^f_A+I'^b_A)=n_2(2d''_b-d'_A+2\bigtriangleup I^b_A)(-d'_A)$

We obtain then;
$\bigtriangleup S >n_1d''_A(2d'_B+2d'_b-2\bigtriangleup I^b_A-d''_A)-n_2d'_A(d''_b+d'_b)$

We can consider that $d''_B=d'_B=d_B$, $d''_A=d'_A=d_A$, $d''_b=d'_b=d_b$
(see figure \ref{fig23n}). We write then:

{ \small
\begin{equation}\label{eq16n}
\bigtriangleup S >n_1d_A(2d_B+2d_b-d_A-2\bigtriangleup I^b_A)-n_2d_A(2d_b-d_A+2\bigtriangleup I^b_A)
\end{equation}
}
The pixel $q$ will be classified as a contour pixel if all values of $S(q,p')$
are greater than $S(q,p)$ and then $\bigtriangleup S>0$.\\

As $d_A>0$, it is sufficient to have:\\
$2d_B+2d_b-d_A-2\bigtriangleup I^b_A>0$ and $2d_b-d_A+2\bigtriangleup I^b_A<0$, this implies:

{ \small
\begin{equation}\label{eq16nn}
2d_b+2\bigtriangleup I^b_A< d_A<2d_B+2d_b-2\bigtriangleup I^b_A
\end{equation}
}

The same reasoning is applied for other combination of positions of $I^b_A, I^b_B, I^f_A, I^f_B$, we found:

- if $I^b_A<I^b_B<I^f_A<I^f_B$, $\bigtriangleup S >0$ if $d_A>2d_b+2\bigtriangleup I^b_A$,
$d_A<2d_b+2d_B$.

-if $I^b_B<I^b_A<I^f_B<I^f_A$, $\bigtriangleup S >0$ if $d_A<2d_b+2d_B-2\bigtriangleup I^b_A$,
 ($d_A>2d_b$ is verified)\\

- if $I^b_B<I^b_A<I^f_A<I^f_B$, $\bigtriangleup S >0$ if $d_A<2d_b+2d_B$,
($d_A>2d_b$ is verified)


Secondly, we assume that $d_A<0$, we obtain:

- if $I^f_A<I^f_B<I^b_B<I^b_A$: $\bigtriangleup S >0$
if $d_A<2d_b-2\bigtriangleup I^b_A$, $d_A>2d_B+2d_b+2\bigtriangleup I^b_A$.

- if $I^f_B<I^f_A<I^b_B<I^b_A$: $\bigtriangleup S >0$ if $d_A<2d_b-2\bigtriangleup I^b_A$,
$d_A>2d_B+2d_b$.

- if $I^f_A<I^f_B<I^b_A<I^b_B$: $\bigtriangleup S >0$ if  $d_A>2d_B+2d_b+2\bigtriangleup I^b_A$,
($d_A<2d_b$ is verified).

- if $I^f_B<I^f_A<I^b_A<I^b_B$: $\bigtriangleup S >0$ if  $d_A>2d_B+2d_b$, ($d_A<2d_b$ is verified)

\end{document}